%% file: root.tex
\documentclass[journal]{IEEEtran}
\usepackage{etex}
\reserveinserts{100}
\usepackage{mathtools}
\usepackage{graphics, theorem, times, amsfonts, graphicx, amssymb, cite}
\usepackage[outdir=figures_eps_to_pdf/]{epstopdf}
\usepackage{tikz}

\usetikzlibrary{shapes,arrows}
\usepackage{standalone}
\usepackage{pgfplots}
\usepackage{color}
\usepackage{bbold}
\usepackage{setspace}
\usepackage{hyperref}
\usepackage{multirow}
\usepackage{rotating}
\usepackage{comment}
\usepackage{flushend}
\usepackage[font=small,labelfont=bf]{caption}
\usepackage{subcaption}
\usepackage[affil-it]{authblk}
\usepackage{bm}
\usepackage{algorithm}
\usepackage{algorithmic}
\usepackage{enumerate}
\usepackage[shortlabels]{enumitem}
\usepackage{morefloats}
\usepackage{setspace, tikz}

\usetikzlibrary{matrix} % for block alignment
\usetikzlibrary{arrows} % for arrow heads
\usetikzlibrary{calc} % for manimulation of coordinates
\usetikzlibrary{shapes}
\usetikzlibrary{arrows}
\usetikzlibrary{positioning}
\usetikzlibrary{backgrounds, fit}

\input{plot_appearance.tex}

\input{mysymbol.tex}
\input{my_sections}

\newcommand{\QED}{\hfill\ensuremath{\blacksquare}}

\newif\ifcompiletikz
\compiletikztrue
\newtheorem{assumption}{\hspace{0pt}\bf Assumption}

\newtheorem{proposition}{\hspace{0pt}\bf Proposition}

\newtheorem{theorem}{\hspace{0pt}\bf Theorem}

\newtheorem{remark}{\hspace{0pt}\bf Remark}

\newtheorem{definition}{\hspace{0pt}\bf Definition}

\title{Graph Neural Networks: \\ Architectures, Stability and Transferability}
\author{Luana Ruiz \qquad Fernando Gama \qquad Alejandro Ribeiro 
\thanks{University of Pennsylvania, email at (rubruiz, fgama, aribeiro) @seas.upenn.edu. Supported by NSF HDR TRIPODS, Award \#1934960.}
}

\begin{document}

\thispagestyle{empty}
\maketitle

\begin{abstract} Graph Neural Networks (GNNs) are information processing architectures for signals supported on graphs. They are presented here as generalizations of convolutional neural networks (CNNs) in which individual layers contain banks of graph convolutional filters instead of banks of classical convolutional filters. Otherwise, GNNs operate as CNNs. Filters are composed with pointwise nonlinearities and stacked in layers. It is shown that GNN architectures exhibit equivariance to permutation and stability to graph deformations. These properties help explain the good performance of GNNs that can be observed empirically. It is also shown that if graphs converge to a limit object, a graphon, GNNs converge to a corresponding limit object, a graphon neural network. This convergence justifies the transferability of GNNs across networks with different number of nodes. Concepts are illustrated by the application of GNNs to recommendation systems, decentralized collaborative control, and wireless communication networks.

\end{abstract}


\input{00-introduction.tex}

%%%%%%%%%%%%%%%%%%%%%%%%%%%%%%%%%%%%%%%%%%%%%%%%%%%%%%%%%%%%%%%%%%%%
%%%   S   E   C   T   I   O   N   %%%%%%%%%%%%%%%%%%%%%%%%%%%%%%%%%%
%%%%%%%%%%%%%%%%%%%%%%%%%%%%%%%%%%%%%%%%%%%%%%%%%%%%%%%%%%%%%%%%%%%%
%

\input{ch8_symbols.tex}
\input{ch8_sec01_classification.tex}

\input{ch8_sec03_gnns.tex}
\input{FG-stability.tex}

\input{LR-transferability.tex}

\input{LR-numerical.tex}

\input{99_conclusions}

\urlstyle{same}
\bibliographystyle{IEEEtran}
\bibliography{%
bib_files/myIEEEabrv,% 
bib_files/graphMachineLearningBiblio,%
bib_files/bib-graphon,%
bib_files/edgeNetsBiblio%
}

\end{document}

%% file: plot_appearance.tex
\def\myfactor{1.0}
\def\unit{\myfactor cm}

\pgfplotsset{ ylabel near ticks,                 % Make x axis label appear close to figure
              xlabel near ticks,                 % Make y axis label appear close to figure
              tick label style = {font=\footnotesize},   % Write numbers in tiny font
              label style = {font=\footnotesize}, % Write axes labels in footnote size font
              title style = {font=\footnotesize},
            }

\tikzstyle{empty node} = [ circle, 
                     draw = black,
                     text = black, 
                     minimum size = 0.8*\unit]

\tikzstyle{node} = [ empty node, 
                     fill = blue!50,
                     draw = blue!50,
                     text = white]

\tikzstyle{edge}                 = [shorten >=1pt, shorten <=1pt]
\tikzstyle{directed edge}        = [edge, -stealth]
\tikzstyle{double directed edge} = [edge, stealth-stealth]
\tikzstyle{tight edge}                 = [shorten >=0pt, shorten <=0pt]

\tikzstyle{block} = [ rectangle,
                      minimum width = \unit,
                      minimum height = \unit,
                      fill = blue!15,
                      draw = black,
                      text = black]

%% file: mysymbol.tex
\newenvironment{myproof}[1][$\!\!$]{{\noindent\bf Proof #1: }}
                         {\hfill\QED\medskip}

\newenvironment{mylist}{\begin{list}{}{  \setlength{\itemsep  }{5pt} \setlength{\parsep    }{  0pt}
                                         \setlength{\parskip  }{0pt} \setlength{\topsep    }{  5pt}
                                         \setlength{\partopsep}{0pt} \setlength{\leftmargin}{ 10pt}
                                         \setlength{\labelsep }{5pt} \setlength{\labelwidth}{- 5pt}}}
                          {\end{list}}

\newcounter{excercise}
\newcounter{excercisepart}

\definecolor{pennblue}{cmyk}{1,0.65,0,0.30}
\definecolor{pennred}{cmyk}{0,1,0.65,0.34}
\definecolor{mygreen}{rgb}{0.10,0.50,0.10}
\newcommand \red[1]         {{\color{red}#1}}
\newcommand \black[1]         {{\color{black}#1}}
\newcommand \blue[1]        {{\color{blue}#1}}

\newcommand \bulletcolor[1] {{\color{pennblue}#1}}
\def \arrowbullet {\bulletcolor{$\ \Rightarrow\ $}}

\def \lam   {\lambda}

\def \ovP {\overline{P}}

\def \ovX {\overline{X}}
\def \ovbbX {\overline{\bbX}}
\def \ovp {\overline{p}}
\def \ovbbp {\overline{\bbp}}
\def \ovr {\overline{r}}
\def \ova {\overline{a}}
\def \ovc {\overline{c}}
\def \ovalpha {\overline{\alpha}}

\def \reals    {{\mathbb R}}

\newcommand{\argmax}{\operatornamewithlimits{argmax}}
\newcommand{\argmin}{\operatornamewithlimits{argmin}}

\def\mbE{{\ensuremath{\mathbb E}}}

\def\mbI{{\ensuremath{\mathbb I}}}

\def\mbZ{{\ensuremath{\mathbb Z}}}
\def\ccalC{{\ensuremath{\mathcal C}}}

\def\ccalE{{\ensuremath{\mathcal E}}}

\def\ccalH{{\ensuremath{\mathcal H}}}
\def\ccalI{{\ensuremath{\mathcal I}}}

\def\ccalN{{\ensuremath{\mathcal N}}}
\def\ccalO{{\ensuremath{\mathcal O}}}
\def\ccalP{{\ensuremath{\mathcal P}}}

\def\ccalT{{\ensuremath{\mathcal T}}}
\def\ccalX{{\ensuremath{\mathcal X}}}

\def\ccal0{{\ensuremath{\mathcal 0}}}
\def\tdu{{\ensuremath{\tilde u}}}

\def\tdx{{\ensuremath{\tilde x}}}

\def\bbone{{\ensuremath{\mathbf 1}}}

\def\bbE{{\ensuremath{\mathbf E}}}
\def\bbF{{\ensuremath{\mathbf F}}}
\def\bbG{{\ensuremath{\mathbf G}}}
\def\bbH{{\ensuremath{\mathbf H}}}
\def\bbI{{\ensuremath{\mathbf I}}}

\def\bbP{{\ensuremath{\mathbf P}}}

\def\bbW{{\ensuremath{\mathbf W}}}
\def\bbU{{\ensuremath{\mathbf U}}}
\def\bbV{{\ensuremath{\mathbf V}}}
\def\bbS{{\ensuremath{\mathbf S}}}

\def\bbX{{\ensuremath{\mathbf X}}}
\def\bbY{{\ensuremath{\mathbf Y}}}
\def\bbZ{{\ensuremath{\mathbf Z}}}
\def\bba{{\ensuremath{\mathbf a}}}

\def\bbe{{\ensuremath{\mathbf e}}}

\def\bbh{{\ensuremath{\mathbf h}}}

\def\bbl{{\ensuremath{\mathbf l}}}

\def\bbp{{\ensuremath{\mathbf p}}}

\def\bbr{{\ensuremath{\mathbf r}}}

\def\bbu{{\ensuremath{\mathbf u}}}
\def\bbv{{\ensuremath{\mathbf v}}}

\def\bbx{{\ensuremath{\mathbf x}}}
\def\bby{{\ensuremath{\mathbf y}}}
\def\bbz{{\ensuremath{\mathbf z}}}
\def\bb0{{\ensuremath{\mathbf 0}}}
\def\hbS{{\hat{\ensuremath{\mathbf S}} }}

\def\hbx{{\hat{\ensuremath{\mathbf x}} }}

\def\tbU{{\tilde{\ensuremath{\mathbf U}} }}

\def\tbX{{\tilde{\ensuremath{\mathbf X}} }}

\def\bbalpha{\boldsymbol{\alpha}}
\def\bbbeta{\boldsymbol{\beta}}
\def\bbgamma{\boldsymbol{\gamma}}
\def\bbdelta{\boldsymbol{\delta}}
\def\bbepsilon{\boldsymbol{\epsilon}}
\def\bbvarepsilon{\boldsymbol{\varepsilon}}
\def\bbzeta{\boldsymbol{\zeta}}
\def\bbeta{\boldsymbol{\eta}}
\def\bbtheta{\boldsymbol{\theta}}
\def\bbvartheta{\boldsymbol{\vartheta}}

\def\bbupsilon{\boldsymbol{\upsilon}}
\def\bbiota{\boldsymbol{\iota}}
\def\bbkappa{\boldsymbol{\kappa}}
\def\bblambda{\boldsymbol{\lambda}}

\def\bbmu{\boldsymbol{\mu}}
\def\bbnu{\boldsymbol{\nu}}
\def\bbxi{\boldsymbol{\xi}}
\def\bbpi{\boldsymbol{\pi}}
\def\bbrho{\boldsymbol{\rho}}
\def\bbvarrho{\boldsymbol{\varrho}}
\def\bbvarsigma{\boldsymbol{\varsigma}}
\def\bbphi{\boldsymbol{\phi}}
\def\bbvarphi{\boldsymbol{\varphi}}
\def\bbchi{\boldsymbol{\chi}}
\def\bbpsi{\boldsymbol{\psi}}
\def\bbomega{\boldsymbol{\omega}}
\def\bbGamma{\boldsymbol{\Gamma}}
\def\bbDelta{\boldsymbol{\Delta}}
\def\bbTheta{\boldsymbol{\Theta}}
\def\bbLambda{\boldsymbol{\Lambda}}
\def\bbXi{\boldsymbol{\Xi}}
\def\bbPi{\boldsymbol{\Pi}}
\def\bbSigma{\boldsymbol{\Sigma}}
\def\bbUpsilon{\boldsymbol{\Upsilon}}
\def\bbPhi{\boldsymbol{\Phi}}
\def\bbPsi{\boldsymbol{\Psi}}

\def \lam    {\lambda}

%% file: my_sections.tex
\usepackage{needspace}

%% file: 00-introduction.tex
% !TEX root = root.tex

%%%%%%%%%%%%%%%%%%%%%%%%%%%%%%%%%%%%%%%%%%%%%%%%%%%%%%%%%%%%%%%%%%%%%%%%%%%%%%%%
%   S   E   C   T   I   O   N   %%%%%%%%%%%%%%%%%%%%%%%%%%%%%%%%%%%%%%%%%%%%%%%%
%%%%%%%%%%%%%%%%%%%%%%%%%%%%%%%%%%%%%%%%%%%%%%%%%%%%%%%%%%%%%%%%%%%%%%%%%%%%%%%%
%
\section{Introduction} \label{sec:intro}

\IEEEPARstart{G}{raphs} can represent lexical relationships in text analysis \cite{Joachims97-TextCategorization, Mikolov13-Word2vec, Defferrard17-ChebNets}, product or customer similarities in recommendation systems \cite{Huang18-RatingGSP, Ying18-LargeScaleGNN, Monti17-Movie}, or agent interactions in multiagent robotics \cite{Tolstaya19-Flocking, Sartoretti19-PRIMAL, Li20-Planning}. {Although otherwise unrelated, these applications share the presence of signals associated with nodes -- words, ratings or perception -- out of which we want to extract some information -- text categories, ratings of other products, or control actions.} If data is available, we can formulate empirical risk minimization (ERM) problems to learn these data-to-information maps. However, it is a form of ERM in which the graph plays a central role in describing relationships between signal components and, therefore, one in which it should be leveraged. Graph Neural Networks (GNNs) are parametrizations of learning problems in general and ERM problems in particular that achieve this goal. 

{In a ERM problem, we are given input-output pairs in a training set, and we want to find a function that best approximates the input-output map according to a given risk (Sec. \ref{sec_ch8_learning}). This function is later used to estimate the outputs associated with inputs that were not part of the training set. We say that the function has been \emph{trained} and that we have \emph{learned} to estimate outputs. {This simple statement hides the fact that ERM problems do not make sense unless we make assumptions on how the function \emph{generalizes} from the training set to unobserved samples (Sec. \ref{sec_parametrizations}).} We can, for instance, assume that the map is linear, or, to be in tune with the times, that the map is a deep neural network \cite{Goodfellow16-DeepLearning}.}

A characteristic shared by arbitrary linear and fully connected neural network parametrizations is that they do not scale well with the dimensionality of the input signals. {This is best known in the case of signals in Euclidean space -- time and images -- where many successful examples of scalable linear processing are based on \emph{convolutional} filters, and of scalable nonlinear processing on \emph{convolutional} neural networks (CNNs)}. In this paper we describe \emph{graph} filters \cite{Sandryhaila13-DSPG, Segarra17-Linear} and \emph{graph} neural networks \cite{Bruna14-DeepSpectralNetworks, Defferrard17-ChebNets, Gama19-Architectures, Kipf17-GCN, Isufi20-EdgeNets} as analogous of convolutional filters and CNNs, but adapted to process signals supported on graphs (Sec. \ref{sec_ch8_gnns}). 
%Both of these concepts are simple. 
A graph filter is a polynomial in a matrix representation of the graph. Out of this definition, we build a graph perceptron with the addition of a pointwise nonlinear function to process the output of a graph filter (Sec. \ref{sec_perceptrons}). Graph perceptrons can be layered to build a multilayer GNN (Sec. \ref{sec_ch8_gnns_multiple_layers}), and individual layers are augmented from single filters to filter banks to build multiple feature GNNs (Sec. \ref{sec_ch8_gnn_multiple_features}).

{At this juncture, an important question is whether graph filters and GNNs do for signals supported on graphs what convolutional filters and CNNs do for Euclidean data. In other words, do they enable scalable processing of signals supported on graphs? A growing body of empirical work shows that this is true to some extent -- although results are not as impressive as in the case of voice and image processing.} As an example that we can use to illustrate the advantages of graph filters and GNNs, consider a recommendation system (Sec. \ref{sec_reco_systems}) in which we want to use past ratings that customers have given to products to predict future ratings \cite{Harper16-MovieLens}. Collaborative filtering solutions build a graph of product similarities and interpret customer ratings as signals supported on the product similarity graph \cite{Huang18-RatingGSP}. We then use past ratings to construct a training set and learn to fill in the ratings that a given customer would give to products not yet rated. {Empirical results do show that graph filters and GNNs work in recommendation systems with large number of products in which linear maps and fully connected neural networks do not \cite{Huang18-RatingGSP, Ying18-LargeScaleGNN, Monti17-Movie}.} {In fact, this example leads to three empirical observations that motivate this paper (Sec. \ref{sec_reco_systems_results}):}
%
%%%%%%%%%%%%%%%%%%%%%%%%%%%%%%%%%%%%%%%%%%%%%%%%%%%%%%%%%%%%%%%%%%%%%%%%%%%%%%%%
%   L   I   S   T   %%%%%%%%%%%%%%%%%%%%%%%%%%%%%%%%%%%%%%%%%%%%%%%%%%%%%%%%%%%%
%%%%%%%%%%%%%%%%%%%%%%%%%%%%%%%%%%%%%%%%%%%%%%%%%%%%%%%%%%%%%%%%%%%%%%%%%%%%%%%%
%
\begin{mylist}
\item[{\bf(O1)}] {Graph filters produce better rating estimates than arbitrary linear parametrizations and GNNs produce better estimates than arbitrary (fully connected) neural networks, provided that sufficient training data is available.}
\item[{\bf(O2)}] GNNs predict ratings better than graph filters.
\item[{\bf(O3)}] A GNN that is trained on a graph with a certain number of nodes can be executed in a graph with a larger number of nodes and still produce good rating estimates.
\end{mylist}
%
%%%%%%%%%%%%%%%%%%%%%%%%%%%%%%%%%%%%%%%%%%%%%%%%%%%%%%%%%%%%%%%%%%%%%%%%%%%%%%%%
%   M   A   I   N       M   A   T   T   E   R   %%%%%%%%%%%%%%%%%%%%%%%%%%%%%%%%
%%%%%%%%%%%%%%%%%%%%%%%%%%%%%%%%%%%%%%%%%%%%%%%%%%%%%%%%%%%%%%%%%%%%%%%%%%%%%%%%
%
Observations (O1)-(O3) support advocacy for the use of GNNs, at least in recommendation systems. But they also spark three interesting questions: (Q1) Why do graph filters and GNNs outperform linear transformations and fully connected neural networks? (Q2) Why do GNNs outperform graph filters? (Q3) Why do GNNs transfer to networks with different number of nodes? In this paper we present three theoretical analyses that help answer these questions:
%
%%%%%%%%%%%%%%%%%%%%%%%%%%%%%%%%%%%%%%%%%%%%%%%%%%%%%%%%%%%%%%%%%%%%%%%%%%%%%%%%
%   L   I   S   T   %%%%%%%%%%%%%%%%%%%%%%%%%%%%%%%%%%%%%%%%%%%%%%%%%%%%%%%%%%%%
%%%%%%%%%%%%%%%%%%%%%%%%%%%%%%%%%%%%%%%%%%%%%%%%%%%%%%%%%%%%%%%%%%%%%%%%%%%%%%%%
%
\begin{mylist}
\item[{\bf Equivariance.}] Graph filters and GNNs are equivariant to permutations of the graph (Sec. \ref{sec_ch8_gnns}). 
\item[{\bf Stability.}] GNNs provide a better tradeoff between discriminability and stability to graph perturbations (Sec. \ref{sec:stability}). 
\item[{\bf Transferability.}] As graphs converge to a limit object, a graphon, GNN outputs converge to outputs of a corresponding limit object, a graphon neural network (Sec. \ref{sec:transferability}). 
\end{mylist}
%
%%%%%%%%%%%%%%%%%%%%%%%%%%%%%%%%%%%%%%%%%%%%%%%%%%%%%%%%%%%%%%%%%%%%%%%%%%%%%%%%
%   M   A   I   N       M   A   T   T   E   R   %%%%%%%%%%%%%%%%%%%%%%%%%%%%%%%%
%%%%%%%%%%%%%%%%%%%%%%%%%%%%%%%%%%%%%%%%%%%%%%%%%%%%%%%%%%%%%%%%%%%%%%%%%%%%%%%%
%
These properties show that GNNs have strong \emph{generalization} potential. Equivariance to permutations implies that nodes with analogous neighbor sets making analogous observations perform the same operations. Thus, we can learn to, say, fill in the ratings of a product from the ratings of another product in another part of the network if the local structures of the graph are the same (Fig. \ref{fig_generalization}). This helps explain why graph filters outperform linear transforms and GNNs outperform fully connected neural networks [cf. observation (O1)]. Stability to graph deformations affords a stronger version of this statement. We can learn to generalize across different products if the local neighborhood structures are similar, not necessarily identical (Fig. \ref{fig_generalization_with_wiggliness}). GNNs possess better stability than graph filters for the same level of discriminability, which helps explain why GNNs outperform graph filters [cf. observation (O2)]. 
%{This property is further illustrated in the decentralized robot control problem of Section \ref{sbs:stab}.} 
The convergence of GNNs towards graphon neural networks delineated under the transferability heading explains why GNNs can be trained and executed in graphs of different sizes [cf. observation (O3)]. 
%{which is further illustrated in the wireless resource allocation problem of Section \ref{sbs:transf}.} 
{It is important to note that analogous of these properties hold for CNNs.} They are equivariant to translations and stable to Euclidean space deformations \cite{Mallat12-Scattering} and have well defined continuous time limits.

We focus on a tutorial introduction to GNNs and on describing some of their fundamental properties. This focus renders several relevant questions out of scope. Most notably, we do not discuss training \cite{assran2020convergence, Kingma15-ADAM}. The role of proper optimization techniques, the selection of proper optimization objectives, and the realization of graph filters is critical in ensuring that the \emph{potential} for generalization implied by equivariance, stability, and transferability is actually \emph{realized.} References for the interested reader are provided in Sec. \ref{sec_further_reading}.

\input{01-context.tex}

%% file: 01-context.tex
% !TEX root = root.tex

\subsection{Context and Further Reading}\label{sec_further_reading}

The field of graph signal processing (GSP) has developed over the last decade \cite{Sandryhaila13-DSPG, Shuman13-SPG, Ortega18-GSP}. Central to developments in GSP is the notion of graph convolutional filters \cite{Sandryhaila13-DSPG, Shuman13-SPG, Segarra17-Linear, Isufi17-ARMA, tremblay2018design}. GNNs arose as nonlinear extensions of graph filters, obtained by the addition of pointwise nonlinearities to the processing pipeline \cite{Bruna14-DeepSpectralNetworks, Defferrard17-ChebNets, Kipf17-GCN, Gama19-Architectures, moura2017topology}. Several implementations of GNNs have been proposed. These include graph convolutional filters implemented in the spectral domain \cite{Bruna14-DeepSpectralNetworks}, implementations of graph filters with Chebyshev polynomials \cite{Defferrard17-ChebNets} and  ordinary polynomials \cite{Gama19-Architectures, moura2018convolution}. One can also encounter GNNs described in terms of local aggregation functions \cite{Kipf17-GCN, Xu19-GIN}. These can be seen as particular cases of GNNs that use graph filters of order 1, {because local aggregation operations can be described as matrix multiplications with some matrix representation of the graph.} This results in a parametrization with lower representation power than those in \cite{Bruna14-DeepSpectralNetworks, Defferrard17-ChebNets, Gama19-Architectures}. 

It is important to point out that the GNNs in \cite{Bruna14-DeepSpectralNetworks, Defferrard17-ChebNets, Gama19-Architectures} are equivalent in the sense that they span the exact same set of maps. Thus, although we use the polynomial description of \cite{Gama19-Architectures}, the results we present apply irrespectively of implementation. The architectures in \cite{Kipf17-GCN, Xu19-GIN}, being restricted to filters of order 1, span a subset of the maps that can be represented by the more generic GNNs in \cite{Bruna14-DeepSpectralNetworks, Defferrard17-ChebNets, Gama19-Architectures}. Hence, results also apply to \cite{Kipf17-GCN, Xu19-GIN}, except for discriminability discussions which require the use of higher order graph filters. Equivalence notwithstanding, these architectures may differ in their ease of training, leading to different performance in practice. 

GNNs using linear transforms other than graph filters have also been proposed \cite{Isufi20-EdgeNets, Velickovic18-GAT, Ruiz20-Nonlinear, Ruiz20-GRNN}. Extension of nonlinearities to encompass neighborhood information is proposed in \cite{Ruiz20-Nonlinear}, {and architectures with residual connections are proposed in \cite{Ioannidis19-Recurrent,bresson2017residual}.} Edge-varying filters \cite{Coutino19-Distributed} can be used to design edge-varying GNNs \cite{Isufi20-EdgeNets} and graph attention networks \cite{Velickovic18-GAT, hacene2019attention}. {Multi-hop attention based GNNs are introduced in \cite{wang2020direct}.} {Architectures considering multi-relational data, i.e., data with support on multiple graphs or graphs with multidimensional edge features, have been proposed in \cite{Ioannidis19-Recurrent, gong2019exploiting}, and} architectures that leverage time dependencies are available in the form of graph recurrent neural networks \cite{Li18-DiffusionRNN, Seo18-GCRN, Ruiz20-GRNN}. We point out that these architectures are different from the GNNs based on graph filters that are described in this paper. To stress this point, GNNs that rely on graph convolutional filters are sometimes called graph convolutional neural networks.

Results on permutation equivariance and stability that we present here are drawn from \cite{Gama19-Stability} and results on transferability are drawn from \cite{ruiz2020wnn}. Other important works on stability of GNNs appear in the context of graph scattering transforms \cite{ZouLerman19-Scattering, Gama19-DiffusionScattering}. {Permutation equivariance is simple to prove, but has nevertheless drawn considerable attention because of its practical importance \cite{ZouLerman19-Scattering, Gama19-DiffusionScattering, Xu19-GIN, Villar19-EquivIsomorphism,frossard2020gnns}.} Our transferability analysis builds upon the concept of graphons and convergent graph sequences \cite{lovasz2012large, borgs2012convergent} which have proven insightful when processing graph data \cite{wolfe2013nonparametric, avella2018centrality, parise2019graphon}. In particular, GSP in the limit has given rise to the topic of graphon signal processing \cite{ruiz2020graphon, morency2017signal, ruiz2020wnn}. {An alternative transferability analysis relying on generic topological spaces such as manifolds where graph Laplacians are sampled from Laplace-Beltrami operators is also possible \cite{levie2019transferability}.}

{Throughout this paper, we use recommendation systems as a running example to illustrate ideas \cite{Huang18-RatingGSP, Ying18-LargeScaleGNN, Monti17-Movie} and, in Secs. \ref{sec_dcis}--\ref{sbs:transf}, present numerical results that illustrate GNN stability in decentralized robot control and GNN transferability in wireless resource allocation. The first two are examples of supervised learning problems, while the latter is an example of unsupervised learning. These are only some of the problems to which GNNs have been applied succesfully; others include identifying brain disorders \cite{ktena2017distance}, learning molecule fingerprints \cite{duvenaud2015convolutional}, web page ranking \cite{1517930}, text categorization \cite{Defferrard17-ChebNets, Gama19-Architectures} and clustering of citation networks \cite{Kipf17-GCN, Velickovic18-GAT, frossard2020choice}. Of particular interest to the Electrical Engineering community are applications to cyberphysical systems such as power grids \cite{Owerko20-OPF}, decentralized collaborative control of multiagent robotic systems \cite{Tolstaya19-Flocking, Li20-Planning} and wireless communication networks \cite{Eisen20-REGNN}.}

%We do not discuss applications here except for recommendation systems \cite{Huang18-RatingGSP, Ying18-LargeScaleGNN, Monti17-Movie}, which we use to illustrate ideas, but applications of GNNs abound. Some other problems where GNNs have been applied successfully are text categorization \cite{Defferrard17-ChebNets, Gama19-Architectures} and clustering of citation networks \cite{Kipf17-GCN, Velickovic18-GAT, frossard2020choice}. Of particular interest to the Electrical Engineering community are applications to cyberphysical systems such as power grids \cite{Owerko20-OPF}, decentralized collaborative control of multiagent robotic systems \cite{Tolstaya19-Flocking, Li20-Planning} and wireless communication networks \cite{Eisen20-REGNN}.

%% file: ch8_symbols.tex
\def\T{\mathsf{T}}
\def\Tr{T}
\def\Hr{H}

\def\nv{\textrm{nv}}
\def\hv{\textrm{hv}}
\def\dc{\text{dc}}

\newcommand*\mycirc[1]{%
	\begin{tikzpicture}
	\vspace{.4cm}
	\node[draw,circle,inner sep=1pt] {#1};
	\end{tikzpicture}}

%% file: ch8_sec01_classification.tex
% !TEX root = root.tex

\def \bbPhi {\Phi}

%%%%%%%%%%%%%%%%%%%%%%%%%%%%%%%%%%%%%%%%%%%%%%%%%%%%%%%%%%%%%%%%%%%%%%%%%%%%%%%%
%   F   I   G   U   R   E   %%%%%%%%%%%%%%%%%%%%%%%%%%%%%%%%%%%%%%%%%%%%%%%%%%%%
%%%%%%%%%%%%%%%%%%%%%%%%%%%%%%%%%%%%%%%%%%%%%%%%%%%%%%%%%%%%%%%%%%%%%%%%%%%%%%%%
%
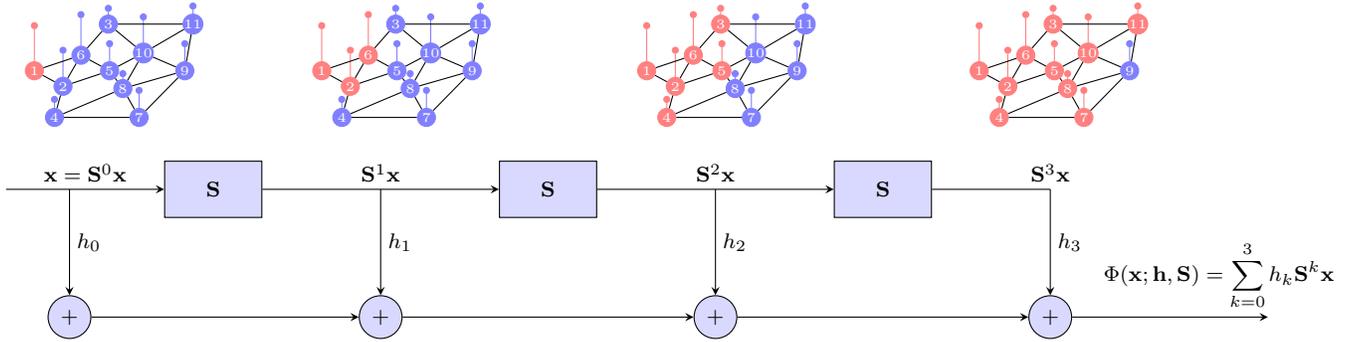
\begin{figure*}[t]
\centering
%\includegraphics [width = 0.245\linewidth]
%                 {plots/geo_graph_aggregation_neihgborhood_0.pdf}
%\includegraphics [width = 0.245\linewidth]
%                 {plots/geo_graph_aggregation_neihgborhood_1.pdf}
%\includegraphics [width = 0.245\linewidth]
%                 {plots/geo_graph_aggregation_neihgborhood_2.pdf}
%\includegraphics [width = 0.245\linewidth]
%                 {plots/geo_graph_aggregation_neihgborhood_3.pdf} \\ \bigskip
\input{figures/my_tikz_defs.tex}
\input{figures/another_graph_0.tex}\hspace{12mm}
\input{figures/another_graph_1.tex}\hspace{17mm}
\input{figures/another_graph_2.tex}\hspace{18mm}
\input{figures/another_graph_3.tex}\hspace{20mm}
\vspace{4mm}
\input{figures/time_convolution_3_with_graph.tex}
\caption{A graph convolutional filter is a polynomial on a matrix representation of the graph $\bbS$. We think of them as operations that propagate information through adjacent nodes. As the order of the filter grows, we aggregate information from nodes that are farther apart. But the integration of this information is always mediated by the neighborhood structure of the graph.} 
\label{fig.evRecMain}
\end{figure*}

%%%%%%%%%%%%%%%%%%%%%%%%%%%%%%%%%%%%%%%%%%%%%%%%%%%%%%%%%%%%%%%%%%%%%%%%%%%%%%%%
%   S   E   C   T   I   O   N   %%%%%%%%%%%%%%%%%%%%%%%%%%%%%%%%%%%%%%%%%%%%%%%%
%%%%%%%%%%%%%%%%%%%%%%%%%%%%%%%%%%%%%%%%%%%%%%%%%%%%%%%%%%%%%%%%%%%%%%%%%%%%%%%%
%
\section{Machine Learning on Graphs}\label{sec_ch8_learning}

Consider a graph $\bbG$ composed of vertices $V=\{1,\ldots n\}$, edges $E$ defined as ordered pairs $(i,j)$ and weights $w_{ij}$ associated with the edges. Our interest in this paper is on machine learning problems defined over this graph. Namely, we are given pairs $(\bbx, \bby)$ composed of an input graph signal $\bbx\in\reals^n$ and a target output graph signal $\bby\in\reals^n$. That $\bbx$ and $\bby$ are graph signals means that the components $x_i$ and $y_i$ are associated with the $i$th node of the graph. The pair $(\bbx, \bby)$ is jointly drawn from a probability distribution $p(\bbx, \bby)$ and our goal is to find a function $\Phi:\reals^n \to \reals^n$ such that $\Phi(\bbx)$ approximates $\bby$ over the probability distribution $p(\bbx, \bby)$. To do so we introduce the nonnegative loss function $\ell \big(\Phi(\bbx), \bby \big)\geq0$ such that $\ell \big(\Phi(\bbx), \bby \big)=0$ when $\Phi(\bbx)=\bby$ in order to measure the dissimilarity between the output $\Phi(\bbx)$ and the target output $\bby$. We can now define the function $\Phi^\dagger$ that best approximates $\bby$ as the one that minimizes the loss $\ell \big(\Phi(\bbx), \bby \big)$ averaged over the probability distribution $p(\bbx, \bby)$,
\begin{align}\label{eqn_ch8 _qtatistical_loss_minimization}
   \Phi^\dagger 
     \!= \argmin_{\Phi} \mbE\Big[\ell \big(\Phi(\bbx), \bby \big) \!\Big] 
     \!= \argmin_{\Phi} \!\int\! \ell \big(\Phi(\bbx), \bby \big) \, dp(\bbx, \bby).
\end{align} 
The expectation $\mbE[\ell (\Phi(\bbx), \bby ) ]$ is said to be a statistical loss and \eqref{eqn_ch8 _qtatistical_loss_minimization} is termed a statistical loss minimization problem.

A critical condition to solve \eqref{eqn_ch8 _qtatistical_loss_minimization} is availability of the probability distribution $p(\bbx, \bby)$. If this is known, the solution to \eqref{eqn_ch8 _qtatistical_loss_minimization} is to compute a posterior distribution that depends on the form of the loss function $\ell \big(\Phi(\bbx), \bby \big)$. The whole idea of machine learning, though, is that $p(\bbx, \bby)$ is not known. Instead, we have access to a collection of $Q$ data samples $(\bbx_q, \bby_q)$ drawn from the distribution $p(\bbx, \bby)$ which we group in the training set $\ccalT:= \{(\bbx_q, \bby_q) \big\}_{q=1}^{Q}$. Assuming these samples are acquired independently and that the number of samples $Q$ is large, a good approximation to the statistical loss in \eqref{eqn_ch8 _qtatistical_loss_minimization} is the empirical average $\bar\ell (\Phi) : =  (1/Q) \sum_{q=1}^{Q} \ell \big(\Phi(\bbx_q), \bby_q \big)$. Therefore, it is sensible to change our objective to search for a function $\Phi^{\ast}$ that minimizes the empirical average $\bar\ell (\Phi)$, 
\begin{align}\label{eqn_ch8_empirical_risk_minimization}
   \Phi^{\ast}   =  \argmin_{\Phi} \frac{1}{Q} \sum_{q=1}^{Q} \ell \big(\Phi(\bbx_q), \bby_q \big) .
%      :=  \argmin_{\Phi}\, \bar\ell (\Phi) .
\end{align} 
We say that \eqref{eqn_ch8_empirical_risk_minimization} is an empirical risk minimization (ERM) problem. The function $\Phi^{\ast}$ is the optimal {\it empirical} function associated with the training set $\ccalT$. 

%%%%%%%%%%%%%%%%%%%%%%%%%%%%%%%%%%%%%%%%%%%%%%%%%%%%%%%%%%%%%%%%%%%%%%%%%%%%%%%%
%   S   E   C   T   I   O   N   %%%%%%%%%%%%%%%%%%%%%%%%%%%%%%%%%%%%%%%%%%%%%%%%
%%%%%%%%%%%%%%%%%%%%%%%%%%%%%%%%%%%%%%%%%%%%%%%%%%%%%%%%%%%%%%%%%%%%%%%%%%%%%%%%
%
\subsection{Learning Parametrizations} \label{sec_parametrizations}

Observe that the solution to \eqref{eqn_ch8_empirical_risk_minimization} is elementary. Since $\ell \big(\Phi(\bbx), \bby \big)=0$ when $\Phi(\bbx)=\bby$ and nonnegative otherwise, it suffices to make $\Phi(\bbx _q)=\bby _q$ for all the observed samples $\bbx _q$ -- or some sort of average if the same input $\bbx_q$ is observed several times. 
%This is as elementary as it is nonsensical. In fact, 
{However, \eqref{eqn_ch8_empirical_risk_minimization} only makes sense as a problem formulation if we have access to all possible samples $\bbx _q$.} But the interest in practice is to infer, or {\it to learn,} the value of $\bby$ for samples $\bbx$ that have not been observed before.

This motivates the introduction of a learning parametrization $\ccalH$ that restricts the family of functions $\Phi$ that are admissible in \eqref{eqn_ch8_empirical_risk_minimization}. Thus, instead of searching over all $\Phi(\bbx)$ we search over functions $\Phi(\bbx; \ccalH)$ so that the ERM problem in \eqref{eqn_ch8_empirical_risk_minimization} is replaced by the alternative ERM formulation,
\begin{align}\label{eqn_ch8_erm_learning}
   \ccalH^*   =  \argmin_{\ccalH} \frac{1}{Q} \sum_{q=1}^{Q} \ell \big(\Phi(\bbx _q;\ccalH), \bby _q \big) .
         %:=  \argmin_{\ccalH}\, \bar\ell \big(\Phi(\xdot; \ccalH) \big) .
\end{align} 
A particular choice of parametrization is the set of linear functions of the form $\Phi(\bbx;\bbH) = \bbH \bbx$, in which case \eqref{eqn_ch8_empirical_risk_minimization} becomes 
\begin{align}\label{eqn_ch8_erm_linear}
   \bbH^*   =  \argmin_{\bbH} \frac{1}{Q} \sum_{q=1}^{Q} \ell \big(\bbH\bbx _q, \bby _q \big) .
         %:=  \argmin_{\bbH}\, \bar\ell \big(\Phi(\xdot; \bbH) \big) .
\end{align} 
Alternatively, one could choose $\Phi(\bbx; \ccalH)$ to be a neural network, or, as we will advocate in Sec. \ref{sec_ch8_gnns}, a graph filter or a GNN. The important point to highlight here is that the design of a machine learning system is tantamount to the selection of the proper learning parametrization. This is because in \eqref{eqn_ch8_erm_learning} the only choice left for a system designer is the class of functions $\Phi(\bbx;\ccalH)$ spanned by different choices of $\ccalH$. But, more importantly, this is also because the choice of parametrization determines how the function $\Phi(\bbx;\ccalH)$ generalizes from (observed) samples in the training set $(\bbx_q,\bby_q)\in\ccalT$ to unobserved signals $\bbx$.

%%%%%%%%%%%%%%%%%%%%%%%%%%%%%%%%%%%%%%%%%%%%%%%%%%%%%%%
%   S   E   C   T   I   O   N   %%%%%%%%%%%%%%%%%%%%%%%
%%%%%%%%%%%%%%%%%%%%%%%%%%%%%%%%%%%%%%%%%%%%%%%%%%%%%%%
%
\subsection{Recommendation Systems}\label{sec_reco_systems}

An example of ERM problem involving graph signals is a collaborative filtering approach to recommendation systems \cite{Huang18-RatingGSP}. In a recommendation system, we want to predict the ratings that customers would give to a certain product using rating histories. Collaborative filtering solutions build a graph of product similarities using past ratings and look at the ratings of each customer as a graph signal supported on the nodes of the product graph.  

\medskip\noindent \textbf{Product similarity graph.} Denote by $x_{ci}$ the rating that customer $c$ gives to product $i$. Typically, product $i$ has been rated by a subset of customers which we denote $\ccalC_i$. We consider the sets of users $\ccalC_{ij}=\ccalC_{i} \cap \ccalC_{j}$ that have rated products $i$ and $j$ and compute correlations
\begin{align}\label{eqn_reco_systems_weights}
   \sigma_{ij} = \frac{1}{|\ccalC_{ij}|}
                 \sum_{ c \in \ccalC_{ij}} 
                        (x_{ci}-\mu_{ij})(x_{cj}-\mu_{ji}),
\end{align} 
where we use the average ratings $\mu_{ij}=(1/|\ccalC_{ij}|)\sum_{c\in \ccalC_{ij}}x_{ci}$ and $\mu_{ji}=(1/|\ccalC_{ij}|)\sum_{c\in \ccalC_{ij}}x_{cj}$. The product graph used in collaborative filtering is the one with normalized weights
\begin{align}\label{eqn_reco_systems_graph_weights}
   w_{ij} \ = \ \sigma_{ij} \,\Big/\, \sqrt{\sigma_{ii}\sigma_{jj}}\ .
\end{align} 
A cartoon illustration of the product graph is shown in Fig. \ref{fig_generalization}-(a). Nodes represent different products, edges stand in for product similarity, and signal components are the product ratings of a given customer. As is typical in practice, a small number of products have been rated. 

\medskip\noindent \textbf{Training set.} To build a training set for this problem define the vector $\bbx_c=[x_{c1};\ldots x_{cn}]$ where $x_{ci}$ is the rating that user $c$ gave to product $i$, if available, or $x_{ci}=0$ otherwise. Further denote as $\ccalI_c$ the set of items rated by customer $c$. Let $i\in\ccalI_c$ be a product rated by customer $c$ and define the sparse vector $\bby_{ci}$ whose unique nonzero entry is $[\bby_{ci}]_i = x_{ci}$. With these definitions we construct the training set
\begin{align}\label{eqn_ch8_reco_system_training_set}
   \ccalT = \bigcup_{c, i \in\ccalI_c} 
                 \big \{ (\bbx_{ci}, \bby_{ci}) \,:\, 
                              \bbx_{ci} = \bbx_c - \bby_{ci} \big\}.
\end{align}
The process of building an input-output pair of the training set is illustrated in Fig. \ref{fig_generalization}-(b). In this particular example we isolate the rating that this customer gave to product $i=3$. This rating is recorded into a graph signal with a single nonzero entry $[\bby_{c3}]_3 = x_{c3}$. The remaining nonzero entries define the rating input $\bbx_{c3} = \bbx_c - \bby_{c3}$. This process is repeated for all the products in the set $i\in\ccalI_c$ of rated items of costumer $c$ and for all customers $c$.

\medskip\noindent \textbf{Loss function.} Our goal is to learn a map that will produce outputs $\bby_{ci}$ when presented with inputs $\bbx_{ci}$. E.g., in the case of Fig. \ref{fig_generalization} we want to present Fig. \ref{fig_generalization}-(b) as an input and fill in a rating of product $i=3$ equal to the rating of product $i=3$ in Fig. \ref{fig_generalization}-(a). To do that we define the loss function
\begin{align}\label{eqn_ch8_reco_system_training_set_arbitrary_linear}
   \ell \big(\Phi(\bbx_{ci}; \ccalH), \bby_{ci} \big) 
      \ = \ \frac{1}{2}
               \Big(\bbe_i^T \Phi(\bbx_{ci}; \ccalH) 
                            - \bbe_i^T\bby_{ci} \Big)^2,
\end{align}
where the vector $\bbe_i$ is the ith entry of the canonical basis of $\reals^n$. 
Since multiplying with $\bbe_i^T$ extracts the $i$th component of a vector, the loss in \eqref{eqn_ch8_reco_system_training_set_arbitrary_linear} compares the predicted rating $\bbe_i^T \Phi(\bbx_{ci}; \ccalH) = [\Phi(\bbx_{ci}; \ccalH)]_i$ with the observed rating $\bbe_i^T\bby_{ci}=[\bby_{ci}]_i = x_{ci}$. At execution time, this map can be used to predict ratings of unrated products from the ratings of rated products. If we encounter the signal in Fig. \ref{fig_generalization}-(b) we know the prediction will be accurate because we encountered this signal during training. If we are given the signals in Fig. \ref{fig_generalization}-(c) or Fig. \ref{fig_generalization}-(d) successful rating predictions depend on the choice of parametrization.

%\begin{remark} \normalfont We point out that we could also build graphs of customer similarities and consider signals for individual products. It is also possible to look at combinations of both graphs and to perform adaptations to individual customers or products. These techniques improve results as demonstrated in \cite{Huang18-RatingGSP}, however, our interest here is not on attaining state-of-the art results, but on using recommendation systems to illustrate ideas. \end{remark}

%% file: figures/my_tikz_defs.tex
% This is a global control on the scale of plots. The constant \myfactor can be defined inside of an individual picture. It is good practice to reset \myfactor to 1 after the end of the picture.
\def\myfactor{1.0}
\def\unit{\myfactor cm}

% This command sets the appearance of axis on pgfplots
\pgfplotsset{ ylabel near ticks,                 % Make x axis label appear close to figure
              xlabel near ticks,                 % Make y axis label appear close to figure
              tick label style = {font=\footnotesize},   % Write numbers in tiny font
              label style = {font=\footnotesize}, % Write axes labels in footnote size font
              title style = {font=\footnotesize},
            }

% Definition of a graph node without filled in color. 
% Controls the appearance of a node in a graph plot
\tikzstyle{empty node} = [ circle, 
                     draw = black,
                     text = black, 
                     minimum size = 0.8*\unit]

% Definition of a graph node. Controls the appearance of a node in a graph plot
\tikzstyle{node} = [ empty node, 
                     fill = blue!50,
                     draw = blue!50,
                     text = white]

\tikzstyle{blue node} = [ empty node, 
                         fill = blue!50,
                         draw = blue!50,
                         text = white]

\tikzstyle{red node} = [ empty node, 
                         fill = red!50,
                         draw = red!50,
                         text = white]

\tikzstyle{green node} = [ empty node, 
                         fill = mygreen!50,
                         draw = mygreen!50,
                         text = white]

\tikzstyle{black node} = [ empty node, 
                           fill = black!40,
                           draw = black!40,
                           text = white]

% Definition of a graph signal dot value without filled in color.
% These are smaller than the nodes and do not allow for text.
% Controls the appearance of a graph signal dot value in a graph plot
\tikzstyle{empty dot} = [ circle, 
                           draw = black,
                           inner sep = 0pt,
                           anchor = center,
                           minimum size = 0.4*\unit]

% Definition of a graph signal dot value.
% Controls the appearance of a graph signal dot value in a graph plot
\tikzstyle{dot} = [ empty dot, 
                    fill = blue!50,
                    draw = blue!50]

\tikzstyle{blue dot} = [ empty dot, 
                         fill = blue!50,
                         draw = blue!50]

\tikzstyle{red dot} = [ empty dot, 
                        fill = red!50,
                        draw = red!50]

\tikzstyle{black dot} = [ empty dot, 
                          fill = black!50,
                          draw = black!50]

% Definition of graph edges. An edge is undirected and doesn't contain arrows. A directed edge has an arrow at the end and a double directed edge contains two arrows. The latter is to be used when we want to emphasize the double directivity.
\tikzstyle{edge}                 = [shorten >=1pt, shorten <=1pt]
\tikzstyle{directed edge}        = [edge, -stealth]
\tikzstyle{double directed edge} = [edge, stealth-stealth]
\tikzstyle{tight edge}                 = [shorten >=0pt, shorten <=0pt]

% Definition of a block for block diagrams
\tikzstyle{block} = [ rectangle,
                      minimum width = \unit,
                      minimum height = \unit,
                      fill = blue!15,
                      draw = black,
                      text = black]

%% file: figures/another_graph_0.tex
%!TEX root = ../root.tex

\def \thisplotscale {0.2}
\def \unit {\thisplotscale cm}

\def\ypos{1.478}
\def\xpos{1.25}
\def\lbl{1.25}
\def\tanAngle{0.577} % tan(pi/6) = 0.577; tan(pi/3) = 1.732; tan (pi/4) = 1
\def\signalLength{3}

\tikzstyle{small blue node} = [ blue node, 
                                inner sep = 0,
                                minimum size = 1.2*\unit]
                                
\tikzstyle{small red node} = [ red node, 
                               inner sep = 0,
                               minimum size = 1.2*\unit]
                               
\tikzstyle{overlay edge} = [tight edge,
                            shorten >= 3]

\def \radius   {3}

{\tiny
\begin{tikzpicture}[scale = \thisplotscale]
    
    % 11 width x 7 height
    
    % Anchor nodes
    %   Origin
    \node at (0,0) (start) {};
    
    \path (start) ++ (0*\xpos + 0*\ypos*\tanAngle, 0*\ypos) node [small red node] (1) {$1$};
 
    \path (start) ++ (2.5*\xpos - 2.1*\ypos*\tanAngle, -2.1*\ypos) node [small blue node] (4) {$4$};   
    \path (start) ++ (2*\xpos - 0.7*\ypos*\tanAngle, -0.7*\ypos) node [small blue node] (2) {$2$};
    \path (start) ++ (2*\xpos+ 0.7*\ypos*\tanAngle, 0.7*\ypos) node [small blue node] (6) {$6$};
    \path (start) ++ (2.5*\xpos+ 2.1*\ypos*\tanAngle, 2.1*\ypos) node [small blue node] (3) {$3$};
    
    \path (start) ++ (4*\xpos + 0*\ypos*\tanAngle, 0*\ypos) node [small blue node] (5) {$5$};   
    \path (start) ++ (5.25*\xpos - 0.8*\ypos*\tanAngle, -0.8*\ypos) node [small blue node] (8) {$8$};
    \path (start) ++ (5.25*\xpos+ 0.8*\ypos*\tanAngle, 0.8*\ypos) node [small blue node] (10) {$10$};
    
    \path (start) ++ (7*\xpos - 2.1*\ypos*\tanAngle, -2.1*\ypos) node [small blue node] (7) {$7$};
    \path (start) ++ (8*\xpos + 0*\ypos*\tanAngle, 0*\ypos) node [small blue node] (9) {$9$};
    \path (start) ++ (7*\xpos + 2.1*\ypos*\tanAngle, 2.1*\ypos) node [small blue node] (11) {$11$};
    
    % Signals
    \path(1) ++ (0, \signalLength) node [red dot] (x1) {};
    \path (x1) edge[overlay edge,  draw = red!50] (x1|-1);
    
    \path(2) ++ (0, 0.8*\signalLength) node [blue dot] (x2) {};
    \path (x2) edge[overlay edge,  draw = blue!50] (x2|-2);
    
    \path(3) ++ (0, 0.35*\signalLength) node [blue dot] (x3) {};
    \path (x3) edge[overlay edge,  draw = blue!50] (x3|-3);
    
    \path(4) ++ (0, 0.4*\signalLength) node [blue dot] (x4) {};
    \path (x4) edge[overlay edge,  draw = blue!50] (x4|-4);
    
    \path(5) ++ (0, 0.6*\signalLength) node [blue dot] (x5) {};
    \path (x5) edge[overlay edge,  draw = blue!50] (x5|-5);
    
    \path(6) ++ (0, 0.9*\signalLength) node [blue dot] (x6) {};
    \path (x6) edge[overlay edge,  draw = blue!50] (x6|-6);
    
    \path(7) ++ (0, 0.6*\signalLength) node [blue dot] (x7) {};
    \path (x7) edge[overlay edge,  draw = blue!50] (x7|-7);
    
    \path(8) ++ (0, 0.33*\signalLength) node [blue dot] (x8) {};
    \path (x8) edge[overlay edge,  draw = blue!50] (x8|-8);

    \path(9) ++ (0, 0.6*\signalLength) node [blue dot] (x9) {};
    \path (x9) edge[overlay edge,  draw = blue!50] (x9|-9);
    
    \path(10) ++ (0, 0.8*\signalLength) node [blue dot] (x10) {};
    \path (x10) edge[overlay edge,  draw = blue!50] (x10|-10);
    
    \path(11) ++ (0, 0.4*\signalLength) node [blue dot] (x11) {};
    \path (x11) edge[overlay edge,  draw = blue!50] (x11|-11);

    % Edges connecting left hexagon  
    \path (1)  edge [tight edge] node {} (2);
    \path (1)  edge [tight edge] node {} (6);
    
    \path (2) edge [tight edge] node {} (6);
    \path (2) edge [tight edge] node {} (4);
    \path (2) edge [tight edge] node {} (5);
    \path (6) edge [tight edge] node {} (5);
    \path (6) edge [tight edge] node {} (3);
    
    \path (5) edge [tight edge] node {} (8);
    \path (5) edge [tight edge] node {} (10);
    
    \path (3) edge [tight edge] node {} (10);
    \path (3) edge [tight edge] node {} (11);
    
    \path (4) edge [tight edge] node {} (8);
    \path (4) edge [tight edge] node {} (7);
    
    \path (10) edge [tight edge] node {} (11);
    \path (10) edge [tight edge] node {} (8);
    \path (10) edge [tight edge] node {} (9);
    
    \path (8) edge [tight edge] node {} (7);
    \path (8) edge [tight edge] node {} (9);
    
    \path (9) edge [tight edge] node {} (11);
    \path (9) edge [tight edge] node {} (7);

\end{tikzpicture}} 

%% file: figures/another_graph_1.tex
{\tiny
\begin{tikzpicture}[scale = \thisplotscale]
    
    % 11 width x 7 height
    
    % Anchor nodes
    %   Origin
    \node at (0,0) (start) {};
    
    \path (start) ++ (0*\xpos + 0*\ypos*\tanAngle, 0*\ypos) node [small red node] (1) {$1$};
 
    \path (start) ++ (2.5*\xpos - 2.1*\ypos*\tanAngle, -2.1*\ypos) node [small blue node] (4) {$4$};   
    \path (start) ++ (2*\xpos - 0.7*\ypos*\tanAngle, -0.7*\ypos) node [small red node] (2) {$2$};
    \path (start) ++ (2*\xpos+ 0.7*\ypos*\tanAngle, 0.7*\ypos) node [small red node] (6) {$6$};
    \path (start) ++ (2.5*\xpos+ 2.1*\ypos*\tanAngle, 2.1*\ypos) node [small blue node] (3) {$3$};
    
    \path (start) ++ (4*\xpos + 0*\ypos*\tanAngle, 0*\ypos) node [small blue node] (5) {$5$};   
    \path (start) ++ (5.25*\xpos - 0.8*\ypos*\tanAngle, -0.8*\ypos) node [small blue node] (8) {$8$};
    \path (start) ++ (5.25*\xpos+ 0.8*\ypos*\tanAngle, 0.8*\ypos) node [small blue node] (10) {$10$};
    
    \path (start) ++ (7*\xpos - 2.1*\ypos*\tanAngle, -2.1*\ypos) node [small blue node] (7) {$7$};
    \path (start) ++ (8*\xpos + 0*\ypos*\tanAngle, 0*\ypos) node [small blue node] (9) {$9$};
    \path (start) ++ (7*\xpos + 2.1*\ypos*\tanAngle, 2.1*\ypos) node [small blue node] (11) {$11$};
    
    % Signals
    \path(1) ++ (0, \signalLength) node [red dot] (x1) {};
    \path (x1) edge[overlay edge,  draw = red!50] (x1|-1);
    
    \path(2) ++ (0, 0.8*\signalLength) node [red dot] (x2) {};
    \path (x2) edge[overlay edge,  draw = red!50] (x2|-2);
    
    \path(3) ++ (0, 0.35*\signalLength) node [blue dot] (x3) {};
    \path (x3) edge[overlay edge,  draw = blue!50] (x3|-3);
    
    \path(4) ++ (0, 0.4*\signalLength) node [blue dot] (x4) {};
    \path (x4) edge[overlay edge,  draw = blue!50] (x4|-4);
    
    \path(5) ++ (0, 0.6*\signalLength) node [blue dot] (x5) {};
    \path (x5) edge[overlay edge,  draw = blue!50] (x5|-5);
    
    \path(6) ++ (0, 0.9*\signalLength) node [red dot] (x6) {};
    \path (x6) edge[overlay edge,  draw = red!50] (x6|-6);
    
    \path(7) ++ (0, 0.6*\signalLength) node [blue dot] (x7) {};
    \path (x7) edge[overlay edge,  draw = blue!50] (x7|-7);
    
    \path(8) ++ (0, 0.33*\signalLength) node [blue dot] (x8) {};
    \path (x8) edge[overlay edge,  draw = blue!50] (x8|-8);
    
    \path(9) ++ (0, 0.6*\signalLength) node [blue dot] (x9) {};
    \path (x9) edge[overlay edge,  draw = blue!50] (x9|-9);
    
    \path(10) ++ (0, 0.8*\signalLength) node [blue dot] (x10) {};
    \path (x10) edge[overlay edge,  draw = blue!50] (x10|-10);
    
    \path(11) ++ (0, 0.4*\signalLength) node [blue dot] (x11) {};
    \path (x11) edge[overlay edge,  draw = blue!50] (x11|-11);

    % Edges connecting left hexagon  
    \path (1)  edge [tight edge] node {} (2);
    \path (1)  edge [tight edge] node {} (6);
    
    \path (2) edge [tight edge] node {} (6);
    \path (2) edge [tight edge] node {} (4);
    \path (2) edge [tight edge] node {} (5);
    \path (6) edge [tight edge] node {} (5);
    \path (6) edge [tight edge] node {} (3);
    
    \path (5) edge [tight edge] node {} (8);
    \path (5) edge [tight edge] node {} (10);
    
    \path (3) edge [tight edge] node {} (10);
    \path (3) edge [tight edge] node {} (11);
    
    \path (4) edge [tight edge] node {} (8);
    \path (4) edge [tight edge] node {} (7);
    
    \path (10) edge [tight edge] node {} (11);
    \path (10) edge [tight edge] node {} (8);
    \path (10) edge [tight edge] node {} (9);
    
    \path (8) edge [tight edge] node {} (7);
    \path (8) edge [tight edge] node {} (9);
    
    \path (9) edge [tight edge] node {} (11);
    \path (9) edge [tight edge] node {} (7);

\end{tikzpicture}} 

%% file: figures/another_graph_2.tex
{\tiny
\begin{tikzpicture}[scale = \thisplotscale]
    
    % 11 width x 7 height
    
    % Anchor nodes
    %   Origin
    \node at (0,0) (start) {};
    
    \path (start) ++ (0*\xpos + 0*\ypos*\tanAngle, 0*\ypos) node [small red node] (1) {$1$};
 
    \path (start) ++ (2.5*\xpos - 2.1*\ypos*\tanAngle, -2.1*\ypos) node [small red node] (4) {$4$};   
    \path (start) ++ (2*\xpos - 0.7*\ypos*\tanAngle, -0.7*\ypos) node [small red node] (2) {$2$};
    \path (start) ++ (2*\xpos+ 0.7*\ypos*\tanAngle, 0.7*\ypos) node [small red node] (6) {$6$};
    \path (start) ++ (2.5*\xpos+ 2.1*\ypos*\tanAngle, 2.1*\ypos) node [small red node] (3) {$3$};
    
    \path (start) ++ (4*\xpos + 0*\ypos*\tanAngle, 0*\ypos) node [small red node] (5) {$5$};   
    \path (start) ++ (5.25*\xpos - 0.8*\ypos*\tanAngle, -0.8*\ypos) node [small blue node] (8) {$8$};
    \path (start) ++ (5.25*\xpos+ 0.8*\ypos*\tanAngle, 0.8*\ypos) node [small blue node] (10) {$10$};
    
    \path (start) ++ (7*\xpos - 2.1*\ypos*\tanAngle, -2.1*\ypos) node [small blue node] (7) {$7$};
    \path (start) ++ (8*\xpos + 0*\ypos*\tanAngle, 0*\ypos) node [small blue node] (9) {$9$};
    \path (start) ++ (7*\xpos + 2.1*\ypos*\tanAngle, 2.1*\ypos) node [small blue node] (11) {$11$};
    
    % Signals
    \path(1) ++ (0, \signalLength) node [red dot] (x1) {};
    \path (x1) edge[overlay edge,  draw = red!50] (x1|-1);
    
    \path(2) ++ (0, 0.8*\signalLength) node [red dot] (x2) {};
    \path (x2) edge[overlay edge,  draw = red!50] (x2|-2);
    
    \path(3) ++ (0, 0.35*\signalLength) node [red dot] (x3) {};
    \path (x3) edge[overlay edge,  draw = red!50] (x3|-3);
    
    \path(4) ++ (0, 0.4*\signalLength) node [red dot] (x4) {};
    \path (x4) edge[overlay edge,  draw = red!50] (x4|-4);
    
    \path(5) ++ (0, 0.6*\signalLength) node [red dot] (x5) {};
    \path (x5) edge[overlay edge,  draw = red!50] (x5|-5);
    
    \path(6) ++ (0, 0.9*\signalLength) node [red dot] (x6) {};
    \path (x6) edge[overlay edge,  draw = red!50] (x6|-6);
    
    \path(7) ++ (0, 0.6*\signalLength) node [blue dot] (x7) {};
    \path (x7) edge[overlay edge,  draw = blue!50] (x7|-7);
    
    \path(8) ++ (0, 0.33*\signalLength) node [blue dot] (x8) {};
    \path (x8) edge[overlay edge,  draw = blue!50] (x8|-8);
    
    \path(9) ++ (0, 0.6*\signalLength) node [blue dot] (x9) {};
    \path (x9) edge[overlay edge,  draw = blue!50] (x9|-9);
    
    \path(10) ++ (0, 0.8*\signalLength) node [blue dot] (x10) {};
    \path (x10) edge[overlay edge,  draw = blue!50] (x10|-10);
    
    \path(11) ++ (0, 0.4*\signalLength) node [blue dot] (x11) {};
    \path (x11) edge[overlay edge,  draw = blue!50] (x11|-11);

    % Edges connecting left hexagon  
    \path (1)  edge [tight edge] node {} (2);
    \path (1)  edge [tight edge] node {} (6);
    
    \path (2) edge [tight edge] node {} (6);
    \path (2) edge [tight edge] node {} (4);
    \path (2) edge [tight edge] node {} (5);
    \path (6) edge [tight edge] node {} (5);
    \path (6) edge [tight edge] node {} (3);
    
    \path (5) edge [tight edge] node {} (8);
    \path (5) edge [tight edge] node {} (10);
    
    \path (3) edge [tight edge] node {} (10);
    \path (3) edge [tight edge] node {} (11);
    
    \path (4) edge [tight edge] node {} (8);
    \path (4) edge [tight edge] node {} (7);
    
    \path (10) edge [tight edge] node {} (11);
    \path (10) edge [tight edge] node {} (8);
    \path (10) edge [tight edge] node {} (9);
    
    \path (8) edge [tight edge] node {} (7);
    \path (8) edge [tight edge] node {} (9);
    
    \path (9) edge [tight edge] node {} (11);
    \path (9) edge [tight edge] node {} (7);

\end{tikzpicture}} 

%% file: figures/another_graph_3.tex
{\tiny
\begin{tikzpicture}[scale = \thisplotscale]
    
    % 11 width x 7 height
    
    % Anchor nodes
    %   Origin
    \node at (0,0) (start) {};
    
    \path (start) ++ (0*\xpos + 0*\ypos*\tanAngle, 0*\ypos) node [small red node] (1) {$1$};
 
    \path (start) ++ (2.5*\xpos - 2.1*\ypos*\tanAngle, -2.1*\ypos) node [small red node] (4) {$4$};   
    \path (start) ++ (2*\xpos - 0.7*\ypos*\tanAngle, -0.7*\ypos) node [small red node] (2) {$2$};
    \path (start) ++ (2*\xpos+ 0.7*\ypos*\tanAngle, 0.7*\ypos) node [small red node] (6) {$6$};
    \path (start) ++ (2.5*\xpos+ 2.1*\ypos*\tanAngle, 2.1*\ypos) node [small red node] (3) {$3$};
    
    \path (start) ++ (4*\xpos + 0*\ypos*\tanAngle, 0*\ypos) node [small red node] (5) {$5$};   
    \path (start) ++ (5.25*\xpos - 0.8*\ypos*\tanAngle, -0.8*\ypos) node [small red node] (8) {$8$};
    \path (start) ++ (5.25*\xpos+ 0.8*\ypos*\tanAngle, 0.8*\ypos) node [small red node] (10) {$10$};
    
    \path (start) ++ (7*\xpos - 2.1*\ypos*\tanAngle, -2.1*\ypos) node [small red node] (7) {$7$};
    \path (start) ++ (8*\xpos + 0*\ypos*\tanAngle, 0*\ypos) node [small blue node] (9) {$9$};
    \path (start) ++ (7*\xpos + 2.1*\ypos*\tanAngle, 2.1*\ypos) node [small red node] (11) {$11$};
    
    % Signals
    \path(1) ++ (0, \signalLength) node [red dot] (x1) {};
    \path (x1) edge[overlay edge,  draw = red!50] (x1|-1);
    
    \path(2) ++ (0, 0.8*\signalLength) node [red dot] (x2) {};
    \path (x2) edge[overlay edge,  draw = red!50] (x2|-2);
    
    \path(3) ++ (0, 0.35*\signalLength) node [red dot] (x3) {};
    \path (x3) edge[overlay edge,  draw = red!50] (x3|-3);
    
    \path(4) ++ (0, 0.4*\signalLength) node [red dot] (x4) {};
    \path (x4) edge[overlay edge,  draw = red!50] (x4|-4);
    
    \path(5) ++ (0, 0.6*\signalLength) node [red dot] (x5) {};
    \path (x5) edge[overlay edge,  draw = red!50] (x5|-5);
    
    \path(6) ++ (0, 0.9*\signalLength) node [red dot] (x6) {};
    \path (x6) edge[overlay edge,  draw = red!50] (x6|-6);
    
    \path(7) ++ (0, 0.6*\signalLength) node [red dot] (x7) {};
    \path (x7) edge[overlay edge,  draw = red!50] (x7|-7);
    
    \path(8) ++ (0, 0.33*\signalLength) node [red dot] (x8) {};
    \path (x8) edge[overlay edge,  draw = red!50] (x8|-8);
    
    \path(9) ++ (0, 0.6*\signalLength) node [blue dot] (x9) {};
    \path (x9) edge[overlay edge,  draw = blue!50] (x9|-9);
    
    \path(10) ++ (0, 0.8*\signalLength) node [red dot] (x10) {};
    \path (x10) edge[overlay edge,  draw = red!50] (x10|-10);
    
    \path(11) ++ (0, 0.4*\signalLength) node [red dot] (x11) {};
    \path (x11) edge[overlay edge,  draw = red!50] (x11|-11);

    % Edges connecting left hexagon  
    \path (1)  edge [tight edge] node {} (2);
    \path (1)  edge [tight edge] node {} (6);
    
    \path (2) edge [tight edge] node {} (6);
    \path (2) edge [tight edge] node {} (4);
    \path (2) edge [tight edge] node {} (5);
    \path (6) edge [tight edge] node {} (5);
    \path (6) edge [tight edge] node {} (3);
    
    \path (5) edge [tight edge] node {} (8);
    \path (5) edge [tight edge] node {} (10);
    
    \path (3) edge [tight edge] node {} (10);
    \path (3) edge [tight edge] node {} (11);
    
    \path (4) edge [tight edge] node {} (8);
    \path (4) edge [tight edge] node {} (7);
    
    \path (10) edge [tight edge] node {} (11);
    \path (10) edge [tight edge] node {} (8);
    \path (10) edge [tight edge] node {} (9);
    
    \path (8) edge [tight edge] node {} (7);
    \path (8) edge [tight edge] node {} (9);
    
    \path (9) edge [tight edge] node {} (11);
    \path (9) edge [tight edge] node {} (7);

\end{tikzpicture}} 

%% file: figures/time_convolution_3_with_graph.tex
%!TEX root = ../root.tex

\def \thisplotscale {1.7}
\def \unit {\thisplotscale cm}

\tikzstyle {Phi} = [block, 
thin,
minimum width = 0.76*\unit, 
minimum height = 0.44*\unit, 
anchor = west]

\tikzstyle {sum} = [circle, 
thin,
minimum width  = 0.26*\unit, 
minimum height = 0.26*\unit, 
anchor = center,
draw = black,
fill = blue!15]

\def \deltax   {1.85}
\def \deltay   {1.0}
\def \sumshift {0.5}

\def\shiftop      {${\bbS}$}
\def\signaln      {${\bbx = \bbS^0}\bbx$}
\def\signalnmone  {${\bbS^1}\bbx$}
\def\signalnmtwo  {${\bbS^2}\bbx$}
\def\signalnmthree{${\bbS^3}\bbx$}
\def\signaloutput {$ \displaystyle{ \Phi (\bbx; \bbh, \bbS)
                                         =\sum_{k=0}^{3} h_k \bbS^k  \bbx }$}

{\footnotesize
\begin{tikzpicture}[x = 1*\unit, y = 1*\unit]

% Begin by drawing an empty node for initializing the chain
\node (origin) [] {};
\path (origin) ++ (0, 0) node (first) [] {};

% Draw the chain of graph shifts
\path (origin) ++ (0.7*\deltax, 0) node (0) [Phi] {\shiftop};
\path (0.east) ++ (\deltax, 0) node (1) [Phi] {\shiftop};
\path (1.east) ++ (\deltax, 0) node (2) [Phi] {\shiftop};

% End by drawing an empty node for finalizing the chain
\path (2.east) ++ (\sumshift*\deltax, 0) node [anchor=west] (last) [] {};

% Draw the chain of additions
\path (origin) ++ (0.6*\sumshift*\deltax, -\deltay) node (sum0) [sum] {$+$};
\path (0.east) ++ (\sumshift*\deltax, -\deltay) node (sum1) [sum] {$+$};
\path (1.east) ++ (\sumshift*\deltax, -\deltay) node (sum2) [sum] {$+$};
\path (2.east) ++ (\sumshift*\deltax, -\deltay) node (sum3) [sum] {$+$};

% Draw the arrows connecting the shift blocks
\path[-stealth] (first) edge [above]         node {\signaln}       (0);	
\path[-stealth] (0)     edge [above]         node {\signalnmone}   (1);	
\path[-stealth] (1)     edge [above]         node {\signalnmtwo}   (2);	
\path[-]        (2)     edge [above, at end] node {\signalnmthree} (sum3|-last);

% Draw the arrows connecting to the sum blocks
\path[-stealth, draw] (sum0 |- first) -- (sum0) node [midway, right] {${h_{0}}$};	
\path[-stealth, draw] (sum1 |- 0)     -- (sum1) node [midway, right] {${h_{1}}$};	
\path[-stealth, draw] (sum2 |- 1)     -- (sum2) node [midway, right] {${h_{2}}$};	
\path[-stealth, draw] (sum3 |- 2)     -- (sum3) node [midway, right] {${h_{3}}$};

% Draw the arrows connecting the sum blocks
\path[-stealth, draw] (sum0) -- (sum1);	
\path[-stealth, draw] (sum1) -- (sum2);	
\path[-stealth, draw] (sum2) -- (sum3);	

% Draw the convolution output
\path[-stealth] (sum3) edge [above, near end] 
                node {\signaloutput} ++ (0.92*\deltax, 0);

\end{tikzpicture}}

%% file: ch8_sec03_gnns.tex
% !TEX root = root.tex

%%%%%%%%%%%%%%%%%%%%%%%%%%%%%%%%%%%%%%%%%%%%%%%%%%%%%%%%%%%%%%%%%%%%%%%%%%%%%%%%
%   F   I   G   U   R   E   %%%%%%%%%%%%%%%%%%%%%%%%%%%%%%%%%%%%%%%%%%%%%%%%%%%%
%%%%%%%%%%%%%%%%%%%%%%%%%%%%%%%%%%%%%%%%%%%%%%%%%%%%%%%%%%%%%%%%%%%%%%%%%%%%%%%%
%
\begin{figure*}[t] 
\def\myfactor{0.57}
\centering
\input{figures/hexagon_graph_signal_node_3_a}\hspace{-6mm}
\input{figures/hexagon_graph_signal_node_3_b}\hspace{-6mm}
\input{figures/hexagon_graph_signal_node_3_c}\hspace{-6mm}
\input{figures/hexagon_graph_signal_node_6}  
\hspace{11mm}
{\bf \footnotesize(a)}\hspace{42mm}
{\bf \footnotesize(b)}\hspace{42mm}
{\bf \footnotesize(c)}\hspace{42mm}
{\bf \footnotesize(d)}
\caption{The graph represents product similarity in a recommendation system. If we are given samples (a) for training, any reasonable parametrization learns to complete the rating of node 3 when observing the signal in (b). The linear parametrization in \eqref{eqn_ch8_erm_linear} also learns to fill the rating of node 3 when observing (c) -- node saturation is proportional to signal value. The graph filter parametrization in \eqref{eqn_ch8_erm_graph_filter} generalizes to (c) but it also generalizes to predicting the rating of node 6 in (d). This is true because of the permutation equivariance result in Proposition \ref{prop_filter_equivariance}. Graph neural networks [cf. \eqref{eqn_ch8_gnn_recursion_filter_matrix}-\eqref{eqn_ch8_gnn_operator_multifeature}] inherit this generalization property (Proposition~\ref{prop_gnn_equivariance}).}
\label{fig_generalization}
\end{figure*}
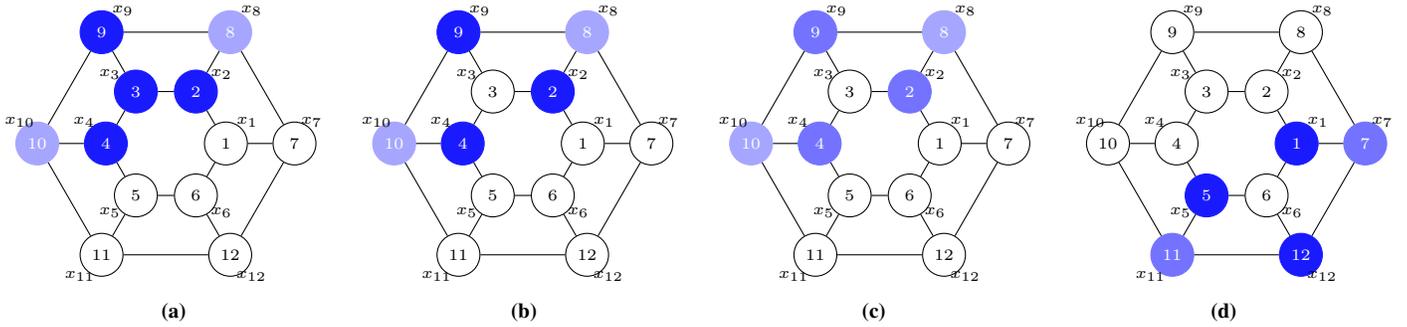

%%%%%%%%%%%%%%%%%%%%%%%%%%%%%%%%%%%%%%%%%%%%%%%%%%%%%%%%%%%%%%%%%%%%%%%%%%%%%%%%
%   S   E   C   T   I   O   N   %%%%%%%%%%%%%%%%%%%%%%%%%%%%%%%%%%%%%%%%%%%%%%%%
%%%%%%%%%%%%%%%%%%%%%%%%%%%%%%%%%%%%%%%%%%%%%%%%%%%%%%%%%%%%%%%%%%%%%%%%%%%%%%%%
%
\section{Graph Neural Networks}\label{sec_ch8_gnns}

As we explained in Sec. \ref{sec_parametrizations}, the choice of parametrization determines the manner in which the function $\Phi(\bbx; \ccalH)$ generalizes from elements of the training set to unobserved samples. A parametrization that is convenient for processing graph signals is a graph convolutional filter \cite{Sandryhaila13-DSPG, Shuman13-SPG, Segarra17-Linear, Isufi17-ARMA}. To define this operation let $\bbS \in \reals^{n \times n}$ denote a matrix representation of the graph and introduce a filter order $K$ along with filter coefficients $h_k$ that we group in the vector $\bbh=[h_0;\ldots; h_K]$. A graph convolutional filter applied to the graph signal $\bbx$ is a polynomial on this matrix representation,
\begin{equation} \label{eqn_filter}
   \bbu \ = \ \sum_{k=0}^{K} h_k \bbS^k \, \bbx
        \ = \ \Phi (\bbx; \bbh, \bbS),
\end{equation}
where we have defined $\Phi (\bbx; \bbh, \bbS)$ in the second equality to represent the output of a graph filter with coefficients $\bbh$ run on the matrix representation $\bbS$ and applied to the graph signal $\bbx$. The output $\bbu = \Phi (\bbx; \bbh, \bbS)$ is also a graph signal. In the context of \eqref{eqn_filter}, the representation $\bbS$ is termed a graph shift operator. If we need to fix ideas we will interpret $\bbS$ as the adjacency matrix of the graph with entries $S_{ij} = w_{ij}$, but nothing really changes if instead we work with the Laplacian or normalized versions of the adjacency or Laplacian \cite{Ortega18-GSP}. 

One advantage of graph filters is their locality. Indeed, we can define the diffusion sequence as the collection of graph signals $\bbz_k = \bbS^k \bbx$ to rewrite the filter in \eqref{eqn_filter} as $\bbu = \sum_{k=0}^{K} h_k \bbz_k$. It is ready to see that the diffusion sequence is given by the recursion $\bbz_k = \bbS\bbz_{k-1}$ with $\bbz_0=\bbx$. Further observing that $S_{ij}\neq 0$ only when the pair $(i,j)$ is an edge of the graph, we see that the entries of the diffusion sequence satisfy
\begin{equation} \label{eqn_diffusion_sequence}
   z_{k,i} \ = \ \sum_{j:(i,j)\in\ccalE} S_{ij} z_{k-1,j} .
\end{equation}
We can therefore interpret the graph filter in \eqref{eqn_filter} as an operation that propagates information through adjacent nodes as we illustrate in Fig. \ref{fig.evRecMain}. This is a property that graph convolutional filters share with regular convolutional filters in time and offers motivation for their use in the processing of graph signals.

In the context of machine learning on graphs, a more important property of graph filters is their \emph{equivariance to permutation.} Use $\bbP$ to denote a permutation matrix -- entries $P_{ij}$ are binary with exactly one nonzero entry in each row and column. The vector $\hbx = \bbP\bbx$ is just a reordering of the entries of $\bbx$ which we can interpret as a graph signal supported on the graph $\hbS = \bbP\bbS\bbP^T$ which is just a reordering of the graph $\bbS$. When processing of $\hbx$ on the graph $\hbS$ with the graph filter $\bbh$ the following proposition from \cite{Gama19-Stability}, {originally proved in \cite{Sandryhaila13-DSPG}}, holds.

%%%%%%%%%%%%%%%%%%%%%%%%%%%%%%%%%%%%%%%%%%%%%%%%%%%%%%%%%%%%%%%%%%%%%%%%%%%%%%%%
%   P   R   O   P   O   S   I   T   I   O   N   %%%%%%%%%%%%%%%%%%%%%%%%%%%%%%%%
%%%%%%%%%%%%%%%%%%%%%%%%%%%%%%%%%%%%%%%%%%%%%%%%%%%%%%%%%%%%%%%%%%%%%%%%%%%%%%%%
%
\begin{proposition}\label{prop_filter_equivariance} Graph filters are permutation equivariant,
\begin{align} \label{eqn_prop_filter_equivariance}
   \Phi (\hbx; \bbh, \hbS)
         =  \Phi (\bbP\bbx; \bbh, \bbP\bbS\bbP^T)
         =  \bbP \Phi (\bbx; \bbh, \bbS),
\end{align} \end{proposition}

%%%%%%%%%%%%%%%%%%%%%%%%%%%%%%%%%%%%%%%%%%%%%%%%%%%%%%%%%%%%%%%%%%%%%%%%%%%%%%%%
%   P   R   O   O   F   %%%%%%%%%%%%%%%%%%%%%%%%%%%%%%%%%%%%%%%%%%%%%%%%%%%%%%%%
%%%%%%%%%%%%%%%%%%%%%%%%%%%%%%%%%%%%%%%%%%%%%%%%%%%%%%%%%%%%%%%%%%%%%%%%%%%%%%%%
%
\begin{myproof} Use the definitions of the graph filter in \eqref{eqn_filter} and of the permutations $\hbx = \bbP\bbx$ and $\hbS = \bbP\bbS\bbP^T$ to write% 
\begin{align} \label{eqn_prop_filter_equivariance_pf}
   \Phi (\hbx; \bbh, \hbS)
        =  \sum_{k=0}^{K} h_k \hbS^k  \hbx
         =  \sum_{k=0}^{K} h_k \Big(\bbP \bbS \bbP^T\Big)^k\bbP\bbx 
\end{align}
Since $\bbP^T\bbP=\bbI$ for any permutation matrix, \eqref{eqn_prop_filter_equivariance} follows. \end{myproof}

%%%%%%%%%%%%%%%%%%%%%%%%%%%%%%%%%%%%%%%%%%%%%%%%%%%%%%%%%%%%%%%%%%%%%%%%%%%%%%%%
%   M   A   I   N       M   A   T   T   E   R   %%%%%%%%%%%%%%%%%%%%%%%%%%%%%%%%
%%%%%%%%%%%%%%%%%%%%%%%%%%%%%%%%%%%%%%%%%%%%%%%%%%%%%%%%%%%%%%%%%%%%%%%%%%%%%%%%
%
We include the proof of Proposition \ref{prop_filter_equivariance} to highlight that this is an elementary result. Its immediate relevance is that it shows that processing a graph signal with a graph filter is independent of node labeling. This is something we know must hold in several applications -- it certainly must hold for the recommendation problem described in Sec. \ref{sec_reco_systems} -- but that is not true of, say, the linear parametrization in \eqref{eqn_ch8_erm_linear}. There is, however, further value in permutation equivariance. To explain this, return to the ERM problem in \eqref{eqn_ch8_erm_learning} and utilize the graph filter in \eqref{eqn_filter} as a learning parametrization. This yields the learning problem
\begin{align}\label{eqn_ch8_erm_graph_filter}
   \bbh^* = \argmin_{\bbh} \frac{1}{Q} \sum_{q=1}^{Q} 
              \ell \bigg(\sum_{k=0}^K h_k\bbS^k \bbx _q,\,\bby _q \bigg) .
%         :=  \argmin_{\bbh}\, \bar\ell \big(\Phi(\xdot; \bbh, \bbS) \big) .
\end{align} 
An important observation is that we know that \eqref{eqn_ch8_erm_linear} must yield a function $\Phi(\bbx;\bbH^*)$ whose average loss is smaller than the average loss attained by the function $\Phi (\bbx; \bbh^*, \bbS)$ obtained from solving \eqref{eqn_ch8_erm_graph_filter}. This is because both are linear transformations and while $\Phi(\bbx;\bbH) = \bbH\bbx$ is generic, the graph filter $\Phi (\bbx; \bbh, \bbS) = \sum_{k=0}^{K} h_k \bbS^k \bbx$ belongs to a particular linear class. This is certainly true on the training set $\ccalT$, but when operating on unobserved samples $\bbx$ the graph filter can and will do better (see results in Sec. \ref{sec_reco_systems_results}) because its permutation equivariance induces better generalization.

%%%%%%%%%%%%%%%%%%%%%%%%%%%%%%%%%%%%%%%%%%%%%%%%%%%%%%%%%%%%%%%%%%%%%%%%%%%%%%%%
%   F   I   G   U   R   E   %%%%%%%%%%%%%%%%%%%%%%%%%%%%%%%%%%%%%%%%%%%%%%%%%%%%
%%%%%%%%%%%%%%%%%%%%%%%%%%%%%%%%%%%%%%%%%%%%%%%%%%%%%%%%%%%%%%%%%%%%%%%%%%%%%%%%
%
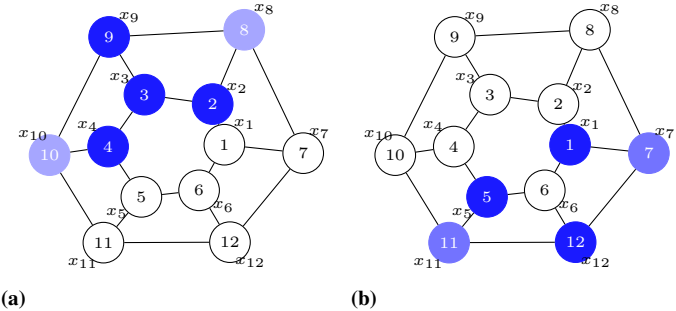
\begin{figure}[t] 
\def\myfactor{0.54}
\centering
\def \randcord {(0.28*rand, 0.28*rand)}
\pgfmathsetseed{13} 
\input{figures/random_hexagon_graph_signal_node_3_a}\hspace{-8mm}
\pgfmathsetseed{13} 
\input{figures/random_hexagon_graph_signal_node_6}
\hspace{8mm} 
{\bf \footnotesize(a)}\hspace{42mm}
{\bf \footnotesize(b)}\hspace{42mm}
\caption{Perfect symmetry as in Fig. \ref{fig_generalization} is unlikely in practice, but near permutation symmetries can and do appear. We still expect some level of generalization from graph filters [cf. \eqref{eqn_ch8_erm_graph_filter}] and GNNs [cf. \eqref{eqn_ch8_gnn_recursion_filter_matrix}-\eqref{eqn_ch8_gnn_operator_multifeature}].}
\label{fig_generalization_with_wiggliness}
\end{figure}

%%%%%%%%%%%%%%%%%%%%%%%%%%%%%%%%%%%%%%%%%%%%%%%%%%%%%%%%%%%%%%%%%%%%%%%%%%%%%%%%
%   M   A   I   N       M   A   T   T   E   R   %%%%%%%%%%%%%%%%%%%%%%%%%%%%%%%%
%%%%%%%%%%%%%%%%%%%%%%%%%%%%%%%%%%%%%%%%%%%%%%%%%%%%%%%%%%%%%%%%%%%%%%%%%%%%%%%%
%
An illustration of this phenomenon is shown in Fig. \ref{fig_generalization}. The graph represents a user similarity network in a recommendation system for which the ratings in (a) are available at training time. Out of these ratings we can create the graph signal in (b) to add to the training set and we assume that both parametrizations, the arbitrary linear transformation $\Phi(\bbx;\bbH^*)$ in \eqref{eqn_ch8_erm_linear} and the graph filter $\Phi (\bbx; \bbh^*, \bbS)$  in \eqref{eqn_ch8_erm_graph_filter}, learn to estimate the rating of user $3$ successfully. If this happens, the functions $\Phi(\bbx;\bbH^*)$ and $\Phi (\bbx; \bbh^*, \bbS)$ also learn to estimate the rating of user 3 when given the signal in (c) -- where we interpret colors as proportional to signal values. Notice that this happens even if signals of this form are not observed during training. We say that $\Phi(\bbx;\bbH^*)$ and $\Phi (\bbx; \bbh^*, \bbS)$ \emph{generalize} to this example. 

If we now consider the signal in (d), the linear parametrization $\Phi(\bbx;\bbH^*)$ may or may not generalize to this example. In principle, it would not. The graph filter $\Phi (\bbx; \bbh^*, \bbS)$, however, does generalize. This can be seen intuitively from the definition of the diffusion sequence in \eqref{eqn_diffusion_sequence}. Whatever operations are done to estimate the rating of user 3 from its adjacent nodes 2, 4 and 9 are the same as those done to estimate the rating of user 6 from its adjacent nodes 1, 5 and 12. {More formally, 
when graphs present symmetries in the sense that they are invariant to some permutation, i.e., $\bbS=\bbP\bbS\bbP^T$, Proposition \ref{prop_filter_equivariance} tells us that $\Phi(\bbP\bbx;\bbh,\bbS)=\bbP\Phi(\bbx;\bbh,\bbS)$, i.e., these operations are equivariant so that the rating prediction is consistent with this relabeling. This is the case of the graph in Fig. \ref{fig_generalization}, which can be permuted onto itself to map the signal in (d) onto the signal in (a). Thus, the graph filter \emph{generalizes} from the example in (a) to fill the rating in (d).}
%(e.g., the graph in Figure \red{X} can be permuted onto itself to map the signal in (d) into the signal in (a)), Proposition \ref{prop_filter_equivariance} says that these operations are equivariant, so that the rating prediction is consistent with this relabeling. The graph filter \emph{generalizes} from the example in (a) to fill the rating in (d).}
%the graph can be permuted onto itself to map the signal in (d) into the signal in (a) and Proposition \ref{prop_filter_equivariance} says that this is an equivariant operation so that the rating prediction is consistent with this relabeling. The graph filter \emph{generalizes} from the example in (a) to fill the rating in (d). 

This illustration highlights the generalization properties of graph filters vis-\`a-vis those of linear transforms. In reality, we are unlikely to encounter the perfect permutation symmetry of Fig. \ref{fig_generalization}. Near permutation symmetry as in Fig. \ref{fig_generalization_with_wiggliness} is more expected. In this case the ability to generalize from \ref{fig_generalization_with_wiggliness}-(a) to \ref{fig_generalization_with_wiggliness}-(b) is not as much as the ability to generalize from \ref{fig_generalization}-(a) to \ref{fig_generalization}-(d) but the continuity of \eqref{eqn_filter} dictates that some amount of predictive power extends from observing samples \ref{fig_generalization_with_wiggliness}-(a) towards the estimation of the rating of user 6 when given the signal in \ref{fig_generalization_with_wiggliness}-(b). 

%%%%%%%%%%%%%%%%%%%%%%%%%%%%%%%%%%%%%%%%%%%%%%%%%%%%%%%%%%%%%%%%%%%%%%%%%%%%%%%%
%   F   I   G   U   R   E   %%%%%%%%%%%%%%%%%%%%%%%%%%%%%%%%%%%%%%%%%%%%%%%%%%%%
%%%%%%%%%%%%%%%%%%%%%%%%%%%%%%%%%%%%%%%%%%%%%%%%%%%%%%%%%%%%%%%%%%%%%%%%%%%%%%%%
%
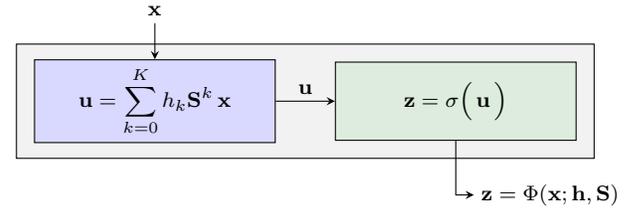
\begin{figure}[t] 
\def\myfactor{1}
\centering
\input{figures/graph_perceptron.tex}
\caption{A graph perceptron composes a graph convolutional filter with a pointwise nonlinearity. It is a minor variation of a graph filter which, among other shared properties, retains permutation equivariance.}
\label{fig_ch8_graph_perceptron}
\end{figure}

%%%%%%%%%%%%%%%%%%%%%%%%%%%%%%%%%%%%%%%%%%%%%%%%%%%%%%%%%%%%%%%%%%%%%%%%%%%%%%%%
%   S   E   C   T   I   O   N   %%%%%%%%%%%%%%%%%%%%%%%%%%%%%%%%%%%%%%%%%%%%%%%%
%%%%%%%%%%%%%%%%%%%%%%%%%%%%%%%%%%%%%%%%%%%%%%%%%%%%%%%%%%%%%%%%%%%%%%%%%%%%%%%%
%
\subsection{Graph Perceptrons}\label{sec_perceptrons}

Graph neural networks (GNNs) extend graph filters by using pointwise nonlinearities which are nonlinear functions that are applied independently to each component of a vector. For a formal definition, begin by introducing a single variable function $\sigma:\reals\to\reals$ which we extend to the vector function $\sigma:\reals^{n}\to\reals^{n}$ by independent application to each component. Thus, if we have $\bbu = [u_1; \ldots; u_{n}] \in \reals^{n}$ the output vector $\sigma(\bbu)$ is such that
\begin{equation} \label{eqn_ch8_pointwise_nonlinearity}
   \sigma\big(\,\bbu\,\big) \ : \ \big[\,\sigma\big(\,\bbu\,\big)\,\big]_i = \sigma\big(\,u_i\,\big).
\end{equation}
I.e., the output vector is of the form $\sigma(\bbu) = [\sigma(u_1); \ldots; \sigma(u_{n})]$. Observe that we are abusing notation and using $\sigma$ to denote both the scalar function and the pointwise vector function.

In a single layer GNN, the graph signal $\bbu$ is passed trough a pointwise nonlinear function satisfying \eqref{eqn_ch8_pointwise_nonlinearity}  to yield
\begin{equation} \label{eqn_ch8_perceptron_nonlinearity}
	\bbz \  = \ \sigma  \Big(\, \bbu \, \Big)
%	     \  = \ \sigma \Big(\,\bbH(\bbS) \bbx \, \Big)	  
	     \  = \ \sigma \Bigg (\sum_{k=0}^{K} h_{k} \bbS^{k} \bbx \Bigg) .
\end{equation}
We say the transform in \eqref{eqn_ch8_perceptron_nonlinearity} is a graph perceptron; see Fig. \ref{fig_ch8_graph_perceptron}. Different from the graph filter in \eqref{eqn_filter}, the graph perceptron is a nonlinear function of the input. It is, however, a very simple form of nonlinear processing because the nonlinearity does not mix signal components. Signal components are mixed by the graph filter but are then processed element-wise through $\sigma$. In particular, \eqref{eqn_ch8_perceptron_nonlinearity} retains the locality properties of graph convolutional filters (cf. Fig. \ref{fig.evRecMain}) as well as their permutation equivariance (cf. Fig. \ref{fig_generalization} and Proposition \ref{prop_filter_equivariance}).

%%%%%%%%%%%%%%%%%%%%%%%%%%%%%%%%%%%%%%%%%%%%%%%%%%%%%%%%%%%%%%%%%%%%%%%%%%%%%%%%
%   F   I   G   U   R   E   %%%%%%%%%%%%%%%%%%%%%%%%%%%%%%%%%%%%%%%%%%%%%%%%%%%%
%%%%%%%%%%%%%%%%%%%%%%%%%%%%%%%%%%%%%%%%%%%%%%%%%%%%%%%%%%%%%%%%%%%%%%%%%%%%%%%%
%
\begin{figure}[t] 
\def\myfactor{0.9}
\centering
\input{figures/gnn_block_diagram_single_feature}%\\ {\small (a)}
\caption{Graph Neural Networks are compositions of layers each of which composes graph filters $\Phi (\bbx; \bbh_l, \bbS)=\sum_{k=0}^K h_{lk}\bbS^k$ with pointwise nonlinearities $\sigma$ [cf.  \eqref{eqn_ch8_gnn_recursion_single_feature_filter} and \eqref{eqn_ch8_gnn_recursion_single_feature_nonlinearity}]. The output $\Phi (\bbx; \bbH, \bbS) = \bbx_L = \bbx_3$ follows at the end of a cascade of 3 layers recursively applied to the input $\bbx$. Layers are defined by sets of coefficients grouped in the {matrix} $\bbH:=\{\bbh_1, \bbh_2, \bbh_3\}$ which is chosen to minimize a training loss for a given shift $\bbS$ [cf. \eqref{eqn_ch8_erm_learning} and \eqref{eqn_ch8_erm_gnn_single_feature}].}
\label{fig_ch8_gnn_block_diagram}
\end{figure}
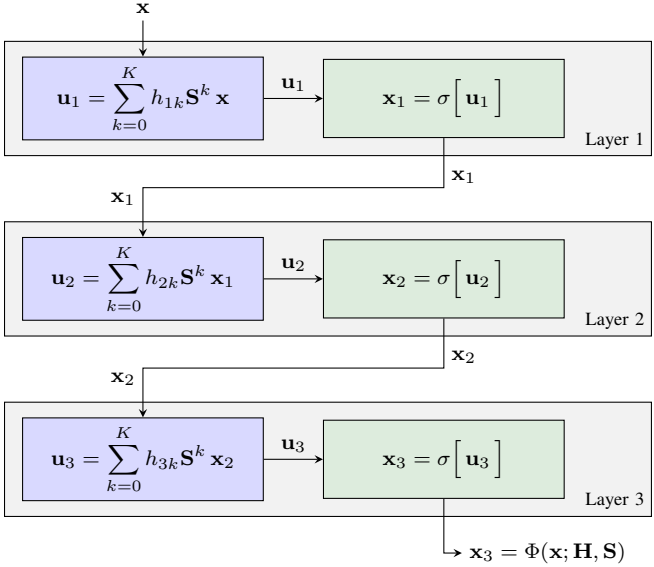

%%%%%%%%%%%%%%%%%%%%%%%%%%%%%%%%%%%%%%%%%%%%%%%%%%%%%%%%%%%%%%%%%%%%%%%%%%%%%%%%
%   S   E   C   T   I   O   N   %%%%%%%%%%%%%%%%%%%%%%%%%%%%%%%%%%%%%%%%%%%%%%%%
%%%%%%%%%%%%%%%%%%%%%%%%%%%%%%%%%%%%%%%%%%%%%%%%%%%%%%%%%%%%%%%%%%%%%%%%%%%%%%%%
%
\subsection{Multiple Layer Networks}\label{sec_ch8_gnns_multiple_layers}

Graph perceptrons can be stacked in layers to create multilayer GNNs -- see Fig. \ref{fig_ch8_gnn_block_diagram}. This stacking is mathematically written as a function composition where the outputs of a layer become inputs to the next layer. For a formal definition let $l=1,\ldots,L$ be a layer index and $\bbh_l=\{h_{lk}\}_{k=0}^K$ be collections of $K+1$ graph filter coefficients associated with each layer. Each of these sets of coefficients define a respective graph filter $\Phi (\bbx; \bbh_l, \bbS) = \sum_{k=0}^K h_{lk}\bbS^k\bbx$. At layer $l$ we take as input the output $\bbx_{l-1}$ of layer $l-1$ which we process with the filter $\Phi (\bbx; \bbh_l, \bbS)$ to produce the intermediate feature
\begin{align}\label{eqn_ch8_gnn_recursion_single_feature_filter}
   \bbu_{l} \ = \ \bbH_l(\bbS)\, \bbx_{l-1} 
            \ = \ \sum_{k=0}^K h_{lk}\bbS^k\, \bbx_{l-1},
\end{align} 
where, by convention, we say that $\bbx_0 = \bbx$ so that the given graph signal $\bbx$ is the GNN input. As for the graph perceptron, this feature is passed through a pointwise nonlinear function (which is the same in all layers) to produce the $l$th layer output
\begin{align}\label{eqn_ch8_gnn_recursion_single_feature_nonlinearity}
   \bbx_{l} \ = \ \sigma(\bbu_l )  
%            \ = \ \sigma\Big(\bbH_l(\bbS)\, \bbx_{l-1} \Big)  
            \ = \ \sigma\Bigg(\sum_{k=0}^K h_{lk}\bbS^k\, \bbx_{l-1}\Bigg) .
\end{align} 
After recursive repetition of \eqref{eqn_ch8_gnn_recursion_single_feature_filter}-\eqref{eqn_ch8_gnn_recursion_single_feature_nonlinearity} for $l=1,\ldots,L$ we reach  $\bbx_L$, which is not further processed and is declared the GNN output $\bbz=\bbx_L$. To represent the GNN output we define the filter {matrix} $\bbH:=\{\bbh_l\}_{l=1}^L$ grouping the $L$ sets of filter coefficients at each layer, and define the operator $\Phi(\,\cdot\,;\bbH,\bbS)$ as
\begin{align}\label{eqn_ch8_gnn_operator_single_feature}
   \Phi (\bbx; \bbH, \bbS) = \bbx_L.
\end{align} 
We stress that in \eqref{eqn_ch8_gnn_operator_single_feature} the GNN output $\Phi (\bbx; \bbH, \bbS) = \bbx_L$ follows from recursive application of \eqref{eqn_ch8_gnn_recursion_single_feature_filter}-\eqref{eqn_ch8_gnn_recursion_single_feature_nonlinearity} for $l=1,\ldots,L$ with $\bbx_0=\bbx$. This operator notation emphasizes that the output of a GNN depends on the filter $\bbH$ and the graph shift operator $\bbS$. 
A block diagram for a GNN with $L=3$ layers is shown in Fig. \ref{fig_ch8_gnn_block_diagram}. 

%The input $\bbx$ is fed to the first layer where it is processed by the filter $\sum_{k=0}^K h_{1k}\bbS^k$ and passed through the pointwise nonlinearity $\sigma$ to produce the first layer output 
%
%\begin{align}\label{eqn_ch8_gnn_single_feature_layer_1}
%   \bbx_{1} = \sigma\Bigg(\sum_{k=0}^K h_{1k}\bbS^k\, \bbx \Bigg).
%\end{align}
%
%This is according to \eqref{eqn_ch8_gnn_recursion_single_feature_filter}-\eqref{eqn_ch8_gnn_recursion_single_feature_nonlinearity} with $l=1$ and $\bbx_0=\bbx$. The output of Layer 1 is sent to Layer 2 where it is processed by the filter $\sum_{k=0}^K h_{2k}\bbS^k$ and passed through the pointwise nonlinearity $\sigma$ to produce the Layer 2 output 
%
%\begin{align}\label{eqn_ch8_gnn_single_feature_layer_2}
%   \bbx_{2} = \sigma\Bigg(\sum_{k=0}^K h_{2k}\bbS^k\, \bbx_1\Bigg),
%\end{align}
%
%as per \eqref{eqn_ch8_gnn_recursion_single_feature_filter}-\eqref{eqn_ch8_gnn_recursion_single_feature_nonlinearity} with $l=2$. This output becomes an input to Layer 3 where it is processed to produce the Layer 3 output 
%
%\begin{align}\label{eqn_ch8_gnn_single_feature_layer_3}
%   \bbx_{3} = \sigma\Bigg(\sum_{k=0}^K h_{3k}\bbS^k\, \bbx_2\Bigg),
%\end{align}
%
%again, as dictated by \eqref{eqn_ch8_gnn_recursion_single_feature_filter}-\eqref{eqn_ch8_gnn_recursion_single_feature_nonlinearity}. Since this is a GNN with $L=3$ layers, this becomes the output of the GNN $\Phi (\bbx; \bbH, \bbS) = \bbx_L = \bbx_3$. Observe that each layer is defined by a set of filter coefficients that are grouped in the tensor $\bbH:=\{\bbh_1, \bbh_2, \bbh_3\}$. 
The sets of filter coefficients $\bbH$ that define the GNN operator in \eqref{eqn_ch8_gnn_operator_single_feature} are chosen to minimize a training loss as in \eqref{eqn_ch8_erm_learning},
\begin{align}\label{eqn_ch8_erm_gnn_single_feature}
   \bbH^* = \argmin_{\bbH} \frac{1}{Q} \sum_{q=1}^{Q} 
              \ell \Big(\Phi \big(\bbx_q; \bbH, \bbS\big),\, \bby _q \Big) .
%         :=  \argmin_{\bbh}\, \bar\ell \big(\Phi(\xdot; \bbh, \bbS) \big) .
\end{align} 
We emphasize that, similar to the case of the graph filters in \eqref{eqn_ch8_erm_graph_filter}, the optimization is over the filter {matrix} $\bbH$ with the shift operator $\bbS$ given. We also note that since each perceptron is permutation equivariant, the whole GNN also inherits the permutation equivariance of graph filters. 

%\paragraph{Increasing Representation Power.} The class of functions that can be represented with all possible perceptrons is larger than the class of functions that can be represented with all possible graph filters. This is because the latter are linear functions while the former are nonlinear. The class of functions that can be represented with a multilayer GNN is, in turn, larger than the class of functions representable with a perceptron. This is true because the composition of perceptrons is not a graph perceptron. This is to be contrasted with a composition of graph filters which is also a graph filter. If, for example, $\bbH_1(\bbS)$ and $\bbH_2(\bbS)$ are graph filters, their composition is the product $\bbH_1(\bbS)\bbH_2(\bbS)$ which can be written as a polynomial on $\bbS$ and is therefore also a graph filter. Thus, a composition of graph filters would not enlarge the class of representable functions. But a composition of perceptrons does.

%%%%%%%%%%%%%%%%%%%%%%%%%%%%%%%%%%%%%%%%%%%%%%%%%%%%%%%%%%%%%%%%%%%%%%%%%%%%%%%%
%   F   I   G   U   R   E   %%%%%%%%%%%%%%%%%%%%%%%%%%%%%%%%%%%%%%%%%%%%%%%%%%%%
%%%%%%%%%%%%%%%%%%%%%%%%%%%%%%%%%%%%%%%%%%%%%%%%%%%%%%%%%%%%%%%%%%%%%%%%%%%%%%%%
%
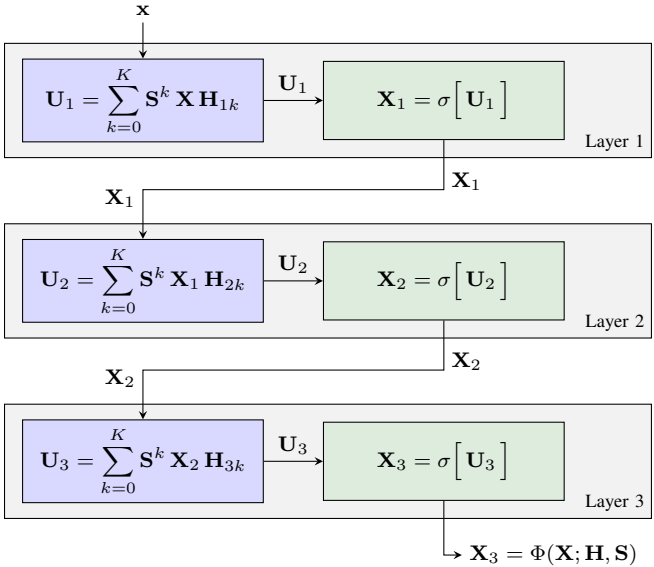
\begin{figure}[t] 
\def\myfactor{0.9}
\centering
\input{figures/gnn_block_diagram_mimo_filter}%\\ {\small (a)}
\caption{We expand the representation power of Graph Neural Networks (GNNs) with the addition of multiple features per layer [cf. \eqref{eqn_ch8_gnn_features}]. The graph filters in each layer are multiple-input-multiple-output graph filters (cf. \eqref{eqn_ch8_gnn_recursion_filter_matrix}]. They take $F_{l-1}$ graph signals as inputs and produce $F_l$ graph signals as outputs. The structure is otherwise identical to the single feature GNN in Fig. \ref{fig_ch8_gnn_block_diagram}.}
\label{fig_ch8_gnn_block_diagram_mimo}
\end{figure}

%%%%%%%%%%%%%%%%%%%%%%%%%%%%%%%%%%%%%%%%%%%%%%%%%%%%%%%%%%%%%%%%%%%%%%%%%%%%%%%%
%   S   E   C   T   I   O   N   %%%%%%%%%%%%%%%%%%%%%%%%%%%%%%%%%%%%%%%%%%%%%%%%
%%%%%%%%%%%%%%%%%%%%%%%%%%%%%%%%%%%%%%%%%%%%%%%%%%%%%%%%%%%%%%%%%%%%%%%%%%%%%%%%
%
\subsection{Multiple Feature Networks}\label{sec_ch8_gnn_multiple_features}

To further increase the representation power of GNNs we incorporate multiple features per layer that are the result of processing multiple input features with a bank of graph filters; see Fig. \ref{fig_ch8_gnn_block_diagram_mimo}. For a formal definition let $F_l$ be the number of features at layer $l$ and define the feature matrix as
\begin{equation} \label{eqn_ch8_gnn_features}
   \bbX_l = \left[ \bbx_{l}^{1},\, \bbx_{l}^{2},\, \ldots,\, \bbx_{l}^{F_l} \right].
\end{equation}
We have that $\bbX_l\in\reals^{n\times F_l}$ and interpret each column of $\bbX_l$ as a graph signal. The outputs of Layer $l-1$ are inputs to Layer $l$ where the set of $F_{l-1}$ features in $\bbX_{l-1}$ are processed by a filterbank made up of $F_{l-1}\times F_l$ filters. For a compact representation of this bank consider coefficient matrices $\bbH_{lk}\in\reals^{F_{l-1}\times F_l}$ to build the intermediate feature matrix
\begin{align}\label{eqn_ch8_gnn_recursion_filter_matrix}
    \bbU_{l}
       \ = \ 
          \sum_{k=0}^K 
             \bbS^{k}\bbX_l \, 
                \bbH_{lk}, 
\end{align} 
Each of the $F_l$ columns of the matrix $\bbU_l\in\reals^{n\times F_l}$ is a separate graph signal. We say that \eqref{eqn_ch8_gnn_recursion_filter_matrix} represents a multiple-input-multiple-output (MIMO) graph filter since it takes $F_{l-1}$ graph signals as inputs and yields $F_{l}$ graph signals at its output. As in the case of the single feature GNN of Sec. \ref{sec_ch8_gnns_multiple_layers} -- and the graph perceptron in \eqref{eqn_ch8_perceptron_nonlinearity} -- the intermediate feature $\bbU_{l}$ is passed through a pointwise nonlinearity to produce the $l$th layer output
\begin{align}\label{eqn_ch8_gnn_recursion_nonlinearity_matrix}
   \bbX_{l} \ = \ \sigma(\bbU_l )  
            \ = \ \sigma\Bigg(\sum_{k=0}^K \bbS^k\, \bbX_{l-1}\, \bbH_{lk} \Bigg).
\end{align} 
When $l=0$ we convene that $\bbX_0=\bbX$ is the input to the GNN which is made of $F_0$ graph signals. The output $\bbX_L$ of layer $L$ is also the output of the GNN which is made up of $F_L$ graph signals. To define a GNN operator we group filter coefficients $\bbH_{lk}$ in the tensor $\bbH = \{\bbH_{lk}\}_{l,k}$ and define the GNN operator
\begin{align}\label{eqn_ch8_gnn_operator_multifeature}
   \Phi (\bbX; \bbH, \bbS) = \bbX_L.
\end{align} 
If the input is a single graph signal as in \eqref{eqn_ch8_perceptron_nonlinearity} and \eqref{eqn_ch8_gnn_operator_single_feature}, we have $F_0=1$ and $\bbX_0=\bbx\in\reals^n$. If the output is also a single graph signal -- as is also the case in \eqref{eqn_ch8_perceptron_nonlinearity} and \eqref{eqn_ch8_gnn_operator_single_feature} -- we have $F_L=1$ and $\bbX_L=\bbx_L\in\reals^n$.

The sets of filter coefficients $\bbH$ that define the  multiple feature GNN operator in \eqref{eqn_ch8_gnn_operator_multifeature} are chosen to minimize a training loss
\begin{align}\label{eqn_ch8_erm_gnn_mf}
   \bbH^* = \argmin_{\bbH} \frac{1}{Q} \sum_{q=1}^{Q} 
              \ell \Big(\Phi \big(\bbX_q; \bbH, \bbS\big),\, \bbY _q \Big) ,
%         :=  \argmin_{\bbh}\, \bar\ell \big(\Phi(\xdot; \bbh, \bbS) \big) .
\end{align} 
which differs from \eqref{eqn_ch8_erm_gnn_single_feature} in that inputs, outputs, and intermediate layers may be composed of multiple features. Each layer of the GNN is made up of filter banks which are permutation equivariant. Since pointwise nonlinearities do not mix signal components, each individual layer is permutation equivariant. It follows that the GNN, being a composition of permutation equivariant operators, is also permutation equivariant. This is a sufficiently important fact that deserves to be highlighted as a proposition that we take from \cite{Gama19-Stability}.

%%%%%%%%%%%%%%%%%%%%%%%%%%%%%%%%%%%%%%%%%%%%%%%%%%%%%%%%%%%%%%%%%%%%%%%%%%%%%%%%
%   P   R   O   P   O   S   I   T   I   O   N   %%%%%%%%%%%%%%%%%%%%%%%%%%%%%%%%
%%%%%%%%%%%%%%%%%%%%%%%%%%%%%%%%%%%%%%%%%%%%%%%%%%%%%%%%%%%%%%%%%%%%%%%%%%%%%%%%
%
\begin{proposition}\label{prop_gnn_equivariance} GNNs are permutation equivariant,
\begin{align} \label{eqn_prop_gnn_equivariance}
   \Phi (\hbx; \bbH, \hbS)
         =  \Phi (\bbP\bbx; \bbH, \bbP\bbS\bbP^T)
         =  \bbP \Phi (\bbx; \bbH, \bbS).
\end{align} \end{proposition}

That Proposition \ref{prop_gnn_equivariance} holds entails that the same comments that follow Proposition \ref{prop_filter_equivariance} hold for GNNs. In particular, GNNs are expected to generalize from observing the signal in Fig. \ref{fig_generalization}-(a) to successfully fill in ratings when presented with the signal in Fig. \ref{fig_generalization}-(d), even if this signal is never observed during training. This is an attribute that is not expected of fully connected neural networks -- and that we verify experimentally  in Sec. \ref{sec_reco_systems_results}. Likewise, we expect generalization to also hold in the case of Fig. \ref{fig_generalization_with_wiggliness}. As we will see in Sec. \ref{sec:stability}, the fundamental difference between GNNs and graph filters is the ability of the former to provide better generalization when signals are close to permutation equivariant but not exactly so.

%%%%%%%%%%%%%%%%%%%%%%%%%%%%%%%%%%%%%%%%%%%%%%%%%%%%%%%%%%%%%%%%%%%%%%%%%%%%%%%%
%   R   E   M   A   R   K   %%%%%%%%%%%%%%%%%%%%%%%%%%%%%%%%%%%%%%%%%%%%%%%%%%%%
%%%%%%%%%%%%%%%%%%%%%%%%%%%%%%%%%%%%%%%%%%%%%%%%%%%%%%%%%%%%%%%%%%%%%%%%%%%%%%%%
%
\begin{remark}\label{rmk_for_dcis}
\normalfont
As is the case of the single feature filter in \eqref{eqn_filter}, we can write the MIMO graph filter in \eqref{eqn_ch8_gnn_recursion_filter_matrix} in terms of a diffusion sequence. To do that, we define $\bbZ_{lk} := \bbS^k\bbX_l$ and observe that we can rewrite the matrices $\bbZ_{lk}$ in the recursive form
\begin{align}\label{eqn_MIMO_diffusion_sequence}
       \bbZ_{lk} = \bbS\bbZ_{l,k-1}, \quad \text{with~} \bbZ_{l0} = \bbX_l.
\end{align} 
With this definition the graph filter in \eqref{eqn_ch8_gnn_recursion_filter_matrix} is rewritten as 
\begin{align}\label{eqn_ch8_gnn_recursion_filter_matrix_diffusion_sequence}
    \bbU_{l}
       \ = \ 
          \sum_{k=0}^K 
             \bbS^{k}\bbX_l \, 
                \bbH_{lk}, 
       \ = \ 
          \sum_{k=0}^K 
             \bbZ_{lk} \, 
                \bbH_{lk}, 
\end{align} 
The use of the diffusion sequence in \eqref{eqn_ch8_gnn_recursion_filter_matrix_diffusion_sequence} highlights that the MIMO graph filter accepts a local implementation [cf, \eqref{eqn_diffusion_sequence}]. This is important in, e.g., the use of GNNs in decentralized collaborative systems (Section \ref{sec_dcis}).
\end{remark}

%%%%%%%%%%%%%%%%%%%%%%%%%%%%%%%%%%%%%%%%%%%%%%%%%%%%%%%%%%%%%%%%%%%%%%%%%%%%%%%%
%   R   E   M   A   R   K   %%%%%%%%%%%%%%%%%%%%%%%%%%%%%%%%%%%%%%%%%%%%%%%%%%%%
%%%%%%%%%%%%%%%%%%%%%%%%%%%%%%%%%%%%%%%%%%%%%%%%%%%%%%%%%%%%%%%%%%%%%%%%%%%%%%%%
%
\begin{remark}
\normalfont
To keep the representation dimension under control, many architectures implement pooling as an intermediate step between the convolutional filter banks and the nonlinearity. Pooling is a summarizing operation that reduces dimensionality by first computing local summaries of the signal and then subsampling it. Permutation equivariance is preserved if the subsampling operation is based on topological features of the graph such as the node degrees \cite{Gama19-Architectures}. Pooling strategies for GNNs have been discussed in \cite{Gama19-Architectures,Bruna14-DeepSpectralNetworks,Defferrard17-ChebNets,cheung2020graph}.
\end{remark}

%%%%%%%%%%%%%%%%%%%%%%%%%%%%%%%%%%%%%%%%%%%%%%%%%%%%%%%%%%%%%%%%%%%%%%%%%%%%%%%%
%   R   E   M   A   R   K   %%%%%%%%%%%%%%%%%%%%%%%%%%%%%%%%%%%%%%%%%%%%%%%%%%%%
%%%%%%%%%%%%%%%%%%%%%%%%%%%%%%%%%%%%%%%%%%%%%%%%%%%%%%%%%%%%%%%%%%%%%%%%%%%%%%%%
%
%\begin{remark}\normalfont Just adding the signals $\bbv_l^{fg}$ in \eqref{eqn_ch8_gnn_filter_addition} seems arbitrary. Having general linear combinations of features and having some output features $\bbx_{l}^g$ being dependent on only a subset of input features $\bbx_{l-1}^f$ seems more general. There is no difference, however. Since the filter coefficients in the tensor $\bbH$ are trained [cf. \eqref{eqn_ch8_erm_gnn_mf}], it is equivalent to search for optimal filter coefficients if they are added up or if they are linearly combined. The latter is just a scaling of filter coefficients. In particular, this includes cases in which some input features $\bbx_{l-1}^f$ do not influence some output features $\bbx_{l}^g$. This could be accomplished by excluding index $f$ from the summation in \eqref{eqn_ch8_gnn_filter_addition} but this is equivalent to having a filter $\bbh_{l}^{fg}$ with all-zero coefficients.
%\end{remark}

%%%%%%%%%%%%%%%%%%%%%%%%%%%%%%%%%%%%%%%%%%%%%%
%%%%%%%%%%%%%%%%% SECTION %%%%%%%%%%%%%%%%%%%%
%%%%%%%%%%%%%%%%%%%%%%%%%%%%%%%%%%%%%%%%%%%%%%

\subsection{Recommendation System Experiments} \label{sec_reco_systems_results}

To illustrate the problem of recommendation systems with a specific numerical example, we consider the MovieLens-100k dataset \cite{Harper16-MovieLens}, which consists of 100,000 ratings given by 943 users to 1,682 movies. These ratings are integers between 1 and 5, and non-existing ratings are set to 0. The movie similarity network is built by computing similarity scores between pairs of movies as described in Sec. \ref{sec_reco_systems}. On this network, each user's rating vector $\bbx_{c}$ can be represented as a graph signal.

\begin{table}[t] 
\centering
{\footnotesize
\begin{tabular}{l|c|c|c|c} \hline
Architecture         & $L$ & Hyperparameters & {Params.} & $\sigma$  \\ \hline
Linear    & -            & $n \times n$ matrix & {2.8E+{6}}  & -             \\
Graph filter     & -            & $F=64$, $K=5$                         & {384} & -              \\
FCNN             & 2            & $N_1=64$, $N_2=32$                    & {1.6{E}+5} & ReLU          \\
Graph perceptron & 1            & $K=5$                                 & {6} & ReLU          \\
GNN & 2            & $F=1$, $K=5$                                 & {11} & ReLU           \\
GNN              & 1            & $F=64$, $K=5$                         & {384} & ReLU           \\
GNN              & 2            & $F_1=64$, $F_2=32$, $K=5$ & {1{E}+4}           & ReLU         \\ \hline
\end{tabular}
}
\caption{{Hyperparameters and total number of parameters of seven parametrizations of $\Phi$ in \eqref{eqn_ch8_erm_learning}. The number of features, filter taps and hidden units are denoted $F$, $K$ and $N$ respectively. For multi-layer models, $F_l/N_l$ indicate the value of these hyperparameters at layer $l$.}}
\label{tab:hyperparam}
\end{table}

\medskip\noindent\textbf{Different parametrizations.} In the first experiment the goal is to predict the ratings to the {six movies with most ratings in the dataset} 
%movie ``Star Wars'' 
by solving the ERM problem in \eqref{eqn_ch8_erm_learning} with different parametrizations of $\Phi$. {In order to do this, we follow the methodology in Sec. \ref{sec_reco_systems} to obtain 3044 input-output pairs corresponding to users who have rated these movies. This data is then split between 90\% for training (of which 10\% are used for validation) and 10\% for testing. 

Seven different parametrizations were considered: a simple linear parametrization; a graph filter \eqref{eqn_filter}; a fully connected neural network; a graph perceptron \eqref{eqn_ch8_perceptron_nonlinearity}; a multi-layer GNN \eqref{eqn_ch8_gnn_recursion_single_feature_nonlinearity}; and a single-layer and a multi-layer multi-feature GNNs \eqref{eqn_ch8_gnn_recursion_nonlinearity_matrix}. Their hyperparameters are presented in Table \ref{tab:hyperparam}. Note that the graph filter and GNNs have a readout layer mapping $F_L$ features per node to a single output feature per node, adding $F_L$ extra parameters. All architectures were trained simultaneously by optimizing the L1 loss on the training set, using ADAM with learning rate $5 \times 10^{-3}$ and decay factors $0.9$ and $0.999$. The number of epochs and batch size were 40 and 5 respectively.}

In Table \ref{tab:movie_results}, we report the average root mean square error (RMSE) achieved by each parametrization for 10 data splits. {We observe that the graph filter achieves a much smaller error than the generic linear parametrization {while having significantly less parameters}, which is empirical evidence of its superior ability to exploit the structure of graph signals through permutation equivariance as discussed in Sec. \ref{sec_ch8_gnns}. The fact that the average RMSE of the fully connected neural network (FCNN), {which also has in the order of $10^5$ parameters}, is worse than those of the GNNs, graph perceptrons and graph filter can be explained by the same reason, even if the FCNN improves upon the linear transformation due to the nonlinearities.} The graph perceptron and multi-layer GNN are not better than the graph filter and showcase similar RMSEs. On the other hand, the addition of multiple features in the single-layer and multi-layer GNNs provide sensible improvements, with the 2-layer GNN performing better than all other architectures. It turns out that nonlinearities also play an important role in GNN performance, which we examine in the stability discussion of Sec. \ref{sec:stability}.

\begin{table}[t]
\centering
{\footnotesize
\begin{tabular}{l|c} \hline
{Parametrization}        & {RMSE}   \\ \hline
{Linear parametrization} & {1.967} \\
{Graph filter}           & {1.054} \\
{FCNN}                   & {1.116} \\
{Graph perceptron, $L=1$} & {1.079} \\
{GNN, $L=2$} & {1.076} \\
{GNN, $L=1, F=64$}              & {1.050} \\
{GNN, $L=2, F_1=64, F_2=32$}              & {\textbf{0.964}} \\ \hline
\end{tabular}}
\caption{{Average RMSE over 10 random data splits for the six movies with most ratings in the dataset.}}
\label{tab:movie_results}
\end{table}

\medskip\noindent\textbf{GNN transferability.} In the second experiment, we aim to analyze whether a GNN trained on a small network generalizes well to a large network. We consider the same parametrization of the {2-layer} GNN in Table \ref{tab:hyperparam} and use the same training parameters of the first experiment. The GNN is trained to predict the ratings of the movie ``Star Wars'' on similarity networks with {$n=118, 203, 338, 603$ and $1682$ nodes}, where one of the nodes is always ``Star Wars'' and the others are picked at random. After training, each GNN is then tested on the full movie network. 

Table \ref{tab:transf} shows the average RMSEs obtained on both the graph where the GNN was trained and the full movie graph for 10 random data splits. It also shows the average relative difference between the RMSE on these graphs. We observe that the prediction error on the full movie network approaches the error realized on the trained network as $n$ increases. 
%Even for $n=250$, this difference is relatively low (under 10\%). 
These results suggest that GNNs are \textit{transferable}, a property that we discuss in more detail in Sec. \ref{sec:transferability}.

\begin{table}[t]
\centering
{\footnotesize
\begin{tabular}{l|ccccc} \hline
     Graph\ /\ $n$        & {$118$} & {$203$} & {$338$} & {$603$} & {$1682$}\\ \hline
$n$ nodes    & {0.829} & {0.818} & {0.863} &
 {0.866} & {0.873}\\
Full graph & {4.069} & {3.908} & {2.150} & {1.201} &  {0.873}\\ \hline
Difference   & {79.5\%} & {79.0\%} & {56.0\%} & {23.4\%} & {0.0\%}\\ \hline
\end{tabular}}
\caption{{Average RMSE achieved on the graph where the GNN is trained ($n$ nodes) and on the full movie graph for the movie ``Star Wars''. Average relative RMSE difference.}}
\label{tab:transf}
\end{table}

%% file: figures/hexagon_graph_signal_node_3_a.tex
\ifcompiletikz 
{\fontsize{6}{6}\selectfont\begin{tikzpicture}[scale = \myfactor]

  % Define center of hexagon
  \node                         []     (center) {};
  % Nodes in the inner hexagon 
    
  \path (center) ++ (  0:1.4) node [empty node]   (1) {$1$}  ++ ( 0.5,  0.5) node {$x_1$};
  \path (center) ++ ( 60:1.4) node [node value a] (2) {$2$}  ++ ( 0.6,  0.4) node {$x_2$};
  \path (center) ++ (120:1.4) node [node value a] (3) {$3$}  ++ (-0.6,  0.4) node {$x_3$};
  \path (center) ++ (180:1.4) node [node value a] (4) {$4$}  ++ (-0.5,  0.5) node {$x_4$};
  \path (center) ++ (240:1.4) node [empty node]   (5) {$5$}  ++ (-0.6, -0.4) node {$x_5$};
  \path (center) ++ (300:1.4) node [empty node]   (6) {$6$}  ++ ( 0.6, -0.4) node {$x_6$};
  % Nodes in the outer hexagon 
  \path (center) ++ (  0:3.0) node [empty node]   (7)  {$7$}  ++ ( 0.4,  0.5) node {$x_7$};
  \path (center) ++ ( 60:3.0) node [node value c] (8)  {$8$}  ++ ( 0.5,  0.5) node {$x_8$};
  \path (center) ++ (120:3.0) node [node value a] (9)  {$9$}  ++ ( 0.5,  0.5) node {$x_9$};
  \path (center) ++ (180:3.0) node [node value c] (10) {$10$} ++ (-0.4,  0.5) node {$x_{10}$};
  \path (center) ++ (240:3.0) node [empty node]   (11) {$11$} ++ (-0.5, -0.5) node {$x_{11}$};
  \path (center) ++ (300:3.0) node [empty node]   (12) {$12$} ++ ( 0.5, -0.5) node {$x_{12}$};

  % Edges connecting inner hexagon  
  \path (1)  edge [tight edge] node {} (2);
  \path (2)  edge [tight edge] node {} (3);
  \path (3)  edge [tight edge] node {} (4);
  \path (4)  edge [tight edge] node {} (5);
  \path (5)  edge [tight edge] node {} (6);
  \path (6)  edge [tight edge] node {} (1);

  % Edges connecting outer hexagon  
  \path (7)   edge [tight edge] node {}  (8);
  \path (8)   edge [tight edge] node {}  (9);
  \path (9)   edge [tight edge] node {} (10);
  \path (10)  edge [tight edge] node {} (11);
  \path (11)  edge [tight edge] node {} (12);
  \path (12)  edge [tight edge] node {}  (7);

  % Edges connecting across hexagons  
  \path (1)  edge [tight edge] node {}  (7);
  \path (2)  edge [tight edge] node {}  (8);
  \path (3)  edge [tight edge] node {}  (9);
  \path (4)  edge [tight edge] node {} (10);
  \path (5)  edge [tight edge] node {} (11);
  \path (6)  edge [tight edge] node {} (12);

\end{tikzpicture}} \fi 

%% file: figures/hexagon_graph_signal_node_3_b.tex
\ifcompiletikz 
{\fontsize{6}{6}\selectfont\begin{tikzpicture}[scale = \myfactor]

  % Define center of hexagon
  \node                         []     (center) {};
  % Nodes in the inner hexagon 
  \path (center) ++ (  0:1.4) node [empty node]   (1) {$1$}  ++ ( 0.5,  0.5) node {$x_1$};
  \path (center) ++ ( 60:1.4) node [node value a] (2) {$2$}  ++ ( 0.6,  0.4) node {$x_2$};
  \path (center) ++ (120:1.4) node [empty node]   (3) {$3$}  ++ (-0.6,  0.4) node {$x_3$};
  \path (center) ++ (180:1.4) node [node value a] (4) {$4$}  ++ (-0.5,  0.5) node {$x_4$};
  \path (center) ++ (240:1.4) node [empty node]   (5) {$5$}  ++ (-0.6, -0.4) node {$x_5$};
  \path (center) ++ (300:1.4) node [empty node]   (6) {$6$}  ++ ( 0.6, -0.4) node {$x_6$};
  % Nodes in the outer hexagon 
  \path (center) ++ (  0:3.0) node [empty node]   (7)  {$7$}  ++ ( 0.4,  0.5) node {$x_7$};
  \path (center) ++ ( 60:3.0) node [node value c] (8)  {$8$}  ++ ( 0.5,  0.5) node {$x_8$};
  \path (center) ++ (120:3.0) node [node value a] (9)  {$9$}  ++ ( 0.5,  0.5) node {$x_9$};
  \path (center) ++ (180:3.0) node [node value c] (10) {$10$} ++ (-0.4,  0.5) node {$x_{10}$};
  \path (center) ++ (240:3.0) node [empty node]   (11) {$11$} ++ (-0.5, -0.5) node {$x_{11}$};
  \path (center) ++ (300:3.0) node [empty node]   (12) {$12$} ++ ( 0.5, -0.5) node {$x_{12}$};

  % Edges connecting inner hexagon  
  \path (1)  edge [tight edge] node {} (2);
  \path (2)  edge [tight edge] node {} (3);
  \path (3)  edge [tight edge] node {} (4);
  \path (4)  edge [tight edge] node {} (5);
  \path (5)  edge [tight edge] node {} (6);
  \path (6)  edge [tight edge] node {} (1);

  % Edges connecting outer hexagon  
  \path (7)   edge [tight edge] node {}  (8);
  \path (8)   edge [tight edge] node {}  (9);
  \path (9)   edge [tight edge] node {} (10);
  \path (10)  edge [tight edge] node {} (11);
  \path (11)  edge [tight edge] node {} (12);
  \path (12)  edge [tight edge] node {}  (7);

  % Edges connecting across hexagons  
  \path (1)  edge [tight edge] node {}  (7);
  \path (2)  edge [tight edge] node {}  (8);
  \path (3)  edge [tight edge] node {}  (9);
  \path (4)  edge [tight edge] node {} (10);
  \path (5)  edge [tight edge] node {} (11);
  \path (6)  edge [tight edge] node {} (12);

\end{tikzpicture}} \fi 

%% file: figures/hexagon_graph_signal_node_3_c.tex
\ifcompiletikz 
{\fontsize{6}{6}\selectfont\begin{tikzpicture}[scale = \myfactor]

  % Define center of hexagon
  \node                         []     (center) {};
  % Nodes in the inner hexagon 
  \path (center) ++ (  0:1.4) node [empty node]   (1) {$1$}  ++ ( 0.5,  0.5) node {$x_1$};
  \path (center) ++ ( 60:1.4) node [node value b] (2) {$2$}  ++ ( 0.6,  0.4) node {$x_2$};
  \path (center) ++ (120:1.4) node [empty node]   (3) {$3$}  ++ (-0.6,  0.4) node {$x_3$};
  \path (center) ++ (180:1.4) node [node value b] (4) {$4$}  ++ (-0.5,  0.5) node {$x_4$};
  \path (center) ++ (240:1.4) node [empty node]   (5) {$5$}  ++ (-0.6, -0.4) node {$x_5$};
  \path (center) ++ (300:1.4) node [empty node]   (6) {$6$}  ++ ( 0.6, -0.4) node {$x_6$};
  % Nodes in the outer hexagon 
  \path (center) ++ (  0:3.0) node [empty node]   (7)  {$7$}  ++ ( 0.4,  0.5) node {$x_7$};
  \path (center) ++ ( 60:3.0) node [node value c] (8)  {$8$}  ++ ( 0.5,  0.5) node {$x_8$};
  \path (center) ++ (120:3.0) node [node value b] (9)  {$9$}  ++ ( 0.5,  0.5) node {$x_9$};
  \path (center) ++ (180:3.0) node [node value c] (10) {$10$} ++ (-0.4,  0.5) node {$x_{10}$};
  \path (center) ++ (240:3.0) node [empty node]   (11) {$11$} ++ (-0.5, -0.5) node {$x_{11}$};
  \path (center) ++ (300:3.0) node [empty node]   (12) {$12$} ++ ( 0.5, -0.5) node {$x_{12}$};

  % Edges connecting inner hexagon  
  \path (1)  edge [tight edge] node {} (2);
  \path (2)  edge [tight edge] node {} (3);
  \path (3)  edge [tight edge] node {} (4);
  \path (4)  edge [tight edge] node {} (5);
  \path (5)  edge [tight edge] node {} (6);
  \path (6)  edge [tight edge] node {} (1);

  % Edges connecting outer hexagon  
  \path (7)   edge [tight edge] node {}  (8);
  \path (8)   edge [tight edge] node {}  (9);
  \path (9)   edge [tight edge] node {} (10);
  \path (10)  edge [tight edge] node {} (11);
  \path (11)  edge [tight edge] node {} (12);
  \path (12)  edge [tight edge] node {}  (7);

  % Edges connecting across hexagons  
  \path (1)  edge [tight edge] node {}  (7);
  \path (2)  edge [tight edge] node {}  (8);
  \path (3)  edge [tight edge] node {}  (9);
  \path (4)  edge [tight edge] node {} (10);
  \path (5)  edge [tight edge] node {} (11);
  \path (6)  edge [tight edge] node {} (12);

\end{tikzpicture}} \fi 

%% file: figures/hexagon_graph_signal_node_6.tex
\ifcompiletikz 
{\fontsize{6}{6}\selectfont\begin{tikzpicture}[scale = \myfactor]

  % Define center of hexagon
  \node                         []     (center) {};
  % Nodes in the inner hexagon 
  \path (center) ++ (  0:1.4) node [node value a] (1) {$1$}  ++ ( 0.5,  0.5) node {$x_1$};
  \path (center) ++ ( 60:1.4) node [empty node]   (2) {$2$}  ++ ( 0.6,  0.4) node {$x_2$};
  \path (center) ++ (120:1.4) node [empty node]   (3) {$3$}  ++ (-0.6,  0.4) node {$x_3$};
  \path (center) ++ (180:1.4) node [empty node]   (4) {$4$}  ++ (-0.5,  0.5) node {$x_4$};
  \path (center) ++ (240:1.4) node [node value a] (5) {$5$}  ++ (-0.6, -0.4) node {$x_5$};
  \path (center) ++ (300:1.4) node [empty node]   (6) {$6$}  ++ ( 0.6, -0.4) node {$x_6$};
  % Nodes in the outer hexagon 
  \path (center) ++ (  0:3.0) node [node value b] (7)  {$7$}  ++ ( 0.4,  0.5) node {$x_7$};
  \path (center) ++ ( 60:3.0) node [empty node]   (8)  {$8$}  ++ ( 0.5,  0.5) node {$x_8$};
  \path (center) ++ (120:3.0) node [empty node]   (9)  {$9$}  ++ ( 0.5,  0.5) node {$x_9$};
  \path (center) ++ (180:3.0) node [empty node]   (10) {$10$} ++ (-0.4,  0.5) node {$x_{10}$};
  \path (center) ++ (240:3.0) node [node value b] (11) {$11$} ++ (-0.5, -0.5) node {$x_{11}$};
  \path (center) ++ (300:3.0) node [node value a] (12) {$12$} ++ ( 0.5, -0.5) node {$x_{12}$};

  % Edges connecting inner hexagon  
  \path (1)  edge [tight edge] node {} (2);
  \path (2)  edge [tight edge] node {} (3);
  \path (3)  edge [tight edge] node {} (4);
  \path (4)  edge [tight edge] node {} (5);
  \path (5)  edge [tight edge] node {} (6);
  \path (6)  edge [tight edge] node {} (1);

  % Edges connecting outer hexagon  
  \path (7)   edge [tight edge] node {}  (8);
  \path (8)   edge [tight edge] node {}  (9);
  \path (9)   edge [tight edge] node {} (10);
  \path (10)  edge [tight edge] node {} (11);
  \path (11)  edge [tight edge] node {} (12);
  \path (12)  edge [tight edge] node {}  (7);

  % Edges connecting across hexagons  
  \path (1)  edge [tight edge] node {}  (7);
  \path (2)  edge [tight edge] node {}  (8);
  \path (3)  edge [tight edge] node {}  (9);
  \path (4)  edge [tight edge] node {} (10);
  \path (5)  edge [tight edge] node {} (11);
  \path (6)  edge [tight edge] node {} (12);

\end{tikzpicture}} \fi 

%% file: figures/random_hexagon_graph_signal_node_3_a.tex
\ifcompiletikz 
{\fontsize{6}{6}\selectfont\begin{tikzpicture}[scale = \myfactor]

  % Define center of hexagon
  \node                         []     (center) {};
  % Nodes in the inner hexagon 
  
  \path (center) ++ (  0:1.4) ++ \randcord node [empty node]   (1) {$1$}  ++ ( 0.5,  0.5) node {$x_1$};
  \path (center) ++ ( 60:1.4) ++ \randcord node [node value a] (2) {$2$}  ++ ( 0.6,  0.4) node {$x_2$};
  \path (center) ++ (120:1.4) ++ \randcord node [node value a] (3) {$3$}  ++ (-0.6,  0.4) node {$x_3$};
  \path (center) ++ (180:1.4) ++ \randcord node [node value a] (4) {$4$}  ++ (-0.5,  0.5) node {$x_4$};
  \path (center) ++ (240:1.4) ++ \randcord node [empty node]   (5) {$5$}  ++ (-0.6, -0.4) node {$x_5$};
  \path (center) ++ (300:1.4) ++ \randcord node [empty node]   (6) {$6$}  ++ ( 0.6, -0.4) node {$x_6$};
  % Nodes in the outer hexagon 
  \path (center) ++ (  0:3.0) ++ \randcord node [empty node]   (7)  {$7$}  ++ ( 0.4,  0.5) node {$x_7$};
  \path (center) ++ ( 60:3.0) ++ \randcord node [node value c] (8)  {$8$}  ++ ( 0.5,  0.5) node {$x_8$};
  \path (center) ++ (120:3.0) ++ \randcord node [node value a] (9)  {$9$}  ++ ( 0.5,  0.5) node {$x_9$};
  \path (center) ++ (180:3.0) ++ \randcord node [node value c] (10) {$10$} ++ (-0.4,  0.5) node {$x_{10}$};
  \path (center) ++ (240:3.0) ++ \randcord node [empty node]   (11) {$11$} ++ (-0.5, -0.5) node {$x_{11}$};
  \path (center) ++ (300:3.0) ++ \randcord node [empty node]   (12) {$12$} ++ ( 0.5, -0.5) node {$x_{12}$};

  % Edges connecting inner hexagon  
  \path (1)  edge [tight edge] node {} (2);
  \path (2)  edge [tight edge] node {} (3);
  \path (3)  edge [tight edge] node {} (4);
  \path (4)  edge [tight edge] node {} (5);
  \path (5)  edge [tight edge] node {} (6);
  \path (6)  edge [tight edge] node {} (1);

  % Edges connecting outer hexagon  
  \path (7)   edge [tight edge] node {}  (8);
  \path (8)   edge [tight edge] node {}  (9);
  \path (9)   edge [tight edge] node {} (10);
  \path (10)  edge [tight edge] node {} (11);
  \path (11)  edge [tight edge] node {} (12);
  \path (12)  edge [tight edge] node {}  (7);

  % Edges connecting across hexagons  
  \path (1)  edge [tight edge] node {}  (7);
  \path (2)  edge [tight edge] node {}  (8);
  \path (3)  edge [tight edge] node {}  (9);
  \path (4)  edge [tight edge] node {} (10);
  \path (5)  edge [tight edge] node {} (11);
  \path (6)  edge [tight edge] node {} (12);

\end{tikzpicture}} \fi 

%% file: figures/random_hexagon_graph_signal_node_6.tex
\ifcompiletikz 
{\fontsize{6}{6}\selectfont\begin{tikzpicture}[scale = \myfactor]

  % Define center of hexagon
  \node                         []     (center) {};
  % Nodes in the inner hexagon 
  \path (center) ++ (  0:1.4) ++ \randcord node [node value a] (1) {$1$}  ++ ( 0.5,  0.5) node {$x_1$};
  \path (center) ++ ( 60:1.4) ++ \randcord node [empty node]   (2) {$2$}  ++ ( 0.6,  0.4) node {$x_2$};
  \path (center) ++ (120:1.4) ++ \randcord node [empty node]   (3) {$3$}  ++ (-0.6,  0.4) node {$x_3$};
  \path (center) ++ (180:1.4) ++ \randcord node [empty node]   (4) {$4$}  ++ (-0.5,  0.5) node {$x_4$};
  \path (center) ++ (240:1.4) ++ \randcord node [node value a] (5) {$5$}  ++ (-0.6, -0.4) node {$x_5$};
  \path (center) ++ (300:1.4) ++ \randcord node [empty node]   (6) {$6$}  ++ ( 0.6, -0.4) node {$x_6$};
  % Nodes in the outer hexagon 
  \path (center) ++ (  0:3.0) ++ \randcord node [node value b] (7)  {$7$}  ++ ( 0.4,  0.5) node {$x_7$};
  \path (center) ++ ( 60:3.0) ++ \randcord node [empty node]   (8)  {$8$}  ++ ( 0.5,  0.5) node {$x_8$};
  \path (center) ++ (120:3.0) ++ \randcord node [empty node]   (9)  {$9$}  ++ ( 0.5,  0.5) node {$x_9$};
  \path (center) ++ (180:3.0) ++ \randcord node [empty node]   (10) {$10$} ++ (-0.4,  0.5) node {$x_{10}$};
  \path (center) ++ (240:3.0) ++ \randcord node [node value b] (11) {$11$} ++ (-0.5, -0.5) node {$x_{11}$};
  \path (center) ++ (300:3.0) ++ \randcord node [node value a] (12) {$12$} ++ ( 0.5, -0.5) node {$x_{12}$};

  % Edges connecting inner hexagon  
  \path (1)  edge [tight edge] node {} (2);
  \path (2)  edge [tight edge] node {} (3);
  \path (3)  edge [tight edge] node {} (4);
  \path (4)  edge [tight edge] node {} (5);
  \path (5)  edge [tight edge] node {} (6);
  \path (6)  edge [tight edge] node {} (1);

  % Edges connecting outer hexagon  
  \path (7)   edge [tight edge] node {}  (8);
  \path (8)   edge [tight edge] node {}  (9);
  \path (9)   edge [tight edge] node {} (10);
  \path (10)  edge [tight edge] node {} (11);
  \path (11)  edge [tight edge] node {} (12);
  \path (12)  edge [tight edge] node {}  (7);

  % Edges connecting across hexagons  
  \path (1)  edge [tight edge] node {}  (7);
  \path (2)  edge [tight edge] node {}  (8);
  \path (3)  edge [tight edge] node {}  (9);
  \path (4)  edge [tight edge] node {} (10);
  \path (5)  edge [tight edge] node {} (11);
  \path (6)  edge [tight edge] node {} (12);

\end{tikzpicture}} \fi 

%% file: figures/graph_perceptron.tex
\ifcompiletikz 
{\footnotesize\begin{tikzpicture}[scale = \myfactor]

  \pgfdeclarelayer{bg}     % declare background layer
  \pgfsetlayers{bg,main}   % set the order of the layers (main is the standard layer)

  % Draw input to GNN
  \node (input) [rectangle, minimum width = 0.1*\unit] {$\bbx$};
  %%%%%%%%%%%%%%%%%%%%%%%%%%%%%%%%%%%%%%%%%%%%%%%%%%
  % Layer 1 %%%%%%%%%%%%%%%%%%%%%%%%%%%%%%%%%%%%%%%%
  %%%%%%%%%%%%%%%%%%%%%%%%%%%%%%%%%%%%%%%%%%%%%%%%%%
  % Draw Layer 1 Filter block and Feature block
  \path (input)      ++ \deltainput node [filter]       (L1 Filter1) {\one};
  \path (L1 Filter1) ++ \deltasigma node [nonlinearity] (L1 F1)      {\sigmaone};
  % Draw connectors for Layer 1
  \path[draw, -stealth] (L1 Filter1.east) -- node [above] {$\bbu$} (L1 F1.west);

  %%%%%%%%%%%%%%%%%%%%%%%%%%%%%%%%%%%%%%%%%%%%%%%%%%
  % Connections across layers %%%%%%%%%%%%%%%%%%%%%%
  %%%%%%%%%%%%%%%%%%%%%%%%%%%%%%%%%%%%%%%%%%%%%%%%%%
  % From input to layer 1
  \path[draw, -stealth] (input.south) -- (L1 Filter1.north);
  % From layer 1 to output
  \path[draw, -stealth] (L1 F1.south) -- ++ \deltaoutput -- ++ (0.3, 0) 
                        node [right]{$\bbz = \bbPhi(\bbx; \bbh, \bbS)$};

  %%%%%%%%%%%%%%%%%%%%%%%%%%%%%%%%%%%%%%%%%%%%%%%%%%
  % Light shades to signify layer groups   %%%%%%%%%
  %%%%%%%%%%%%%%%%%%%%%%%%%%%%%%%%%%%%%%%%%%%%%%%%%%
  \begin{pgfonlayer}{bg} 
      % Shading for Layer 1  
      \path (L1 Filter1.west |- L1 F1.south) ++ (-0.3,-0.3)
           node [filter, anchor = south west,
                 fill = black!5, 
                 minimum width  = 9.6*\unit,
                 minimum height = 1.9*\unit,] 
        (layer)
        {}; 
  \end{pgfonlayer}

\end{tikzpicture}} \fi 

%% file: figures/gnn_block_diagram_single_feature.tex
\ifcompiletikz 
{\footnotesize \begin{tikzpicture}[scale = \myfactor]

  \pgfdeclarelayer{bg}     % declare background layer
  \pgfsetlayers{bg,main}   % set the order of the layers (main is the standard layer)

  % Draw input to GNN
  \node (input) [rectangle, minimum width = 0.1*\unit] {$\bbx$};
  %%%%%%%%%%%%%%%%%%%%%%%%%%%%%%%%%%%%%%%%%%%%%%%%%%
  % Layer 1 %%%%%%%%%%%%%%%%%%%%%%%%%%%%%%%%%%%%%%%%
  %%%%%%%%%%%%%%%%%%%%%%%%%%%%%%%%%%%%%%%%%%%%%%%%%%
  % Draw Layer 1 Filter block and Feature block
  \path (input)      ++ \deltainput node [filter]       (L1 Filter1) {\one};
  \path (L1 Filter1) ++ \deltasigma node [nonlinearity] (L1 F1)      {\sigmaone};
  % Draw connectors for Layer 1
  \path[draw, -stealth] (L1 Filter1.east) -- node [above] {$\bbu_1$} (L1 F1.west);

  %%%%%%%%%%%%%%%%%%%%%%%%%%%%%%%%%%%%%%%%%%%%%%%%%%
  % Layer 2 %%%%%%%%%%%%%%%%%%%%%%%%%%%%%%%%%%%%%%%%
  %%%%%%%%%%%%%%%%%%%%%%%%%%%%%%%%%%%%%%%%%%%%%%%%%%
  % Draw Layer 2 Filter block and Feature block
  \path (L1 Filter1) ++ (0,-\deltalayer) node [filter]       (L2 Filter1) {\two};
  \path (L2 Filter1) ++ \deltasigma      node [nonlinearity] (L2 F1)      {\sigmatwo};
  % Draw connectors for Layer 2
  \path[draw, -stealth] (L2 Filter1.east) --  node [above] {$\bbu_2$} (L2 F1.west);
  
  %%%%%%%%%%%%%%%%%%%%%%%%%%%%%%%%%%%%%%%%%%%%%%%%%%
  % Layer 3 %%%%%%%%%%%%%%%%%%%%%%%%%%%%%%%%%%%%%%%%
  %%%%%%%%%%%%%%%%%%%%%%%%%%%%%%%%%%%%%%%%%%%%%%%%%%
  % Draw Layer 2 Filter block and Feature block
  \path (L2 Filter1) ++ (0,-\deltalayer) node [filter]       (L3 Filter1) {\three};
  \path (L3 Filter1) ++ \deltasigma      node [nonlinearity] (L3 F1)      {\sigmathree};
  % Draw connectors for Layer 2
  \path[draw, -stealth] (L3 Filter1.east) --  node [above] {$\bbu_3$} (L3 F1.west);

  %%%%%%%%%%%%%%%%%%%%%%%%%%%%%%%%%%%%%%%%%%%%%%%%%%
  % Connections across layers %%%%%%%%%%%%%%%%%%%%%%
  %%%%%%%%%%%%%%%%%%%%%%%%%%%%%%%%%%%%%%%%%%%%%%%%%%
  % From input to layer 1
  \path[draw, -stealth] (input.south) -- (L1 Filter1.north);
  % From layer 1 to layer 2
  \path (L1 F1.south) ++ (0,-\deltaconnector) node [] (aux1) {};
  \path[draw, -stealth] (L1 F1.south) -- node [below right] {$\bbx_1$} (aux1.north) 
                                      --                         (aux1.north -| L2 Filter1.north) 
                                      -- node [above left]  {$\bbx_1$} (L2 Filter1.north);
  % From layer 2 to layer 3
  \path (L2 F1.south) ++ (0,-\deltaconnector) node [] (aux1) {};
  \path[draw, -stealth] (L2 F1.south) -- node [below right] {$\bbx_2$} (aux1.north) 
                                      --                         (aux1.north -| L2 Filter1.north) 
                                      -- node [above left]  {$\bbx_2$} (L3 Filter1.north);
  % From layer 3 to output
  \path[draw, -stealth] (L3 F1.south) -- ++ \deltaoutput -- ++ (0.3, 0) 
                        node [right]{$\bbx_3 = \bbPhi(\bbx; \bbH, \bbS)$};

  %%%%%%%%%%%%%%%%%%%%%%%%%%%%%%%%%%%%%%%%%%%%%%%%%%
  % Light shades to signify layer groups   %%%%%%%%%
  %%%%%%%%%%%%%%%%%%%%%%%%%%%%%%%%%%%%%%%%%%%%%%%%%%
  \begin{pgfonlayer}{bg} 
      % Shading for Layer 1  
      \path (L1 Filter1.west |- L1 F1.south) ++ (-0.3,-0.3)
           node [filter, anchor = south west,
                 fill = black!5, 
                 minimum width  = 10.75*\unit,
                 minimum height = 1.9*\unit,] 
        (layer)
        {}; 
      \path (layer.south east) ++ (0.0,0.0) node [above left] {{\fontsize{7}{7}\selectfont Layer 1}};
      % Shading for Layer 2  
      \path (L1 Filter1.west |- L2 F1.south) ++ (-0.3,-0.3)
           node [filter, anchor = south west,
                 fill = black!5, 
                 minimum width  = 10.75*\unit,
                 minimum height = 1.9*\unit,] 
        (layer)
        {}; 
      \path (layer.south east) ++ (0.0,0.0) node [above left] {{\fontsize{7}{7}\selectfont Layer 2}};
      % Shading for Layer 3  
      \path (L1 Filter1.west |- L3 F1.south) ++ (-0.3,-0.3)
           node [filter, anchor = south west,
                 fill = black!5, 
                 minimum width  = 10.75*\unit,
                 minimum height = 1.9*\unit,] 
        (layer)
        {}; 
      \path (layer.south east) ++ (0.0,0.0) node [above left] {{\fontsize{7}{7}\selectfont Layer 3}};

  \end{pgfonlayer}

\end{tikzpicture}} \fi 

%% file: figures/gnn_block_diagram_mimo_filter.tex
\ifcompiletikz 
{\footnotesize\begin{tikzpicture}[scale = \myfactor]

  \pgfdeclarelayer{bg}     % declare background layer
  \pgfsetlayers{bg,main}   % set the order of the layers (main is the standard layer)

  % Draw input to GNN
  \node (input) [rectangle, minimum width = 0.1*\unit] {$\bbx$};
  %%%%%%%%%%%%%%%%%%%%%%%%%%%%%%%%%%%%%%%%%%%%%%%%%%
  % Layer 1 %%%%%%%%%%%%%%%%%%%%%%%%%%%%%%%%%%%%%%%%
  %%%%%%%%%%%%%%%%%%%%%%%%%%%%%%%%%%%%%%%%%%%%%%%%%%
  % Draw Layer 1 Filter block and Feature block
  \path (input)      ++ \deltainput node [filter]       (L1 Filter1) {\one};
  \path (L1 Filter1) ++ \deltasigma node [nonlinearity] (L1 F1)      {\sigmaone};
  % Draw connectors for Layer 1
  \path[draw, -stealth] (L1 Filter1.east) -- node [above] {$\bbU_1$} (L1 F1.west);

  %%%%%%%%%%%%%%%%%%%%%%%%%%%%%%%%%%%%%%%%%%%%%%%%%%
  % Layer 2 %%%%%%%%%%%%%%%%%%%%%%%%%%%%%%%%%%%%%%%%
  %%%%%%%%%%%%%%%%%%%%%%%%%%%%%%%%%%%%%%%%%%%%%%%%%%
  % Draw Layer 2 Filter block and Feature block
  \path (L1 Filter1) ++ (0,-\deltalayer) node [filter]       (L2 Filter1) {\two};
  \path (L2 Filter1) ++ \deltasigma      node [nonlinearity] (L2 F1)      {\sigmatwo};
  % Draw connectors for Layer 2
  \path[draw, -stealth] (L2 Filter1.east) --  node [above] {$\bbU_2$} (L2 F1.west);
  
  %%%%%%%%%%%%%%%%%%%%%%%%%%%%%%%%%%%%%%%%%%%%%%%%%%
  % Layer 3 %%%%%%%%%%%%%%%%%%%%%%%%%%%%%%%%%%%%%%%%
  %%%%%%%%%%%%%%%%%%%%%%%%%%%%%%%%%%%%%%%%%%%%%%%%%%
  % Draw Layer 2 Filter block and Feature block
  \path (L2 Filter1) ++ (0,-\deltalayer) node [filter]       (L3 Filter1) {\three};
  \path (L3 Filter1) ++ \deltasigma      node [nonlinearity] (L3 F1)      {\sigmathree};
  % Draw connectors for Layer 2
  \path[draw, -stealth] (L3 Filter1.east) --  node [above] {$\bbU_3$} (L3 F1.west);

  %%%%%%%%%%%%%%%%%%%%%%%%%%%%%%%%%%%%%%%%%%%%%%%%%%
  % Connections across layers %%%%%%%%%%%%%%%%%%%%%%
  %%%%%%%%%%%%%%%%%%%%%%%%%%%%%%%%%%%%%%%%%%%%%%%%%%
  % From input to layer 1
  \path[draw, -stealth] (input.south) -- (L1 Filter1.north);
  % From layer 1 to layer 2
  \path (L1 F1.south) ++ (0,-\deltaconnector) node [] (aux1) {};
  \path[draw, -stealth] (L1 F1.south) -- node [below right] {$\bbX_1$} (aux1.north) 
                                      --                         (aux1.north -| L2 Filter1.north) 
                                      -- node [above left]  {$\bbX_1$} (L2 Filter1.north);
  % From layer 2 to layer 3
  \path (L2 F1.south) ++ (0,-\deltaconnector) node [] (aux1) {};
  \path[draw, -stealth] (L2 F1.south) -- node [below right] {$\bbX_2$} (aux1.north) 
                                      --                         (aux1.north -| L2 Filter1.north) 
                                      -- node [above left]  {$\bbX_2$} (L3 Filter1.north);
  % From layer 3 to output
  \path[draw, -stealth] (L3 F1.south) -- ++ \deltaoutput -- ++ (0.3, 0) 
                        node [right]{$\bbX_3 = \bbPhi(\bbX; \bbH, \bbS)$};

  %%%%%%%%%%%%%%%%%%%%%%%%%%%%%%%%%%%%%%%%%%%%%%%%%%
  % Light shades to signify layer groups   %%%%%%%%%
  %%%%%%%%%%%%%%%%%%%%%%%%%%%%%%%%%%%%%%%%%%%%%%%%%%
  \begin{pgfonlayer}{bg} 
      % Shading for Layer 1  
      \path (L1 Filter1.west |- L1 F1.south) ++ (-0.3,-0.3)
           node [filter, anchor = south west,
                 fill = black!5, 
                 minimum width  = 10.75*\unit,
                 minimum height = 1.9*\unit,] 
        (layer)
        {}; 
      \path (layer.south east) ++ (0.0,0.0) node [above left] {{\fontsize{7}{7}\selectfont Layer 1}};
      % Shading for Layer 2  
      \path (L1 Filter1.west |- L2 F1.south) ++ (-0.3,-0.3)
           node [filter, anchor = south west,
                 fill = black!5, 
                 minimum width  = 10.75*\unit,
                 minimum height = 1.9*\unit,] 
        (layer)
        {}; 
      \path (layer.south east) ++ (0.0,0.0) node [above left] {{\fontsize{7}{7}\selectfont Layer 2}};
      % Shading for Layer 3  
      \path (L1 Filter1.west |- L3 F1.south) ++ (-0.3,-0.3)
           node [filter, anchor = south west,
                 fill = black!5, 
                 minimum width  = 10.75*\unit,
                 minimum height = 1.9*\unit,] 
        (layer)
        {}; 
      \path (layer.south east) ++ (0.0,0.0) node [above left] {{\fontsize{7}{7}\selectfont Layer 3}};

  \end{pgfonlayer}

\end{tikzpicture}} \fi 

%% file: FG-stability.tex
% !TEX root = root.tex

%%%%%%%%%%%%%%%%%%%%%%%%%%%%%%%%%%%%%%%%%%%%%%%%%%%%%%%%%%%%%%%%%%%%%%%%%%%%%%%%
%%%%                                                                        %%%%
%%%%                           T E M P O R A R Y                            %%%%
%%%%                                                                        %%%%
%%%%%%%%%%%%%%%%%%%%%%%%%%%%%%%%%%%%%%%%%%%%%%%%%%%%%%%%%%%%%%%%%%%%%%%%%%%%%%%%

%%%%% LABEL EQUIVALENCE %%%%%%
\def\eqGraphFilterSingleFeature{\eqref{eqn_ch8_perceptron_graph_filter}}
\def\eqERMparam{\eqref{eqn_ch8_erm_learning}}
\def\eqNonlinearity{\eqref{eqn_ch8_pointwise_nonlinearity}}
\def\eqGraphPerceptron{\eqref{eqn_ch8_perceptron_nonlinearity}}
\def\eqMultiLayerSingleFeatureLinear{\eqref{eqn_ch8_gnn_recursion_single_feature_filter}}
\def\eqMultiLayerSingleFeatureNonlinear{\eqref{eqn_ch8_gnn_recursion_single_feature_nonlinearity}}
\def\eqGNNsingleFeature{\eqref{eqn_ch8_gnn_operator_single_feature}}
\def\eqGraphSignal{\eqref{eqn_ch8_gnn_features}} % MutiFeature
\def\eqGraphFilter{\eqref{eqn_ch8_gnn_recursion_filter_matrix}} % MultiFeature
\def\eqGraphFilterSingleFilter{\eqref{eqn_ch8_gnn_filter}}
\def\eqGNNlayer{\eqref{eqn_ch8_gnn_recursion_nonlinearity_matrix}}
\def\eqGNN{\eqref{eqn_ch8_gnn_operator_multifeature}}
\def\eqRatingSimilarity{\eqref{eqn_reco_systems_weights}}
\def\secRecSysDescription{\ref{sec_reco_systems}}
\def\secRecSysResults{\ref{sec_reco_systems_results}}
\def\propPermEquivGraphFilter{\ref{prop_filter_equivariance}}
\def\propPermEquivGNN{\ref{prop_gnn_equivariance}}
\def\figPermEquiv{\ref{fig_generalization}}
\def\figWiggly{\ref{fig_generalization_with_wiggliness}}
%%%%%%%%%%%%%%%%%%%%%%%%%%%%%%

%%%%% OTHER DEFINITIONS %%%%%%
\definecolor{pennpurple}{cmyk}{0,1,0.11,0.71} 
\newcommand{\purple}[1]{{\color{pennpurple} #1}}
%%%%%%%%%%%%%%%%%%%%%%%%%%%%%%

%%%%%%%%% REMINDERS %%%%%%%%%%
% Phi(x; \ccalH) is a general parametrization
% Phi(x; \bbh, \bbS) is a graph *convolutional* filter: order K
%%%%%%%%%%%%%%%%%%%%%%%%%%%%%%

%%%% WHAT WE ALREADY KNOW %%%%
% Graph perceptron (and GNNs) are nonlinear processing
% Graph perceptron retains locality of graph filters
% Representation: Graph filters \subset Graph perceptron \subset Multi layer GNN
%%%%%%%%%%%%%%%%%%%%%%%%%%%%%%

%%%%%%%%%%%%%%%%%%%%%%%%%%%%%%%%%%%%%%%%%%%%%%%%%%%%%%%%%%%%%%%%%%%%%%%%%%%%%%%%
%%%%                                                                        %%%%
%%%%                             S E C T I O N                              %%%%
%%%%                                                                        %%%%
%%%%%%%%%%%%%%%%%%%%%%%%%%%%%%%%%%%%%%%%%%%%%%%%%%%%%%%%%%%%%%%%%%%%%%%%%%%%%%%%

\section{Stability Properties of GNNs} \label{sec:stability}

Permutation equivariance is a fundamental property of graph filters (Prop.~\propPermEquivGraphFilter) and GNNs (Prop.~\propPermEquivGNN), since it allows them to exploit the graph structure and thus generalize better to unseen samples coming from the same graph \cite{Gama19-Stability, ZouLerman19-Scattering}. However, graphs rarely exhibit perfect symmetries as illustrated in Fig.~\figPermEquiv, but rather show near permutation symmetries, as seen in Fig.~\figWiggly.

Stability to graph support perturbations quantifies how much the output of the graph filter changes in relation to the size of the perturbation. That is, if the graph support has changed slightly (with respect to a permutation of itself), then the output of a trained graph filter or GNN will also change slightly \cite{Gama19-Stability}. This property is particularly important in graph data where the structure of the graph, described by $\bbS$, is generally given in the problem and might not be known precisely \cite{pasdeloup2019uncertainty}. {For example, in the problem of movie recommendation (Sec.~\secRecSysDescription), the edges of the graph are built based on the rating similarity between the items [cf. \eqRatingSimilarity]. Estimating this value depends on the training set and thus there is an error incurred in obtaining it. Therefore, we usually train over an inferred graph that is not exactly the true graph over which the data is actually defined. The stability property guarantees that the trained parametrization (either a graph filter or a GNN) will yield the expected performance as long as the estimation of the support is good enough \cite{Gama19-Stability}.}

{In this section, we present the stability property of graph filters and GNNs for a relative perturbation model (Sec.~\ref{subsec:relative}). Stability is thus another fundamental property that complements permutation equivariance, establishing the mechanisms by which graph filters and GNNs adequately exploit the graph structure to offer better generalization capabilities.}

Both permutation equivariance and stability are properties shared by graph fiters and GNNs, and thus they explain their superior performance with respect to arbitrary linear transforms or FCNNs, as observed in the recommendation problem (Sec.~\ref{sec_reco_systems_results}). In this example, we further observe that GNNs perform better than graph filters. Herein, we leverage the stability theorems and the effect of nonlinearities to explain why GNNs perform better than graph filters. We show that nonlinearities have a demodulating effect on the frequency domain that allows GNNs to be simultaneously stable and discriminative, a feat that cannot be achieved by graph filters (Sec.~\ref{subsec:discussion}).

{In what follows, we focus on undirected graphs and parametrizations given either by graph convolutional filters with $F$ input features and $G$ output features [cf. \eqGraphFilter] or by GNNs [cf. \eqGNN]. We consider GNNs that satisfy the following assumptions.}
%
%%%%%%%%%%%%%%%%%%%%%%%%%%%%%%%%%%%%%%%%
%%%%            ASSUMPTION          %%%%
%%%%%%%%%%%%%%%%%%%%%%%%%%%%%%%%%%%%%%%%
%
\begin{assumption}[GNN architecture] \label{asn:GNNarchit}
    Let $\Phi$ be a GNN parametrization \eqGNN\ with the following architecture.
    \begin{enumerate}[(i)]
        \item Consists of $L>0$ layers.
        \item Obtains $F_{l}$ features at the output of each layer.
        \item The graph filters [cf. \eqGraphFilter] are described by the tensor of coefficients $\bbH=\{\bbH_{lk}\}_{l,k}$, with $\bbH_{lk} \in \reals^{F_{l-1} \times F_{l}}$.
        \item The output of the filtering stage of each layer $l$ satisfies $\|\bbU_{l}\| \leq B \| \bbX_{l-1}\|$ [cf. \eqGraphFilter] for some $B > 0$.
        \item The chosen nonlinearity $\sigma$ is normalized Lipschitz continuous, $|\sigma(a) - \sigma(b)| \leq |a-b|$ for $a,b \in \reals$, and satisfies $\sigma(0)=0$.
    \end{enumerate}
\end{assumption}
%
%%
%%%%        End of ASSUMPTION       %%%%
%%%%%%%%%%%%%%%%%%%%%%%%%%%%%%%%%%%%%%%%
%
We note that Assumption~\ref{asn:GNNarchit} is made on the resulting trained GNN. Assumptions (i)-(iii) are determined by the hyperparameters of the architecture and, as such, are a design choice. {Assumption (iv) needs to be satisfied only on some finite interval $[\lambda_{\min},\lambda_{\max}]$ and is always the case, in theory, for graph convolutional filters \eqGraphFilter\ with finite coefficients. In practical terms, some choices of $\bbS$ may lead to numerical instabilities when computing $\bbS^{k}$. There are several ways  to address this as discussed in \cite{Kipf17-GCN}}. Assumption (v) is satisfied by most of the commonly chosen nonlinearities ($\tanh$, $\text{ReLU}$, $\text{sigmoid}$).

\subsection{Relative perturbations} \label{subsec:relative}

{Permutations are a very particular case of a modification or \emph{perturbation} to which the graph support $\bbS$ can be subjected (see Fig.~\figPermEquiv). We are interested, however, in more general perturbations $\hbS$ (see Fig.~\figWiggly), and in analyzing how the parametrization $\Phi$ changes under these perturbations of the graph support. To measure the change in the parametrization, and in light of the permutation equivariance property of Propositions~\propPermEquivGraphFilter~and~\propPermEquivGNN, we define the operator distance modulo permutations.}
%
%%%%%%%%%%%%%%%%%%%%%%%%%%%%%%%%%%%%%%%%
%%%%            DEFINITION          %%%%
%%%%%%%%%%%%%%%%%%%%%%%%%%%%%%%%%%%%%%%%
%
\begin{definition}[Operator distance modulo permutations] \label{def:opDist}
    Let $\bbS$ be the support matrix of a graph $\bbG$, and let $\hbS$ be the support matrix of a perturbed graph $\hat{\bbG}$. Let $\bbH$ be the tensor of filter coefficients that describe the parametrization $\Phi$ [cf. \eqGraphFilter\ or \eqGNN]. Then, the \emph{operator distance modulo permutation} is defined as
    % eq:opDist
    \begin{equation} \label{eq:opDist}
    \begin{aligned}
    & \big\| \Phi( \cdot ; \bbH, \bbS) - \Phi( \cdot ; \bbH, \hbS) \big\|_{\ccalP} \\
    & = \min_{\bbP \in \ccalP} \max_{\bbX : \|\bbX\|=1} \| \Phi( \bbX ; \bbH, \bbS) - \Phi( \bbX ; \bbH, \bbP^{\Tr} \hbS \bbP) \|
    \end{aligned}
    \end{equation}
    where, for any $\bbU \in \reals^{n \times G}$, we define $\|\bbU\| = \sum_{g=1}^{G} \|\bbu^{g}\|_{2}$.
\end{definition}
%
%%
%%%%        End of DEFINITION       %%%%
%%%%%%%%%%%%%%%%%%%%%%%%%%%%%%%%%%%%%%%%
%
{We note that $\ccalP$ denotes the set of all possible permutations
% eq:permutationSet
\begin{equation} \label{eq:permutationSet}
\ccalP = \big\{ \bbP \in \{0,1\}^{n \times n} : \bbP \bbone = \bbone \ , \ \bbP^{\Tr} \bbone = \bbone \big\}.
\end{equation}
The operator distance modulo permutations measures how much the output of the parametrization $\Phi$ changes for a unit-norm signal $\bbX$ that makes the difference maximum, and for a permutation that makes the difference minimum. Note that, in terms of the operator distance in Def.~\ref{def:opDist}, the permutation equivariance property (Proposition~\propPermEquivGraphFilter~and~\propPermEquivGNN) implies that
% eq:permEquivOpDist
\begin{equation}\label{eq:permEquivOpDist}
    \big\| \Phi( \cdot ; \bbH, \bbS) - \Phi( \cdot ; \bbH, \bbP^{\Tr} \bbS \bbP) \big\|_{\ccalP} = 0
\end{equation}
for both graph filters and GNN parametrizations of $\Phi$.}

{To better analyze how the output of the parametrization $\Phi$ changes when the underlying graph is perturbed, we proceed in the graph frequency domain, as is customary in signal processing. To do this, we consider the eigendecomposition of the support matrix $\bbS=\bbV \bbLambda \bbV^{\Tr}$ to be given by an orthonormal set of eigenvectors collected in the columns of $\bbV$. We define the graph Fourier transform (GFT) of a graph signal $\bbX$ as a projection of the signal onto the eigenvectors of the support matrix $\bbS$ \cite{Sandryhaila13-DSPG, Shuman13-SPG, Sandryhaila14-DSPGfreq,ricaud2019fourier}
% eq:GFT
\begin{equation} \label{eq:GFT}
    \tbX = \bbV^{\Tr} \bbX.
\end{equation}
Note that, since $\bbV$ is an orthonormal matrix, then the inverse GFT is immediately defined as $\bbX = \bbV \tbX$.}

%%%%%%%%%%%%%%%%%%%%%%%%%%%%%%%%%%%%%%%%
%%%%             FIGURE             %%%%
%%%%%%%%%%%%%%%%%%%%%%%%%%%%%%%%%%%%%%%%
%
\def \thisplotscale {2.9}
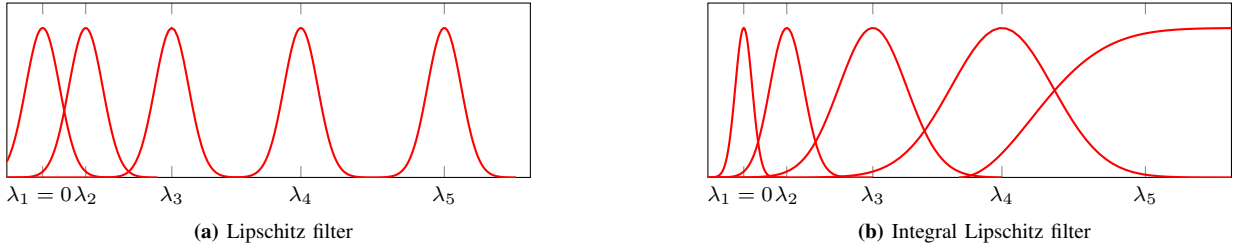
\begin{figure*}
    \centering
    \begin{subfigure}{0.4\textwidth}
        \centering
        \input{plots_stability/lipschitz_filters.tex}
        \vspace{-0.1cm}
        \caption{Lipschitz filter}
        \label{subfig:LipschitzFilters}
    \end{subfigure}
    \hspace{1.5cm}
    \begin{subfigure}{0.4\textwidth}
        \centering
        \input{plots_stability/integral_lipschitz_filters.tex}
        \vspace{-0.1cm}
        \caption{Integral Lipschitz filter}
        \label{subfig:integralLipschitzFilters}
    \end{subfigure}
    \caption{Frequency response (Def.~\ref{def:freqResponse}) of bank of graph filters [cf. \eqGraphFilter]. { (\subref{subfig:LipschitzFilters}) Lipschitz filter with $F=1$ input feature and $G=5$ output features. The frequency response of a Lipschitz filter has $5$ functions of the form \eqref{eq:freqResponse} and all satisfy Lipschitz continuity $|h^{fg}(\lambda_{1}) - h^{fg}(\lambda_{2})| \leq C |\lambda_{1}-\lambda_{2}|$. In this illustrative plot, this condition is met exactly. The minimum width of the functions \eqref{eq:freqResponse} is determined by $C$ since this value limits the maximum value of the derivative. The minimum width is the same throughout the spectrum.} (\subref{subfig:integralLipschitzFilters}) Integral Lipschitz filter (Def.~\ref{def:integralLipschitzFiters}) with $F=1$ input feature and $G=5$ output features. The frequency response of an integral Lipschitz filter has $5$ functions of the form \eqref{eq:freqResponse} and all satisfy \eqref{eq:integralLipschitzFilters}. In this plot, this condition is met exactly. The minimum width of the functions \eqref{eq:freqResponse} depends on their location in the spectrum, since the maximum value of the derivative is bounded by $2C/|\lambda_{1}+\lambda_{2}|$. Therefore, filters located in smaller eigenvalues (i.e. $\lambda_{1}$) can be narrower than filters located in larger eigenvalues (i.e. $\lambda_{5}$).}
    \label{fig:filters}
\end{figure*}
%%
%%%%          End of FIGURE         %%%%
%%%%%%%%%%%%%%%%%%%%%%%%%%%%%%%%%%%%%%%%

{With this definition in place, we can compute the GFT of the graph filter output $\bbU = \sum_{k=0}^{\infty} \bbS^{k} \bbX \bbH_{k}$ [cf. \eqGraphFilter] as \cite{Segarra17-Linear}
% eq:GFTfilter
\begin{equation}
    \tbU = \bbV^{\Tr} \bbU = \sum_{k=0}^{\infty} \bbLambda^{k} \tbX \bbH_{k}
\end{equation}
where, due to the diagonal nature of $\bbLambda$, we can obtain the GFT as a pointwise multiplication in the graph frequency domain, akin to the convolution theorem \cite[Sec. 2.9.6]{OppenheimSchafer10-DTSP}, \cite{Sandryhaila14-DSPGfreq, Ortega18-GSP}. To see this more clearly, consider the $i$th frequency component of $\bbU$ for the $g$th feature, that is, the element $(i,g)$ of $\tbU$ which we denote as $[\tbU]_{ig} = \tdu_{i}^{g}$. Then, we note that
% eq:GFTfilterSingle
\begin{equation} \label{eq:GFTfilterSingle}
    \tdu_{i}^{g} = \sum_{f=1}^{F} h^{fg}(\lambda_{i}) \tdx_{i}^{f}
\end{equation}
for $\tdx_{i}^{f}$ the $i$th frequency component of the $f$th feature of the input, and where $h^{fg}(\lambda_{i})$ is the frequency response of the $(f,g)$ graph convolutional filter in \eqGraphFilter, evaluated at $\lambda_{i}$. We formally define the frequency response of a graph filter [cf. \eqGraphFilter].}
%
%%%%%%%%%%%%%%%%%%%%%%%%%%%%%%%%%%%%%%%%
%%%%            DEFINITION          %%%%
%%%%%%%%%%%%%%%%%%%%%%%%%%%%%%%%%%%%%%%%
%
\begin{definition}[Graph filter frequency response] \label{def:freqResponse}
    Given a graph filter [cf. \eqGraphFilter] with a tensor of filter coefficients $\bbH = \{\bbH_{k}\}_{k}$, $\bbH_{k} \in \reals^{F \times G}$, the frequency response of the graph filter is the set of $F \times G$ polynomial functions $h^{fg}(\lambda)$, with
    % eq:opDist
    \begin{equation} \label{eq:freqResponse}
        h^{fg}(\lambda) = \sum_{k=0}^{K} h_{k}^{fg} \lambda^{k}
    \end{equation}
    for a continuous variable $\lambda$, and where $h_{k}^{fg} = [\bbH_{k}]_{fg}$ is the $(f,g)$th element of $\bbH_{k}$, corresponding to the $k$th filter coefficient of the $(f,g)$ graph convolutional filter in the corresponding filterbank.
%[cf. \eqGraphFilterSingleFilter].
\end{definition}
%
%%
%%%%        End of DEFINITION       %%%%
%%%%%%%%%%%%%%%%%%%%%%%%%%%%%%%%%%%%%%%%
%
{Per Def.~\ref{def:freqResponse}, the frequency response of a filter is a collection of polynomial functions characterized solely by the filter coefficients and so it is independent of the graph. The effect of the specific support matrix $\bbS$ on a graph filter is observed by instantiating the frequency response on the specific eigenvalues [cf. \eqref{eq:GFTfilterSingle}]. But the shape of the frequency response is actually independent of the graph and determined by the filter coefficients.}

{It is evident from \eqref{eq:GFTfilterSingle} that the GFT of the output of a graph filter is a pointwise multiplication of the GFT of the input and the frequency response of the filter. An important distinction with traditional signal processing, however, is that the GFT of a signal depends on the eigenvectors of the support matrix $\bbS$ and the GFT of a filter depends on the eigenvalues of $\bbS$ \cite{Sandryhaila14-DSPGfreq}, while in traditional SP the FT of both the signal and the filter only depend on the eigenvalues $e^{-j2\pi n/N}$.}
%is that the GFT of a signal depends on the eigenvectors of the support matrix $\bbS$, while the GFT of a filter depends on the eigenvalues of $\bbS$ \cite{Sandryhaila14-DSPGfreq}.}

{We are particularly interested in filters that satisfy the integral Lipschitz condition. While traditional Lipschitz filters are those whose frequency response is Lipschitz continuous \cite[Def. 2]{Gama19-Stability}, \emph{integral} Lipschitz filters are those that are Lipschitz continuous, but with a constant that depends on the midpoint of the values considered. See Fig.~\ref{fig:filters} for an illustrative comparison between Lipschitz filters and integral Lipschitz filters. We formally define integral Lipschitz filters as follows.}
%
%%%%%%%%%%%%%%%%%%%%%%%%%%%%%%%%%%%%%%%%
%%%%            DEFINITION          %%%%
%%%%%%%%%%%%%%%%%%%%%%%%%%%%%%%%%%%%%%%%
%
\begin{definition}[Integral Lipschitz graph filters] \label{def:integralLipschitzFiters}
    Given a filter [cf. \eqGraphFilter] with a tensor of filter coefficients $\bbH = \{\bbH_{k}\}_{k}$ with $\bbH_{k} \in \reals^{F \times G}$, we say it is an \emph{integral Lipschitz graph filter} if its frequency response [cf. Def.~\ref{def:freqResponse}] satisfies
    % eq:opDist
    \begin{equation} \label{eq:integralLipschitzFilters}
    |h^{fg}(\lambda_{1}) - h^{fg}(\lambda_{2})| \leq  \frac{C}{|\lambda_{1}+\lambda_{2}|/2} |\lambda_{1}-\lambda_{2}|
    \end{equation}
    for some $C>0$, and for all $\lambda_{1},\lambda_{2} \in \reals, \lambda_{1} \neq \lambda_{2}$ and all $f=1,\ldots,F$ and $g=1,\ldots,G$.
\end{definition}
%
%%
%%%%        End of DEFINITION       %%%%
%%%%%%%%%%%%%%%%%%%%%%%%%%%%%%%%%%%%%%%%
%
Integral Lipschitz filters (Def.~\ref{def:integralLipschitzFiters}) are those filters whose frequency response (Def.~\ref{def:freqResponse}) is Lipschitz continuous on continuous variable $\lambda$ with a Lipschitz constant that is inversely proportional to the midpoint of the interval. For example, if $\lambda_{1}$ or $\lambda_{2}$ are large, the resulting Lipschitz constant $2C/(\lambda_{1}+\lambda_{2})$ is small. This implies that these filters need to be flat for large values of $\lambda$ (i.e. they do not change), but can be arbitrarily thin for values of $\lambda$ near zero (i.e. they can change arbitrarily). See Fig.~\ref{subfig:integralLipschitzFilters} for an example of an illustration of the frequency response of a graph filter that satisfies the integral Lipschitz condition. Note that \eqref{eq:integralLipschitzFilters} implies $|\lambda (h^{fg}(\lambda))'| \leq C$ for $(h^{fg}(\lambda))'$ being the derivative of $h^{fg}(\lambda)$. {This condition is reminiscent of the scale invariance of wavelet filter banks \cite{Daubechies92-Wavelets} and there are several graph wavelet banks that satisfy it, see \cite{Hammond11-Wavelets, Shuman15-Wavelets}.}

{To measure the distance between a graph $\bbS$ and its corresponding perturbation $\hbS$, we adopt a \emph{relative perturbation} model, which ties the changes of the graph to the underlying structure.}
%
%%%%%%%%%%%%%%%%%%%%%%%%%%%%%%%%%%%%%%%%
%%%%            DEFINITION          %%%%
%%%%%%%%%%%%%%%%%%%%%%%%%%%%%%%%%%%%%%%%
%
\begin{definition}[Relative perturbations] \label{def:relPerturbation}
    Given a support matrix $\bbS$ and a perturbed support $\hbS$, define the \emph{relative error set} as
    % eq:absErrorSet
    \begin{equation} \label{eq:relErrorSet}
    \begin{aligned}
    \ccalE(\bbS,\hbS) = \Big\{\bbE \in \reals^{n \times n} : &\ \bbP^{\Tr} \hbS \bbP = \bbS+ \frac{1}{2} (\bbS\bbE+\bbE\bbS) \ , \\ &\ \bbP \in \ccalP \ ,\  \bbE = \bbE^{\Tr}
    \Big\}.
    \end{aligned}
    \end{equation}
    The size of the \emph{relative perturbation} is
    % eq:absPerturbation
    \begin{equation} \label{eq:relPerturbation}
    d(\bbS,\hbS) = \min_{\bbE \in \ccalE(\bbS,\hbS)} \| \bbE \|.
    \end{equation}
\end{definition}
%
%%
%%%%        End of DEFINITION       %%%%
%%%%%%%%%%%%%%%%%%%%%%%%%%%%%%%%%%%%%%%%
%
The relative error set \eqref{eq:relErrorSet} is defined as the set of all symmetric error matrices $\bbE$ such that, when multiplied by the shift operator and added back to it, yield a permutation of the perturbed support $\hbS$. The relative perturbation size \eqref{eq:relPerturbation} is given by the minimum norm of all such relative error matrices, and thus measures how close $\bbS$ and $\hbS$ are to being permutations of each other, as determined by the multiplicative factor $\bbE$.

%%%%%%%%%%%%%%%%%%%%%%%%%%%%%%%%%%%%%%%%
%%%%             FIGURE             %%%%
%%%%%%%%%%%%%%%%%%%%%%%%%%%%%%%%%%%%%%%%
%
\def \thisplotscale {2.9}
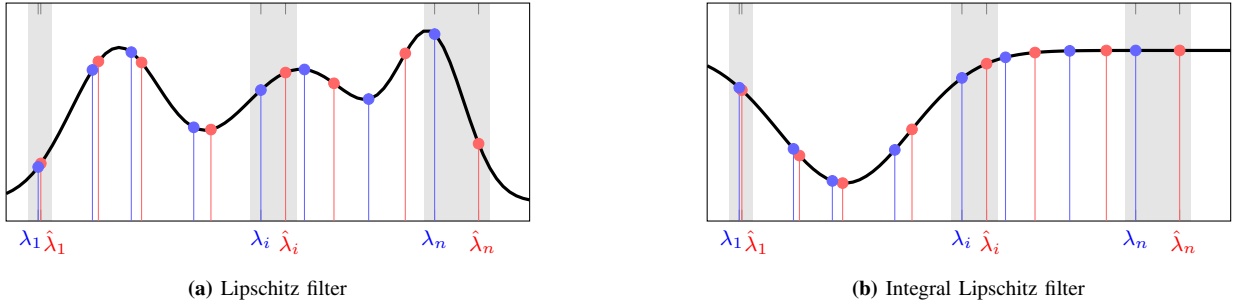
\begin{figure*}
    \centering
    \begin{subfigure}{0.4\textwidth}
        \centering
        \input{plots_stability/graph_dilation_lipschitz.tex}
        \caption{Lipschitz filter}
        \label{subfig:graphDilationLipschitz}
    \end{subfigure}
    \hspace{1.5cm}
    \begin{subfigure}{0.4\textwidth}
        \centering
        \input{plots_stability/graph_dilation_integral_lipschitz.tex}
        \caption{Integral Lipschitz filter}
        \label{subfig:graphDilationIntegralLipschitz}
    \end{subfigure}
    \caption{Effect of a graph dilation $\hbS = (1+\varepsilon) \bbS$. The eigenvalues move from $\lambda_{i}$ (in blue) to $\hat{\lambda}_{i} = (1+\varepsilon) \lambda_{i}$ (in red). Even if $\varepsilon \approx 0$, large eigenvalues change more than small eigenvalues. (\subref{subfig:graphDilationLipschitz}) Lipschitz filters are not stable. A small perturbation causes a large change in the output of the filter due to the large change in large eigenvalues. (\subref{subfig:graphDilationIntegralLipschitz}) Integral Lipschitz filters are stable. For small eigenvalues, the filter can change, but the eigenvalues do not change much. For large eigenvalues, the filter is flat, and thus the large change in eigenvalues still yields the same output.}
    \label{fig:graphDilation}
\end{figure*}
%%
%%%%          End of FIGURE         %%%%
%%%%%%%%%%%%%%%%%%%%%%%%%%%%%%%%%%%%%%%%

{The relative perturbation model takes into consideration the structure of the graph when measuring the change in the support by tying the changes in the edge weights of the graph to its local structure. To see this, note that the difference between the edge weight $[\bbS]_{ij}$ of the original graph $\bbS$ and the corresponding edge $[\bbP_0^{\Tr}\hbS\bbP_0]_{ij}$ of the perturbed graph $\hbS$ is given by the corresponding entry $[\bbE\bbS+\bbS\bbE]_{ij}$ of the perturbation factor $\bbE\bbS+\bbS\bbE$. It is ready to see that this quantity is proportional to the sum of the degrees of nodes $i$ and $j$ scaled by the entries of $\bbE$. As the norm of $\bbE$ grows, the entries of the graphs $\bbS$ and $\bbP_0^{\Tr}\hbS\bbP_0$ become more dissimilar. But parts of the graph that are characterized by weaker connectivity change by amounts that are proportionally smaller to the changes that are observed in parts of the graph characterized by stronger links. This is in contrast to absolute perturbations where edge weights change by the same amount irrespective of the local topology of the graph.}

{Relative perturbations arise in many practical problems and, as a matter of fact, the diffeomorphism used in the seminal work by \cite{Mallat12-Scattering} can be modeled a relative perturbation, since each point in the Euclidean space is perturbed depending on the position of the point (i.e. it takes into account the original structure of the space). Most notable, though, is the case of covariance-based graphs, where the edge weights are a function of the correlation between the nodes. We typically estimate this correlation from a given dataset and this estimation incurs an error that is proportional to the true value of the correlation \cite{Wishart28-CovarianceError, Bishop06-PAML}. Thus, the relationship between the estimate $\hbS$ and the true graph $\bbS$ follows the relative perturbation model. We note that this is precisely the case in the problem of movie recommendation (Sec.~\secRecSysDescription), where perturbations arising from imperfect estimation of the rating similarities \eqRatingSimilarity\ fall under the relative perturbation model.}

{Integral Lipschitz filters (Def.~\ref{def:integralLipschitzFiters}) are stable to relative perturbations (Def.~\ref{def:relPerturbation}) per the following theorem \cite[Theorem 2]{Gama19-Stability}.}
%
%%%%%%%%%%%%%%%%%%%%%%%%%%%%%%%%%%%%%%%%
%%%%             THEOREM            %%%%
%%%%%%%%%%%%%%%%%%%%%%%%%%%%%%%%%%%%%%%%
%
\begin{theorem}[Graph filter stability to relative perturbations] \label{thm:relStabilityGraphFilter}
    Let $\bbS$ and $\hbS$ be the support matrices of a graph $\bbG$ and its perturbation $\hat{\bbG}$, respectively. Let $\Phi$ be a graph filter [cf. \eqGraphFilter] with a tensor of filter coefficients $\bbH = \{\bbH_{k}\}_{k}$, $\bbH_{k} \in \reals^{F \times G}$. If $\Phi$ is an integral Lipschitz filter (Def.~\ref{def:integralLipschitzFiters}) with $C>0$ and if the relative perturbation size satisfies $d(\bbS,\hbS) \leq \varepsilon$ (Def.~\ref{def:relPerturbation}), then
    % eq:absStabilityGraphFilter
    \begin{equation} \label{eq:relStabilityGraphFilter}
    \big\| \bbPhi(\cdot;\bbH,\bbS) - \bbPhi(\cdot;\bbH,\hbS) \big\|_{\ccalP} \leq \varepsilon (1+\delta \sqrt{n}) CG + \ccalO(\varepsilon^{2})
    \end{equation}
    where $\delta = (\|\bbU-\bbV\|_{2} + 1)^{2}-1$ is the \emph{eigenvector misalignment constant} for $\bbU$ the eigenvector basis of the absolute error matrix $\bbE$ that solves \eqref{eq:relPerturbation}.
\end{theorem}
%
%%
%%%%         End of THEOREM         %%%%
%%%%%%%%%%%%%%%%%%%%%%%%%%%%%%%%%%%%%%%%
%
Theorem~\ref{thm:relStabilityGraphFilter} asserts that a change in the output of a graph filter caused by a relative perturbation of the graph support is upper bounded in proportion to the size of the perturbation \eqref{eq:relPerturbation}. This property of stability to relative perturbations is inherited by GNNs as is shown next \cite[Theorem 4]{Gama19-Stability}.
%
%%%%%%%%%%%%%%%%%%%%%%%%%%%%%%%%%%%%%%%%
%%%%             THEOREM            %%%%
%%%%%%%%%%%%%%%%%%%%%%%%%%%%%%%%%%%%%%%%
%
\begin{theorem}[GNN stability to relative perturbations] \label{thm:relStabilityGNN}
    Let $\bbS$ and $\hbS$ be the support matrices of a graph $\bbG$ and its perturbation $\hat{\bbG}$, respectively. Let $\Phi$ be a GNN [cf. \eqGNN] that satisfies Assumption~\ref{asn:GNNarchit}. If the filters used in $\Phi$ are integral Lipschitz (Def.~\ref{def:integralLipschitzFiters}) with $C>0$ and if the relative perturbation size satisfies $d(\bbS,\hbS) \leq \varepsilon$ (Def.~\ref{def:relPerturbation}), then
    % eq:absStabilityGraphFilter
    \begin{equation} \label{eq:relStabilityGNN}
    \big\| \bbPhi(\cdot;\bbH,\bbS) - \bbPhi(\cdot;\bbH,\hbS) \big\|_{\ccalP} \leq \varepsilon (1+\delta \sqrt{n}) CB^{L-1}\prod_{l=1}^{L} F_{l} + \ccalO(\varepsilon^{2})
    \end{equation}
    where $\delta = (\|\bbU-\bbV\|_{2} + 1)^{2}-1$ is the \emph{eigenvector misalignment constant} for $\bbU$ the eigenvector basis of the relative error matrix $\bbE$ that solves \eqref{eq:relPerturbation}.
\end{theorem}
%
%%
%%%%         End of THEOREM         %%%%
%%%%%%%%%%%%%%%%%%%%%%%%%%%%%%%%%%%%%%%%
%

Theorem~\ref{thm:relStabilityGNN} states that the change in the output of the GNN caused by a relative perturbation of the graph support is upper bounded in a proportional manner to the size of the perturbation \eqref{eq:relPerturbation}. Theorem~\ref{thm:relStabilityGNN} thus complements Theorem~\ref{thm:relStabilityGraphFilter}, quantifying how the stability of graph filters gets inherited by GNNs.

%%%%%%%%%%%%%%%%%%%%%%%%%%%%%%%%%%%%%%%%
%%%%             FIGURE             %%%%
%%%%%%%%%%%%%%%%%%%%%%%%%%%%%%%%%%%%%%%%
%
\def \thisplotscale {2.9}
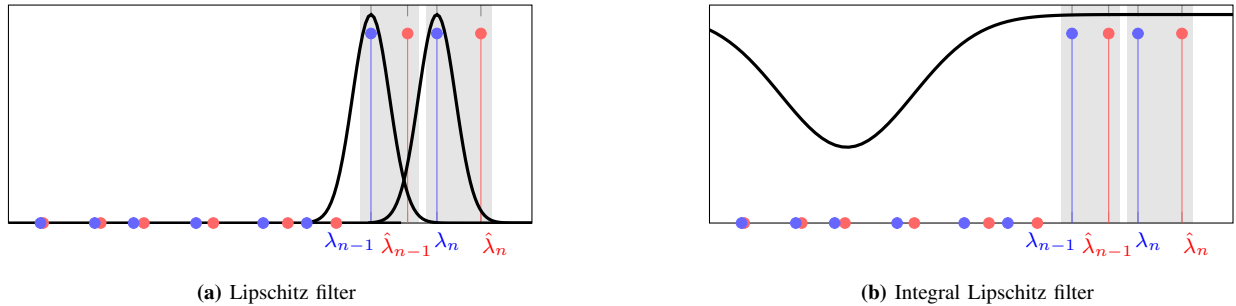
\begin{figure*}
    \centering
    \begin{subfigure}{0.4\textwidth}
        \centering
        \input{plots_stability/discriminability_lipschitz.tex}
        \caption{Lipschitz filter}
        \label{subfig:discriminabilityLipschitz}
    \end{subfigure}
    \hspace{1.5cm}
    \begin{subfigure}{0.4\textwidth}
        \centering
        \input{plots_stability/discriminability_integral_lipschitz.tex}
        \caption{Integral Lipschitz filter}
        \label{subfig:discriminabilityIntegralLipschitz}
    \end{subfigure}
    \caption{Discriminability of large eigenvalues. Let $\bbx = \bbv_{n}$ and $\bby= \bbv_{n-1}$ be two different signals that we want to discriminate. (\subref{subfig:discriminabilityLipschitz}) This can be done by using a Lipschitz graph filter with $G=2$ output features, and a reasonable value of $C$. However, if the graph is subject to an edge dilation, then the eigenvalues will fall out of the passband of the frequency response, and thus yield an output of zero. Therefore, Lipschitz filters can discriminate signal with large eigenvalue content, but cannot do so in a stable manner. (\subref{subfig:discriminabilityIntegralLipschitz}) An integral Lipschitz filter is not able to discriminate between $\bbx$ and $\bby$ since it cannot be narrow for large eigenvalues (unless the integral Lipschitz constant $C$ is very large, compromising the stability). In summary, Lipschitz filters can discriminate large eigenvalue content, but are not stable; while integral Lipschitz filters are stable, but cannot discriminate large eigenvalue content.}
    \label{fig:discriminability}
\end{figure*}
%%
%%%%          End of FIGURE         %%%%
%%%%%%%%%%%%%%%%%%%%%%%%%%%%%%%%%%%%%%%%

{The main conclusion and key takeaway of Theorems~\ref{thm:relStabilityGraphFilter}~and~\ref{thm:relStabilityGNN} is that the stability bound of both graph filters and GNNs is linear on the size of the perturbation, making both parametrizations stable to relative perturbations of the graph support.} This bound also holds for all graphs with the same size $n$. We emphasize that this bound establishes Lipschitz continuity of graph filters and GNNs \emph{with respect to changes in the underlying support}, not with respect to the input\footnote{GNNs and graph filters are also Lipschitz continuous with respect to the input, and this is trivial to show by using operator norms.}. We further emphasize that the results in Theorem~\ref{thm:relStabilityGraphFilter}~and~\ref{thm:relStabilityGNN} hold for parametrizations using the same tensor filter coefficients $\bbH$. More specifically, stability to relative perturbations requires that the graph filters obtained after training be integral Lipschitz (Def.~\ref{def:integralLipschitzFiters}). This condition is trivial on a bounded support $[\lambda_{\min},\lambda_{\max}]$ for filters given by an analytic frequency response \eqGraphFilter. {As a matter of fact, the actual value of $C$ can be impacted during training by adding the integral Lipschitz condition \eqref{eq:integralLipschitzFilters} as a penalty on the loss function of the corresponding ERM problem \eqERMparam.}

{The stability bound of Theorems~\ref{thm:relStabilityGraphFilter}~and~\ref{thm:relStabilityGNN} is proportional to the size of the perturbation. The proportionality constant is given by two terms. The first term is $(1+\delta \sqrt{n})$ and involves the eigenvector misalignment constant $\delta$, which measures the change in the graph frequency basis caused by the perturbation. This term is given by the admissible perturbations of the specific problem under consideration. We note that while $\delta$ provided here applies for any graph and any relative perturbation (Def.~\ref{def:relPerturbation}), it is a coarse bound which can be improved if we know that the space of possible perturbations is restricted by extraneous information, as is the case of Euclidean data \cite{Mallat12-Scattering}. For a numerical experiment showing how conservative the bound is, please see \cite[Fig. 6]{Gama19-Stability}.}
        
{The second term is $CG$ for graph filters or $CB^{L-1}\prod_{l=1}^{L} F_{l}$ for GNNs, and is a direct consequence of the design choices that result in the specific graph filters used in the parametrization. The values of $G$ or $\prod_{l=1}^{L} F_{l}$ are design choices, while the values of $C$ and $B$ result from the training phase. As discussed earlier, both of these values can be impacted by an appropriate choice of penalty function during training, if stability is to be increased. We note that the resulting filters can thus compensate for the specific perturbation characteristics.}

\begin{remark}[{Absolute perturbations}]
\normalfont
{An alternative to the relative perturbation model is the absolute one \cite{Gama19-Stability}. In this case, the distance between $\bbS$ and $\hbS$ is given by the norm of a matrix $\bbE$ such that we can write $\bbP\hbS\bbP^{\Tr} = \bbS + \bbE$ for some perturbation matrix $\bbP$. Note, however, that this model can be misleading in that the graph structure can be altered completely without this being reflected in the value of $\varepsilon$. To see this, consider a stochastic block model with two disconnected communities. An absolute perturbation given by the identity matrix results in a perturbed graph that still respects this two-block structure. However, an absolute perturbation given by the anti-diagonal identity matrix would disrupt this two-block structure by forcing connections between the blocks. Yet, both perturbations have the same absolute size $\varepsilon$. This is also evident in that the sparsity of the graph is completely lost. As we can see, absolute perturbations do not capture the specifics of the graph support they affect, so we choose to focus on relative perturbations. Details on the stability under absolute perturbation model can be found in \cite{Gama19-Stability}.}
\end{remark}

\begin{remark}[Computation of the bound]
\normalfont
{The key contribution from Theorems~\ref{thm:relStabilityGraphFilter}~and~\ref{thm:relStabilityGNN} is that the change in the output of a GNN due to a change in the graph support is proportional to the size of the perturbation. This has important implications in that a GNN trained on one graph can be used on another graph as long as the graphs are similar. This may entail computing $d(\bbS,\hbS)$ directly, which would lead to a combinatorial problem. To avoid this, we can estimate $d(\bbS,\hbS)$ by computing $\|\bbS - \hbS\|/\|\bbS\|$. As for the proportionality constant, we emphasize that the stability of the architecture can be affected by changing the integral Lipschitz constant of the filter, which can be done through training. With respect to the eigenvector misalignment constant, knowing its exact value does not alter the conceptual implications of Theorems~\ref{thm:relStabilityGraphFilter}~and~\ref{thm:relStabilityGNN}. This constant depends on the specific perturbation, and if more knowledge is available, it can be computed directly, as is the case of the diffeomorphism in \cite{Mallat12-Scattering}. Alternatively, more restrictions can be imposed on it \cite[Theorem~3]{Gama19-Stability}. In any case, we note that $\delta \leq 8$ always holds since it is related to the norm of unitary matrices.}
\end{remark}

%%%%%%%%%%%%%%%%%%%%%%%%%%%%%%%%%%%%%%%%%%%%%%%%%%%%%%%%%%%%%%%%%%%%%%%%%%%%%%%%
%%%%                              SUBSECTION                                %%%%
%%%%%%%%%%%%%%%%%%%%%%%%%%%%%%%%%%%%%%%%%%%%%%%%%%%%%%%%%%%%%%%%%%%%%%%%%%%%%%%%

\subsection{Discussion and insights} \label{subsec:discussion}

Graph signals $\bbX$ can be completely characterized by their frequency content $\tbX$ given the one-to-one correspondence between the GFT and the inverse GFT [cf. \eqref{eq:GFT}]. Therefore, to analyze, understand, and learn from signals, we need to use functions $\Phi$ that adequately capture the difference and similarities of signals throughout the frequency spectrum \cite{Sandryhaila14-DSPGfreq}. This concept is known in signal processing as filter discriminability, and is concerned with how well a function $\Phi$ can tell apart different sections of the frequency spectrum.

{In graphs, the spectrum is discrete and given by the eigenvalues $\lambda_{1} < \cdots < \lambda_{n}$ of the graph support $\bbS$. Perturbations to the graph structure $\bbS$ alter the eigenvalues and, therefore, alter the location of the different frequency coefficients of the signal within the given spectrum. It is evident, then, that the concept of discriminability is related to the concept of stability, since relevant parts of the spectrum that need to be told apart (discriminability) change under perturbations of the graph support (stability). Thus, to analyze both the discriminability and stability of a graph filter, we need to analyze the shape of its frequency response (Def.~\ref{def:freqResponse}).}

%%%%%%%%%%%%%%%%%%%%%%%%%%%%%%%%%%%%%%%%
%%%%             FIGURE             %%%%
%%%%%%%%%%%%%%%%%%%%%%%%%%%%%%%%%%%%%%%%
%
\def \thisplotscale {2.8}
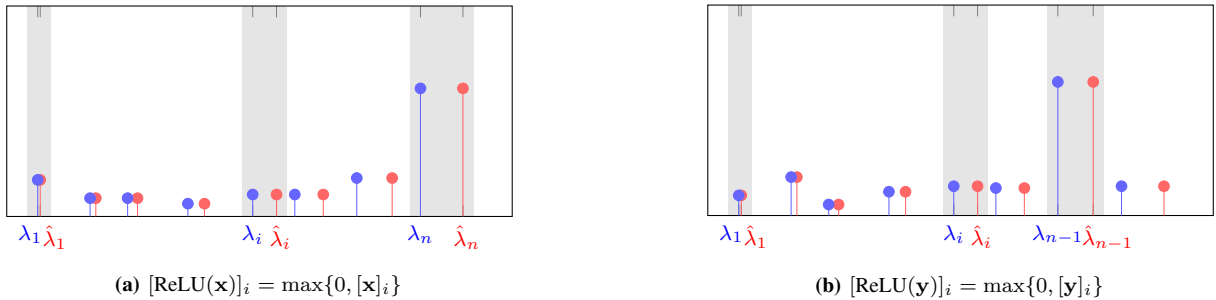
\begin{figure*}
    \centering
    \begin{subfigure}{0.4\textwidth}
        \centering
        \input{plots_stability/relu_spectrum_1.tex}
        \caption{$[\text{ReLU}(\bbx)]_{i} = \max\{0,[\bbx]_{i}\}$}
        \label{subfig:ReLUx}
    \end{subfigure}
    \hspace{1.5cm}
    \begin{subfigure}{0.4\textwidth}
        \centering
        \input{plots_stability/relu_spectrum_2.tex}
        \caption{$[\text{ReLU}(\bby)]_{i} = \max\{0,[\bby]_{i}\}$}
        \label{subfig:ReLUy}
    \end{subfigure}
    \caption{Effect of applying nonlinearities. (\subref{subfig:ReLUx}) Frequency content of signal $\sigma(\bbx) = \text{ReLU}(\bbv_{n})$. (\subref{subfig:ReLUy}) Frequency content of signal $\sigma(\bby) = \text{ReLU}(\bbv_{n-1})$. The use of nonlinearities creates frequency content in parts of the spectrum that there were none. The nonlinearity spreads the frequency content throughout the spectrum, in an effect akin to demodulation. This is a fundamental contribution of nonlinearities, since frequency content at low eigenvalues can be stably discriminated by the graph filters used in the following layer. While we cannot control what shape the signal will have after being applied a nonlinearity, we observe that this content will likely be different, and thus, will be further discriminated. The effect of nonlinearities allows GNNs to process content in large eigenvalues in a stable manner (by spreading it into low eigenvalues).}
    \label{fig:ReLU}
\end{figure*}
%%
%%%%          End of FIGURE         %%%%
%%%%%%%%%%%%%%%%%%%%%%%%%%%%%%%%%%%%%%%%

Stability to relative perturbations (Def.~\ref{def:relPerturbation}) requires integral Lipschitz filters (Def.~\ref{def:integralLipschitzFiters}) as per Theorems~\ref{thm:relStabilityGraphFilter}~and~\ref{thm:relStabilityGNN}. The maximum discriminability of integral Lipschitz filters, however, is not only determined by the integral Lipschitz constant $C$, but also by the position in the spectrum. Recall that integral Lipschitz filters are Lipschitz with a constant $2C/(\lambda_{1}+\lambda_{2})$ that depends on the spectrum. Thus, if we are in a portion of the spectrum where $\lambda$ is large, then the discriminability is very poor since the maximum derivative has to be almost zero, irrespective of $C$. On the contrary, if we are on the low-eigenvalue part of the spectrum, the discriminability can be arbitrarily high, since the derivative of the frequency response can be arbitrarily large. In a way, the value of $C$ helps to determine the eigenvalue at which the integral Lipschitz filters enter the \emph{flat} zone (larger $C$ implies that larger eigenvalues can be discriminated before the filter becomes flat), but do not affect the overall discrminability for small eigenvalues. The value of $C$, however, does affect the stability of both graph filters and GNNs, where lower values of $C$ means more stable representations (Theorems~\ref{thm:relStabilityGraphFilter}~and~\ref{thm:relStabilityGNN}).

{This implies that, under the relative perturbation model, the discriminability of the filters is independent of their stability, meaning that around low eigenvalues they can be arbitrarily discriminative, while at high eigenvalues, they cannot discriminate any frequency coefficient. All of this, irrespective of the value of $C$. This suggests that integral Lipschitz graph filters are well equipped to successfully learn from signals, as long as the relevant information is located in low-eigenvalue content. This limits their use to this specific class of signals. GNNs, however, can successfully capture information from high-eigenvalues by leveraging the nonlinearity and the subsequent graph filters. This can be better understood by looking at a specific, illustrative and conceptual example as we do next.}

Consider the particular case of a perturbation that is given by an edge dilation, that is $\hbS = (1+\varepsilon) \bbS$, where $\varepsilon \approx 0$ is small. This is a particular instance of a relative perturbation model [cf. Def.~\ref{def:relPerturbation}]. In the case of the movie recommendation problem, this can happen if we use a biased estimator to compute the rating similarities, and thus $\hbS$, the graph on which we operate, is an edge dilation of the actual graph $\bbS$. Note that $\hbS$ and $\bbS$ share the same eigenvectors, so that the eigenvector misalignment constant of Theorems~\ref{thm:relStabilityGraphFilter}~and~\ref{thm:relStabilityGNN} is $\delta=0$. The eigenvalues get perturbed as $\hat{\lambda}_{i} = (1+\varepsilon) \lambda_{i}$. This implies that larger eigenvalues get perturbed more than smaller eigenvalues.

{In the context of this very simple edge dilation perturbation, we see in Fig.~\ref{subfig:graphDilationLipschitz} an illustration that Lipschitz filters are not stable.} This is because for large eigenvalues, the change in the output of a filter is very large, even if the perturbation $\varepsilon$ is small. To see this, notice that $|h(\hat{\lambda}_{i}) - h(\lambda_{i})| \leq C |\hat{\lambda}_{i} - \lambda_{i}| = C \varepsilon \lambda_{i}$, so that if $\lambda_{i}$ is large, the difference in the filter output $|h(\hat{\lambda}_{i}) - h(\lambda_{i})|$ can be very large, even if $\varepsilon$ is small.

{In contrast, integral Lipschitz filters are stable, as illustrated in Fig.~\ref{subfig:graphDilationIntegralLipschitz}.} For low eigenvalues these filters can have arbitrary variations, but since small $\varepsilon$ does not cause a big change in the eigenvalues, the output is similar. For large eigenvalues, the frequency response is flat; thus, even if there is a high variability of the eigenvalues, the filter output remains constant. This follows from the integral Lipschitz condition, where $|h(\hat{\lambda}_{i}) - h(\lambda_{i})| \leq 2C |\hat{\lambda}_{i} - \lambda_{i}|/|\hat{\lambda}_{i} + \lambda_{i}| %= 2C\varepsilon \lambda_{i}/(2\lambda_{i}+ \varepsilon \lambda_{i}) = 2C\varepsilon/(2\varepsilon+1) 
\approx 2C\varepsilon$ only depends on $\varepsilon$ but not on the specific eigenvalue, leading to stability.

The price that integral Lipschitz filters pay for stability is that they cannot discriminate information located at high eigenvalues. Consider that we want to tell apart two single-feature signals, $\bbx = \bbv_{n}$ and $\bby = \bbv_{n-1}$, where $\bbv_{i}$ is the eigenvector associated to $\lambda_{i}$ (or $\hat{\lambda}_{i}$ in the perturbed graph). {As we can see on the illustration in Fig.~\ref{subfig:discriminabilityIntegralLipschitz}, this is not doable by means of integral Lipschitz filters. On the contrary, we could easily discriminate between these two signals by using Lipschitz filters, as illustrated in Fig.~\ref{subfig:discriminabilityLipschitz}.} However, this leads to an unstable filter, as discussed before. Therefore, when using linear graph filters as parametrizations $\Phi$, we are faced with the trade-off between discriminability and stability (where we need to increase the $C$ of integral Lipschitz filters to achieve discriminability at high eigenvalues) or, alternatively, stick to processing graph signals whose relevant information is located on low eigenvalues.

GNNs are stable under relative perturbations by employing integral Lipschitz filters (Theorem~\ref{thm:relStabilityGNN}). While, as discussed above, integral Lipschitz filters are unable to discriminate information located in high eigenvalues, GNNs can do so by leveraging the pointwise nonlinearity. Essentially, applying a nonlinearity to a signal spreads its information content throughout the spectrum, creating frequency content in locations where it was not before. {As we can see on the illustration in Fig.~\ref{subfig:ReLUx}, the frequency content of $\bbx=\bbv_{n}$ after applying the nonlinearity is located throughout the frequency spectrum. The same happens when applying $\sigma$ to $\bby=\bbv_{n-1}$, as shown on the illustration in Fig.~\ref{subfig:ReLUy}.} Even more so, the resulting frequency content is different in both resulting signals. Once the frequency content has been spread throughout the spectrum, the integral Lipschitz graph filters can, indeed, discriminate between these two signals by processing only the low-eigenvalue frequency content. In essence, the nonlinearities in GNNs act as frequency demodulators, spreading the information content throughout the spectrum. This allows for subsequent filters to process this information in a stable manner. Thus, GNNs improve on graph filters, by processing information in a way that is simultaneously discriminative and stable.

%\begin{remark}[Perturbation models] \label{rmk:perturbationModels} \normalfont
%    The absolute and relative perturbation models have been described separately for ease of exposition. However, both models are complementary and can be jointly analyzed as a single perturbation model. If this is the case, stability of graph filters and GNNs follows immediately from the proofs of Theorems~\ref{thm:absStabilityGraphFilter} through \ref{thm:relStabilityGNN}, and determines that the filters involved have to be simultaneously Lipschitz (Def.~\ref{def:LipschitzFiters}) and integral Lipschitz (Def.~\ref{def:integralLipschitzFiters}). This means that the filters are flat for large eigenvalues, but how narrow they can be (even around the zero eigenvalue) is also restricted. Thus, the need to be able to process mid-range eigenvalues becomes of greater importance. In this sense, the use of nonlinearities and layers of bank filters help GNNs outperform linear graph filters as well. By spreading the information throughout the spectrum and using various filters, creates several instances where the information can be collected and successfully discriminated.
%\end{remark}

%% file: plots_stability/lipschitz_filters.tex
%!TEX root = ../root.tex

%\def \thisplotscale {3.68}
\def \unit {\thisplotscale cm}

\def \frequencyresponse 
%{ 0.9 - 0.9*exp(-(0.25*(x+0.5))^2) }
{ 0.9*exp(-(3.0*(x-0.0))^2) }
\def \frequencyresponseTwo
{ 0.9*exp(-(3.0*(x-0.6))^2) }
\def \frequencyresponseThree
{ 0.9*exp(-(3.0*(x-1.8))^2) }
\def \frequencyresponseFour
{ 0.9*exp(-(3.0*(x-3.6))^2) }
\def \frequencyresponseFive
{ 0.9*exp(-(3.0*(x-5.6))^2) }

\begin{tikzpicture}[x = 1*\unit, y=1*\unit]
\begin{axis}[scale only axis,
             width  = 2.4*\unit,
             height = 0.8*\unit,
             xmin = -0.5, xmax=6.8,
             xtick = {0, 0.6, 1.8, 3.6, 5.6},
             xticklabels = {$\lam_1=0\ $, $\lam_2$, $\lam_3$, $\lam_4$, $\lam_5$},
             ymin = -0, ymax = 1.05,
             ytick = {-1}]

\addplot[domain=-0.5:1.0,samples = 80, color = red, thick] {\frequencyresponse};
\addplot[domain=-0.5:1.6,samples = 80, color = red, thick] {\frequencyresponseTwo};
\addplot[domain=-0.5:2.8,samples = 80, color = red, thick] {\frequencyresponseThree};
\addplot[domain=2.6:4.6,samples = 80, color = red, thick] {\frequencyresponseFour};
\addplot[domain=4.6:6.6,samples = 80, color = red, thick] {\frequencyresponseFive};

\end{axis}
\end{tikzpicture}

%% file: plots_stability/integral_lipschitz_filters.tex
%!TEX root = ../root.tex

%\def \thisplotscale {3.68}
\def \unit {\thisplotscale cm}

\def \frequencyresponse 
%{ 0.9 - 0.9*exp(-(0.25*(x+0.5))^2) }
{ 0.9*exp(-(6.0*(x-0.0))^2) }
\def \frequencyresponseTwo
{ 0.9*exp(-(3.0*(x-0.6))^2) }
\def \frequencyresponseThree
{ 0.9*exp(-(1.5*(x-1.8))^2) }
\def \frequencyresponseFour
{ 0.9*exp(-(1.0*(x-3.6))^2) }
\def \frequencyresponseFive
{ 0.9 - 0.9*exp(-(0.7*(x-3.0))^2) }

\begin{tikzpicture}[x = 1*\unit, y=1*\unit]
\begin{axis}[scale only axis,
             width  = 2.4*\unit,
             height = 0.8*\unit,
             xmin = -0.5, xmax=6.8,
             xtick = {0, 0.6, 1.8, 3.6, 5.6},
             xticklabels = {$\lam_1=0\ $, $\lam_2$, $\lam_3$, $\lam_4$, $\lam_5$},
             ymin = -0, ymax = 1.05,
             ytick = {-1}]

\addplot[domain=-0.5:0.6,samples = 80, color = red, thick] {\frequencyresponse};
\addplot[domain=-0.5:1.8,samples = 80, color = red, thick] {\frequencyresponseTwo};
\addplot[domain=-0.5:3.6,samples = 80, color = red, thick] {\frequencyresponseThree};
\addplot[domain=-0.5:6.8,samples = 80, color = red, thick] {\frequencyresponseFour};
\addplot[domain= 3.0:6.8,samples = 80, color = red, thick] {\frequencyresponseFive};

\end{axis}
\end{tikzpicture}

%% file: plots_stability/graph_dilation_lipschitz.tex
%!TEX root = ../root.tex

%\def \thisplotscale {3.2}
\def \unit {\thisplotscale cm}

\def \frequencyresponse 
     {   0.8*exp(-(1*(x-1.2))^2) 
       + 0.7*exp(-(0.7*(x-4))^2) 
       + 0.8*exp(-(1.4*(x-6))^2) 
       + 0.1}

\hspace{-2.9mm}
\begin{tikzpicture}[x = 1*\unit, y=1*\unit]

\def \factorx {2.4/8}
\def \deltax  {0.5*\factorx}
\def \shadeshift  {0.05}

\path [fill=black!20, opacity = 0.5] 
              (\deltax - 0.001*\factorx - \shadeshift, 0.00) rectangle 
              (\deltax + 0.030*\factorx + \shadeshift, 1.00);
\path [fill=black!20, opacity = 0.5] 
              (\deltax + 3.393*\factorx - \shadeshift, 0.00) rectangle 
              (\deltax + 3.770*\factorx + \shadeshift, 1.00);
\path [fill=black!20, opacity = 0.5] 
              (\deltax + 6.048*\factorx - \shadeshift, 0.00) rectangle 
              (\deltax + 6.720*\factorx + \shadeshift, 1.00);

\begin{axis}[scale only axis,
             width  = 2.4*\unit,
             height = 1*\unit,
             xmin = -0.5, xmax=7.5,
             xtick = {0.03, -0.01, 3.77, 3.393, 6.72, 6.048},
             xticklabels = {\red{$\qquad\hat{\lambda}_1\phantom{\lambda}$},
                            \blue{$\lambda_1\ \ $}, 
                            \red{$\quad\hat{\lambda}_i\phantom{\lambda}$}, 
                            \blue{$\lambda_i$},
                            \red{$\quad\hat{\lambda}_{n}\phantom{\lambda}$},
                            \blue{$\lambda_n$}},
             ymin = -0, ymax = 1.15,
             ytick = {-1},
             enlarge x limits=false]

\addplot+[samples at = {0.03, 0.91, 1.57, 
                        2.63, 3.77, 4.51, 
                        5.60, 6.72}, 
          color = red!60, 
          ycomb, 
          mark=otimes*, 
          mark options={red!60}]
         {\frequencyresponse};

\addplot+[samples at = {-0.01, 0.819, 1.413, 
                        2.367, 3.393, 4.059, 
                        5.04, 6.048}, 
          color = blue!60, 
          ycomb, 
          mark=oplus*, 
          mark options={blue!60}]
         {\frequencyresponse};

\addplot[ domain=-0.5:7.5, 
          samples = 80, 
          color = black,
          line width = 1.2]
         {\frequencyresponse};

\end{axis}
\end{tikzpicture}

%{-0.22, 0.25, 0.61, 1.05, 1.37, 2.11, 
%  2.43, 3.07, 3.67, 4.13, 4.31, 5.05, 
%  5.20, 6.00, 6.42, 7.00}

%% file: plots_stability/graph_dilation_integral_lipschitz.tex
%!TEX root = ../root.tex

%\def \thisplotscale {3.68}
\def \unit {\thisplotscale cm}

\def \frequencyresponse 
     { 0.9 - 0.7*exp(-(0.7*(x-1.6))^2) }

\hspace{-2.9mm}
\begin{tikzpicture}[x = 1*\unit, y=1*\unit]

\def \factorx {2.4/8}
\def \deltax  {0.5*\factorx}
\def \shadeshift  {0.05}

\path [fill=black!20, opacity = 0.5] 
              (\deltax - 0.001*\factorx - \shadeshift, 0.00) rectangle 
              (\deltax + 0.030*\factorx + \shadeshift, 1.00);
\path [fill=black!20, opacity = 0.5] 
              (\deltax + 3.393*\factorx - \shadeshift, 0.00) rectangle 
              (\deltax + 3.770*\factorx + \shadeshift, 1.00);
\path [fill=black!20, opacity = 0.5] 
              (\deltax + 6.048*\factorx - \shadeshift, 0.00) rectangle 
              (\deltax + 6.720*\factorx + \shadeshift, 1.00);

\begin{axis}[scale only axis,
             width  = 2.4*\unit,
             height = 1*\unit,
             xmin = -0.5, xmax=7.5,
             xtick = {0.03, -0.01, 3.77, 3.393, 6.72, 6.048},
             xticklabels = {\red{$\qquad\hat{\lambda}_1\phantom{\lambda}$},
                            \blue{$\lambda_1\ \ $}, 
                            \red{$\quad\hat{\lambda}_i\phantom{\lambda}$}, 
                            \blue{$\lambda_i$},
                            \red{$\quad\hat{\lambda}_{n}\phantom{\lambda}$},
                            \blue{$\lambda_n$}},
             ymin = -0, ymax = 1.15,
             ytick = {-1},
             enlarge x limits=false]

\addplot+[samples at = {0.03, 0.91, 1.57, 
                        2.63, 3.77, 4.51, 
                        5.60, 6.72}, 
          color = red!60, 
          ycomb, 
          mark=otimes*, 
          mark options={red!60}]
         {\frequencyresponse};

\addplot+[samples at = {-0.01, 0.819, 1.413, 
                        2.367, 3.393, 4.059, 
                        5.04, 6.048}, 
          color = blue!60, 
          ycomb, 
          mark=oplus*, 
          mark options={blue!60}]
         {\frequencyresponse};

\addplot[ domain=-0.5:7.5, 
          samples = 80, 
          color = black,
          line width = 1.2]
         {\frequencyresponse};

\end{axis}
\end{tikzpicture}

%{-0.22, 0.25, 0.61, 1.05, 1.37, 2.11, 
%  2.43, 3.07, 3.67, 4.13, 4.31, 5.05, 
%  5.20, 6.00, 6.42, 7.00}

%% file: plots_stability/discriminability_lipschitz.tex
%!TEX root = ../root.tex

%\def \thisplotscale {3.68}
\def \unit {\thisplotscale cm}

\def \frequencyresponse 
     {1.1*exp(-(2.5*(x-6.048))^2}
\def \frequencyresponsetwo 
     {1.1*exp(-(2.5*(x-5.04))^2}

\hspace{-2.9mm}
\begin{tikzpicture}[x = 1*\unit, y=1*\unit]

\def \factorx {2.4/8}
\def \deltax  {0.5*\factorx}
\def \shadeshift  {0.05}

\path [fill=black!20, opacity = 0.5] 
              (\deltax + 6.048*\factorx - \shadeshift, 0.00) rectangle 
              (\deltax + 6.720*\factorx + \shadeshift, 1.00);

\path [fill=black!20, opacity = 0.5] 
              (\deltax + 5.04*\factorx - \shadeshift, 0.00) rectangle 
              (\deltax + 5.60*\factorx + \shadeshift, 1.00);

\begin{axis}[scale only axis,
             width  = 2.4*\unit,
             height = 1*\unit,
             xmin = -0.5, xmax=7.5,
             xtick = {5.60, 5.04, 6.72, 6.048},
             xticklabels = {\red{$\ \qquad\hat{\lambda}_{n-1}\phantom{\lambda_{n-1}}$},
                            \blue{$\lambda_{n-1}\qquad$}, 
                            \red{$\qquad\hat{\lambda}_{n}\phantom{\lambda}$},
                            \blue{$\quad\lambda_n$}},
             ymin = -0, ymax = 1.15,
             ytick = {-1},
             enlarge x limits=false]

\addplot+[samples at = {0.03, 0.91, 1.57, 
                        2.63, 3.77, 4.51}, 
          color = red!60, 
          ycomb, 
          mark=otimes*, 
          mark options={red!60}]
         {0};

\addplot+[samples at = {6.72, 5.60}, 
          color = red!60, 
          ycomb, 
          mark=otimes*, 
          mark options={red!60}]
         {1};

\addplot+[samples at = {-0.01, 0.819, 1.413, 
                        2.367, 3.393, 4.059}, 
          color = blue!60, 
          ycomb, 
          mark=oplus*, 
          mark options={blue!60}]
         {0};

\addplot+[samples at = {6.048, 5.04}, 
          color = blue!60, 
          ycomb, 
          mark=oplus*, 
          mark options={blue!60}]
         {1};

\addplot[ domain=-0.5:5.5, 
          samples = 2, 
          color = black,
          line width = 1.2]
         {0};

\addplot[ domain=5.0:7.5, 
          samples = 70, 
          color = black,
          line width = 1.2]
         {\frequencyresponse};

\addplot[ domain=4.0:7.0, 
          samples = 70, 
          color = black,
          line width = 1.2]
         {\frequencyresponsetwo};
\addplot[ domain=7.0:7.5, 
          samples = 2, 
          color = black,
          line width = 1.2]
         {0};

\end{axis}
\end{tikzpicture}

%{-0.22, 0.25, 0.61, 1.05, 1.37, 2.11, 
%  2.43, 3.07, 3.67, 4.13, 4.31, 5.05, 
%  5.20, 6.00, 6.42, 7.00}

%% file: plots_stability/discriminability_integral_lipschitz.tex
%!TEX root = ../root.tex

%\def \thisplotscale {3.68}
\def \unit {\thisplotscale cm}

\def \frequencyresponse 
	{ 1.1 - 0.7*exp(-(0.7*(x-1.6))^2) }

\hspace{-2.9mm}
\begin{tikzpicture}[x = 1*\unit, y=1*\unit]

\def \factorx {2.4/8}
\def \deltax  {0.5*\factorx}
\def \shadeshift  {0.05}

\path [fill=black!20, opacity = 0.5] 
(\deltax + 6.048*\factorx - \shadeshift, 0.00) rectangle 
(\deltax + 6.720*\factorx + \shadeshift, 1.00);

\path [fill=black!20, opacity = 0.5] 
(\deltax + 5.04*\factorx - \shadeshift, 0.00) rectangle 
(\deltax + 5.60*\factorx + \shadeshift, 1.00);

\begin{axis}[scale only axis,
width  = 2.4*\unit,
height = 1*\unit,
xmin = -0.5, xmax=7.5,
xtick = {5.60, 5.04, 6.72, 6.048},
xticklabels = {\red{$\ \qquad\hat{\lambda}_{n-1}\phantom{\lambda_{n-1}}$},
    \blue{$\lambda_{n-1}\qquad$}, 
    \red{$\qquad\hat{\lambda}_{n}\phantom{\lambda}$},
    \blue{$\quad\lambda_n$}},
ymin = -0, ymax = 1.15,
ytick = {-1},
enlarge x limits=false]
             
\addplot+[samples at = {0.03, 0.91, 1.57, 
    2.63, 3.77, 4.51}, 
color = red!60, 
ycomb, 
mark=otimes*, 
mark options={red!60}]
{0};

\addplot+[samples at = {6.72, 5.60}, 
color = red!60, 
ycomb, 
mark=otimes*, 
mark options={red!60}]
{1};

\addplot+[samples at = {-0.01, 0.819, 1.413, 
                        2.367, 3.393, 4.059}, 
          color = blue!60, 
          ycomb, 
          mark=oplus*, 
          mark options={blue!60}]
         {0};

\addplot+[samples at = {6.048, 5.04}, 
          color = blue!60, 
          ycomb, 
          mark=oplus*, 
          mark options={blue!60}]
         {1};

\addplot[ domain=-0.5:7.5, 
samples = 80, 
color = black,
line width = 1.2]
{\frequencyresponse};

\end{axis}
\end{tikzpicture}

%{-0.22, 0.25, 0.61, 1.05, 1.37, 2.11, 
%  2.43, 3.07, 3.67, 4.13, 4.31, 5.05, 
%  5.20, 6.00, 6.42, 7.00}

%% file: plots_stability/relu_spectrum_1.tex
%!TEX root = ../root.tex

%\def \thisplotscale {3.68}
\def \unit {\thisplotscale cm}

\def \frequencyresponse 
     {   0.8*exp(-(1*(x-1.2))^2) 
       + 0.7*exp(-(0.7*(x-4))^2) 
       + 0.8*exp(-(1.4*(x-6))^2) 
       + 0.1}

\hspace{-2.9mm}
\begin{tikzpicture}[x = 1*\unit, y=1*\unit]

\def \factorx {2.4/8}
\def \deltax  {0.5*\factorx}
\def \shadeshift  {0.05}

\path [fill=black!20, opacity = 0.5] 
              (\deltax - 0.001*\factorx - \shadeshift, 0.00) rectangle 
              (\deltax + 0.030*\factorx + \shadeshift, 1.00);
\path [fill=black!20, opacity = 0.5] 
              (\deltax + 3.393*\factorx - \shadeshift, 0.00) rectangle 
              (\deltax + 3.770*\factorx + \shadeshift, 1.00);
\path [fill=black!20, opacity = 0.5] 
              (\deltax + 6.048*\factorx - \shadeshift, 0.00) rectangle 
              (\deltax + 6.720*\factorx + \shadeshift, 1.00);

\begin{axis}[scale only axis,
             width  = 2.4*\unit,
             height = 1*\unit,
             xmin = -0.5, xmax=7.5,
             xtick = {0.03, -0.01, 3.77, 3.393, 6.72, 6.048},
             xticklabels = {\red{$\qquad\hat{\lambda}_1\phantom{\lambda}$},
                            \blue{$\lambda_1\ \ $}, 
                            \red{$\quad\hat{\lambda}_i\phantom{\lambda}$}, 
                            \blue{$\lambda_i$},
                            \red{$\quad\hat{\lambda}_{n}\phantom{\lambda}$},
                            \blue{$\lambda_n$}},
             ymin = -0, ymax = 1.15,
             ytick = {-1},
             enlarge x limits=false]

\addplot+[color = red!60, 
          ycomb, 
          mark=otimes*, 
          mark options={red!60}]
          coordinates { (0.03, 0.20)
                        (0.91, 0.10)
                        (1.57, 0.10)
                        (2.63, 0.07)
                        (3.77, 0.12)
                        (4.51, 0.12)
                        (5.60, 0.21)
                        (6.72, 0.70)};

\addplot+[samples at = {-0.01, 0.819, 1.413, 
                        2.367, 3.393, 4.059, 
                        5.04, 6.048}, 
          color = blue!60, 
          ycomb, 
          mark=oplus*, 
          mark options={blue!60}]
          coordinates { (-0.010, 0.20)
                        ( 0.819, 0.10)
                        ( 1.413, 0.10)
                        ( 2.367, 0.07)
                        ( 3.393, 0.12)
                        ( 4.059, 0.12)
                        ( 5.040, 0.21)
                        ( 6.048, 0.70)};

\end{axis}
\end{tikzpicture}

%% file: plots_stability/relu_spectrum_2.tex
%!TEX root = ../root.tex

%\def \thisplotscale {3.68}
\def \unit {\thisplotscale cm}

\def \frequencyresponse 
     {   0.8*exp(-(1*(x-1.2))^2) 
       + 0.7*exp(-(0.7*(x-4))^2) 
       + 0.8*exp(-(1.4*(x-6))^2) 
       + 0.1}

\hspace{-2.9mm}
\begin{tikzpicture}[x = 1*\unit, y=1*\unit]

\def \factorx {2.4/8}
\def \deltax  {0.5*\factorx}
\def \shadeshift  {0.05}

\path [fill=black!20, opacity = 0.5] 
              (\deltax - 0.001*\factorx - \shadeshift, 0.00) rectangle 
              (\deltax + 0.030*\factorx + \shadeshift, 1.00);
\path [fill=black!20, opacity = 0.5] 
              (\deltax + 3.393*\factorx - \shadeshift, 0.00) rectangle 
              (\deltax + 3.770*\factorx + \shadeshift, 1.00);
\path [fill=black!20, opacity = 0.5] 
              (\deltax + 5.040*\factorx - \shadeshift, 0.00) rectangle 
              (\deltax + 5.6*\factorx + \shadeshift, 1.00);

\begin{axis}[scale only axis,
             width  = 2.4*\unit,
             height = 1*\unit,
             xmin = -0.5, xmax=7.5,
             xtick = {0.03, -0.01, 3.77, 3.393, 5.60, 5.040},
             xticklabels = {\red{$\qquad\hat{\lambda}_1\phantom{\lambda}$},
                            \blue{$\lambda_1\ \ $}, 
                            \red{$\quad\hat{\lambda}_i\phantom{\lambda}$}, 
                            \blue{$\lambda_i$},
                            \red{$\qquad\hat{\lambda}_{n-1}\phantom{\lambda}$},
                            \blue{$\lambda_{n-1}$}},
             ymin = -0, ymax = 1.15,
             ytick = {-1},
             enlarge x limits=false]

\addplot+[color = red!60, 
          ycomb, 
          mark=otimes*, 
          mark options={red!60}]
          coordinates { (0.03, 0.11)
                        (0.91, 0.21)
                        (1.57, 0.06)
                        (2.63, 0.13)
                        (3.77, 0.16)
                        (4.51, 0.15)
                        (5.60, 0.73)
                        (6.72, 0.16)};

\addplot+[samples at = {-0.01, 0.819, 1.413, 
                        2.367, 3.393, 4.059, 
                        5.04, 6.048}, 
          color = blue!60, 
          ycomb, 
          mark=oplus*, 
          mark options={blue!60}]
          coordinates { (-0.010, 0.11)
                        ( 0.819, 0.21)
                        ( 1.413, 0.06)
                        ( 2.367, 0.13)
                        ( 3.393, 0.16)
                        ( 4.059, 0.15)
                        ( 5.040, 0.73)
                        ( 6.048, 0.16)};

\end{axis}
\end{tikzpicture}

%% file: LR-transferability.tex
% !TEX root = root.tex

%%%%%%%%%%%%%%%%%%%%%%%%%%%%%%%%%%%%%%%%%%%%%%%%%%%%%%%%%%%%%%%%%%%%%%%%%%%%%%%%
%%%%                                                                        %%%%
%%%%                             S E C T I O N                              %%%%
%%%%                                                                        %%%%
%%%%%%%%%%%%%%%%%%%%%%%%%%%%%%%%%%%%%%%%%%%%%%%%%%%%%%%%%%%%%%%%%%%%%%%%%%%%%%%%

\section{Transferability of GNNs} \label{sec:transferability}

In different instances of the same network problem, it is not uncommon for different graphs, even of different sizes, to ``look similar'' in the sense that they share certain defining structural characteristics. This motivates studying groups of graphs---or  \textit{graph families}---and investigating whether graph filters and GNNs are transferable within them. Transferability of information processing architectures is key because it allows re-using systems without the need to re-train or re-design. This is especially useful in applications where the network size is dynamic, e.g. recommendation systems for a growing product portfolio (Secs. \ref{sec_reco_systems}, \ref{sec_reco_systems_results}).
% {or resource allocation in wireless networks (Sec. \ref{sbs:transf}).}

From the architecture perspective, transferability is akin to replacing the graph by another graph in the same family, which, in itself, is a kind of perturbation. Therefore, transferability can be seen as a type of stability. In this section, we analyze the transferability of graph filters and GNNs in a similar fashion to Sec. \ref{sec:stability}, with particular focus on {families of undirected graphs} identified by objects called \textit{graphons}.  
%We start by reviewing graphons in Sec. \ref{subsec:graphons}, where their limit object interpretation and their role as a generating models for deterministic graphs are also discussed. The graphon signal processing framework is then introduced in Sec. \ref{subsec:wsp}, where we define graphon filters and study both how they can be used to generate graph filters and how graph filters may be used to approximate graphon filters arbitrarily well. These analyses culminate in the transferability analysis of graph filters, which is presented in Sec. \ref{subsec:transf1}. Graphon neural networks are then discussed in Sec. \ref{subsec:wnns}. The concept of graphon neural networks is important because they too can be interpreted as generating models for GNNs, which allows showing that, on very large graphs, GNNs provide a good approximation of WNNs. The existence of a formal transferability bound for GNNs, which is discussed in Sec. \ref{subsec:transf2}, is a direct consequence of this result.
All analyses assume the multi-layer, single feature architecture of Sec. \ref{sec_ch8_gnns_multiple_layers}.

\subsection{Graphons and graph families} \label{subsec:graphons}

Graphons are bounded, symmetric and measurable functions~$\bbW: [0,1]^2 \to [0,1]$ which can be thought of as representations of undirected graphs with an uncountable number of nodes. {An example is the exponential graphon $\bbW(u,v) = \exp (-\beta(u-v)^2)$ with parameter $\beta > 0$. Assigning nodes $i$ and $j$ to points $u_i$ and $u_j$ of the unit interval, the weight of the edge $(i,j)$ is given by $\bbW(u_i,u_j)$. This weight is largest when $u_i$ is close to $u_j$, therefore, the exponential graphon can be used to model graphs with cyclic or ring structure.} As suggested by their infinite-dimensional structure, graphons are also the limit objects of convergent sequences of graphs.

A convergent sequence of graphs, denoted $\{\bbG_n\}$, is characterized by the convergence of the density of certain structures, or \textit{motifs}, in the graphs $\bbG_n$. We define these motifs as graphs $\bbF = (V', E')$ that are unweighted and undirected. 
%Homomorphisms of $\bbF$ into $\bbG = (V,E,\bbS)$ are adjacency preserving maps~$\beta: V'\to V$, i.e., maps in which~$(i,j)\in E'$ implies $(\beta(i),\beta(j))\in E$. Observing that there are a total of $|V|^{|V'|}$ possible maps between $V'$ and $V$, and denoting the number of homomorphisms from $\bbF$ into $\bbG$ $\mbox{hom}(\bbF,\bbG)$, it is then possible to define a \textit{density of homomorphisms} from $\bbF$ into $\bbG$ \red{\cite[Chapter ?]{lovasz2012large}} as
%\begin{equation} \label{eqn:hom_density}
%  t(\bbF, \bbG) = \frac{\mbox{hom}(\bbF,\bbG)}{n^{n'}} = \frac{\sum_{\beta} \prod_{(i,j) \in E'} \ccalW(\beta(i),\beta(j))}{n^{n'}} .
%\end{equation}
%Since the quantity $\mbox{hom}(\bbF,\bbG)$ counts the number of matches of the graph motif $\bbF$ in $\bbG$, the density of homomorphisms $t(\bbF,\bbG)$ corresponds to the relative frequency of this motif in $\bbG$.
%Generalized homomorphism densities between graphs and graphons are defined analogously by replacing the sum with an integral. Explicitly, the homomorphism density $t(\bbF,\bbW)$ is given by
%\begin{equation}\label{eqn_hom_density_graphon}
%   t(\bbF,\bbW) = \int_{{[0,1]}^{V'}} 
%                     \prod_{(i,j) \in E'} \bbW(u_i,u_j) 
%                            \prod_{i \in V'} du_i .
%\end{equation}
Homomorphisms of $\bbF$ into $\bbG = (V,E,\bbS)$ are defined as adjacency preserving maps.
% ~$\beta: V'\to V$ in which~$(i,j)\in E'$ implies $(\beta(i),\beta(j)) \in E$. 
There are $|V|^{|V'|} = n^{n'}$ maps from~$V'$ to $V$, but only some of them are homomorphisms. 
%Weighing each homomorphism of $\bbF$ into $\bbG$ by the corresponding edge weights in $\bbG$, we write the total number of homomorphisms that map $\bbF$ into $\bbG$ as $\mbox{hom}(\bbF,\bbG)$, and define the density of  homomorphisms $t(\bbF, \bbG)$ as \citep{lovasz2012large}
%~$\beta: \ccalV'\to \ccalV$, i.e.,
% = \sum_{\beta} \prod_{(i,j) \in \ccalE'} \ccalW(\beta(i),\beta(j))
%\begin{equation} \label{eqn:hom_density}
%   t(\bbF, \bbG) = \frac{\mbox{hom}(\bbF,\bbG)}{n^{n'}} .
%\end{equation} % = \frac{\sum_{\beta} \prod_{(i,j) \in \ccalE'} \ccalW(\beta(i),\beta(j))}{n^{n'}} .
%
%For unweighted graphs $\bbG$, 
%The quantity $\mbox{hom}(\bbF,\bbG)$ counts the number of ways in which the graph motif $\bbF$ can be mapped into $\bbG$. 
Hence, we can define a density of homomorphisms $t(\bbF,\bbG)$, which represents the relative frequency with which the motif $\bbF$ appears in $\bbG$.

%Generalized homomorphism densities between graphs and graphons can be defined by replacing the sum with an integral. The density of homomorphisms $t(\bbF,\bbW)$ is defined as
%
%\begin{equation}\label{eqn_hom_density_graphon}
%   t(\bbF,\bbW) = \int_{{[0,1]}^{\ccalV'}} 
%                      \prod_{(i,j) \in \ccalE'} \bbW(u_i,u_j) 
%                            \prod_{i \in \ccalV'} du_i .
%\end{equation}
%
Homomorphisms of graphs into graphons are defined analogously and denoted $t(\bbF,\bbW)$ for a motif $\bbF$ and a graphon $\bbW$.
The graph sequence $\{\bbG_n\}$ converges to the graphon $\bbW$ if, for all finite, unweighted and undirected graphs $\bbF$, 
\begin{equation} \label{eqn_graphon_convergence}
\lim_{n\to \infty} t(\bbF,\bbG_n) = t(\bbF,\bbW) .
\end{equation}

All graphons are limit objects of convergent graph sequences, and every convergent graph sequence converges to a graphon \cite[Chapter 11]{lovasz2012large}. 
%This means that if the limit $\lim_{n\to\infty} t(\bbF,\bbG_n)$ exists, it can be written as in equation \ref{eqn_hom_density_graphon} for some limit graphon $\bbW$. 
This allows associating graphons with \textit{families} of graphs of different sizes that share structural similarities. The simplest
examples of such graphs 
%instances of 
are those obtained by evaluation of $\bbW$. In particular, our transferability results will hold for \textit{deterministic graphs} $\bbG_n$ constructed by associating the regular partition $u_i = {(i-1)}/{n}$ to nodes $1 \leq i \leq n$, and the weights $\bbW(u_i,u_j)$ to edges $(i,j)$. Explicitly, 
\begin{equation} \label{eqn:deterministic_graph}
[\bbS_n]_{ij} = s_{ij} = \bbW(u_i,u_j)
\end{equation}
where $\bbS_n$ is the adjacency matrix of $\bbG_n$. 
%An example of a stochastic block model graphon and of an $8$-node deterministic graph drawn from it are shown at the top of Figure \ref{fig:gnns_wnns}, from left to right. 
This sequence of deterministic graphs satisfies the condition in \eqref{eqn_graphon_convergence}, and therefore converges to the graphon $\bbW$ \cite[Chapter 11]{lovasz2012large}. The convergence mode in equation \ref{eqn_graphon_convergence} also allows for other, more general graph sequences than those consisting of deterministic graphs.

%In the same way that graphs can be generated from graphons, graphons induced by graphs can be defined. The graphon $\bbW_{\bbG_n}$ induced by $\bbG_n$ is defined by constructing the regular partition $I_1 \cup \ldots \cup I_n$ of $[0,1]$ with $I_i = [(i-1)/n,i/n]$ and writing
%\begin{equation} \label{eqn:graphon_induced}
%\bbW_{\bbG_n}(u,v) = \bbW_{n}(u,v) = {[\bbS_n]_{ij}} \times \mbI(u \in I_i)\mbI(v \in I_j).
%\end{equation}

%%%%%%%%%%%%%%%%%%%%%%%%%%%%%%%%%%%%%%%%%%%%%%%%%%%%%%%%%%%%%%%%%%%%%%%%%%%%%%%%
%%%%                              SUBSECTION                                %%%%
%%%%%%%%%%%%%%%%%%%%%%%%%%%%%%%%%%%%%%%%%%%%%%%%%%%%%%%%%%%%%%%%%%%%%%%%%%%%%%%%

\subsection{Graphon filters} \label{subsec:wsp}

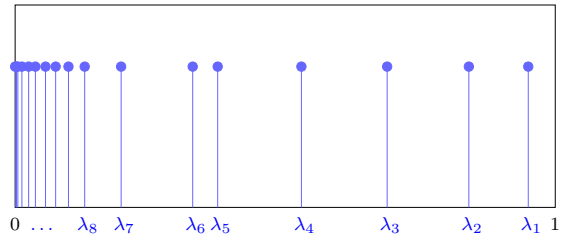
\begin{figure}[t]
    \centering
    \scalebox{.88}{\input{plots_transferability/eigenvalues_graphon.tex}}  
    \caption{Graphon eigenvalues. A graphon has an infinite number of eigenvalues $\lambda_j$ but for any fixed constant $c$ the number of eigenvalues $|\lambda_j|>c$ is finite. Thus, eigenvalues accumulate at $0$ and this is the only accumulation point for graphon eigenvalues.}
    \label{fig:eigenvalues_graphon}
\end{figure}

To understand the behavior of data that may be supported on the graphs belonging to a graphon family, it is also natural to consider the abstractions of \textit{graphon data} and \textit{graphon information processing systems}. 
Graphon data, or graphon signals, are defined as functions $X: [0,1] \to \reals$ of $L^2$. These signals can be modified through graphon operations parametrized by the integral operator
\begin{equation} \label{eqn:graphon_shift}
(T_\bbW X)(v) := \int_0^1 \bbW(u,v)X(u)du
\end{equation}
which is called \textit{graphon shift operator} (WSO) in analogy with the GSO \cite{ruiz2020graphon}. Because $\bbW$ is bounded and symmetric, the WSO is a self-adjoint Hilbert-Schmidt operator, allowing to express $\bbW$ in the operator's spectral basis---the \textit{graphon spectra}---as
\begin{equation}
\bbW(u,v) = \sum_{i \in \mbZ\setminus \{0\}} \lambda_i \varphi_i(u)\varphi_i(v) .
\end{equation}
The operator $T_\bbW$ can thus be rewritten as
\begin{equation} \label{eqn:graphon_spectra}
(T_\bbW X)(v) = \sum_{i \in \mbZ\setminus \{0\}}\lambda_i \varphi_i(v) \int_0^1 \varphi_i(u)X(u)du
\end{equation}
where $\lambda_i$ are the graphon eigenvalues, $\varphi_i$ are the graphon eigenfunctions and $i \in \mbZ\setminus \{0\}$. The eigenvalues are ordered according to their sign and in decreasing order of absolute value, i.e., $1 \geq \lambda_1 \geq \lambda_2 \geq \ldots \geq \ldots \geq \lambda_{-2} \geq \lambda_{-1} \geq -1$. The eigenvalues accumulate around 0 as $|i| \to \infty$, as depicted in Fig. \ref{fig:eigenvalues_graphon} \cite[Theorem 3, Chapter 28]{lax02-functional}. 

Graphon convolutions are defined as shift-and-sum operations where the shift is implemented by the graphon shift operator. Explicitly, a graphon convolutional filter is given by
\begin{align}\begin{split} \label{eqn:lsi-wf}
&\Phi (X; \bbh, \bbW) = \sum_{k=0}^{K} h_k (T_{\bbW}^{(k)} X)(v) = (T_\bbH X)(v) \quad \mbox{with} \\
&(T_{\bbW}^{(k)}X)(v) = \int_0^1 \bbW(u,v)(T_\bbW^{(k-1)} X)(u)du
\end{split}\end{align}
and where $T_{\bbW}^{(0)} = \bbI$ is the identity operator \cite{ruiz2020graphon}. The vector $\bbh = [h_0, \ldots, h_{K}]$ collects the filter coefficients. Using the spectral decomposition in \eqref{eqn:graphon_spectra}, $\Phi (X; \bbh, \bbW)$ can also be written as
\begin{align} \label{eqn:spec-graphon_filter}
\begin{split}
\Phi (X; \bbh, \bbW) &= \sum_{i \in \mbZ\setminus \{0\}} \sum_{k=0}^{K} h_k \lambda_i^k \varphi_i(v) \int_0^1 \varphi_i(u)X(u)du \\
&= \sum_{i \in \mbZ\setminus \{0\}} h(\lambda_i) \varphi_i(v) \int_0^1 \varphi_i(u)X(u)du .
\end{split}
\end{align}
Note that the spectral representation of $\Phi (X; \bbh, \bbW)$ is given by $h(\lambda) = \sum_{k=0}^{K} h_k \lambda^k$, which only depends on the graphon eigenvalues and on the coefficients $h_k$.

\subsubsection{Generating graph filters from graphon filters}

Like the spectral representation of the graphon filter, the spectral representation of the graph filter as shown in Definition \ref{def:freqResponse} depends uniquely on the graph eigenvalues and on the filter coefficients. This allows making the coefficients $h_k$ in equations \eqref{eq:freqResponse} and \eqref{eqn:spec-graphon_filter} the same. Put differently, graphon filters can serve as generating models for graph filters on graphs evaluated from the graphon. Take the graphon filter $\Phi (X; \bbh, \bbW)$ from \eqref{eqn:lsi-wf} and construct a partition $u_i = (i-1)/n$, $1 \leq i \leq n$, of $[0,1]$. The graph filter $\Phi (\bbx_n; \bbh, \bbS_n)= \sum_{k=0}^{K} h_k \bbS_n^k\bbx_n$ can be obtained by defining
\begin{align}\begin{split} \label{eqn:det_graph_filter}
&[\bbS_n]_{ij} = \bbW(u_i,u_j) \quad \mbox{and} \\
&[\bbx_n]_i = X(u_i)
\end{split}\end{align}
where $\bbS_n$ is the GSO of $\bbG_n$, the deterministic graph obtained from $\bbW$ as in equation \eqref{eqn:deterministic_graph}, and $\bbx_n$ is the corresponding \textit{deterministic graph signal} obtained by evaluating $X$ at $u_i$.

Generating graph filters from graphon filters is helpful because it allows designing filters on graphons and applying them to graphs. This decouples the filter design from a specific graph realization. Conversely, it is also possible to define graphon filters induced by graph filters. The graphon filter induced by $\Phi (\bbx_n; \bbh, \bbS_n)=\sum_{k=0}^{K} h_k \bbS_n^k\bbx_n$ is given by
\begin{align}\begin{split} \label{eqn:lsi-wf-induced}
&\Phi (X_n; \bbh, \bbW_n)= \sum_{k=0}^{K} h_k (T_{\bbW_n}^{(k)} X_n)(v) =  \quad \mbox{with} \\
&(T_{\bbW_n}^{(k)}X_n)(v) = \int_0^1 \bbW_n(u,v)(T_{\bbW_n}^{(k-1)} X_n)(u)du
\end{split}\end{align}
where the graphon $\bbW_n$ is the \textit{graphon induced by} $\bbG_n$ and $X_n$ is the \textit{graphon signal induced by the graph signal} $\bbx_n$, i.e.,
\begin{align}\begin{split} \label{eqn:graphon_filter_ind}
&\bbW_{n}(u,v) = {[\bbS_n]_{ij}} \times \mbI(u \in I_i)\mbI(v \in I_j) \quad \mbox{and} \\
&X_n(u) = [\bbx_n]_i \times \mbI(u \in I_i) .
\end{split}\end{align}
This definition allows comparing graph and graphon filters directly, and analyzing the transferability of graph filters to graphs of different sizes.

%The interpretation of GNNs as instantiantions of a WNN is important because it explicitly decouples the GNN from the graph. In this interpretation, the graph is not a fixed hyperparameter of the GNN, but a parameter that can be changed according to the underlying graphon and the value of $n$. This has two important consequences. First, it reveals the ability of GNNs to \textit{scale}. Second, it  allows GNNs to be adapted both by optimizing the weights in $\bbH$ and by changing the graph $\bbG_n$, which adds degrees of freedom to the architecture at no additional computational cost.

\subsubsection{Approximating graph filters with graphon filters}

Consider graph filters obtained from a graphon filter as in \eqref{eqn:det_graph_filter}. For increasing $n$, $\bbG_n$ converges to $\bbW$, which means that these graph filters become increasingly similar to the graphon filter itself. Thus, the graph filter $\Phi (\bbx_n; \bbh, \bbS_n)$ can be used to approximate $\Phi (X; \bbh, \bbW)$. In Theorem \ref{thm:graphon-graph-filter}, we quantify how good this approximation is for different values of $n$. Because the continuous output $Y = \Phi (X; \bbh, \bbW)$ cannot be compared with the discrete output $\bby_n = \Phi (\bbx_n; \bbh, \bbS_n)$ directly, we consider the output of the graphon filter induced by $\Phi (\bbx_n; \bbh, \bbS_n)$, which is given by $Y_n = \Phi (X_n; \bbh, \bbW_n)$ [cf. \eqref{eqn:graphon_filter_ind}]. We also consider the following definitions and assumptions.

\begin{definition}[$c$-band cardinality of $\bbG_n$] \label{def:c_band_card}
The $c$-band cardinality of $\bbG_n$, denoted $B_{nc}$, is the number of eigenvalues $\lambda_i^n$ of $\bbW_{n}$ with absolute value larger or equal to $c$, i.e., $$B_{nc} = \#\{\lambda_i^n\ :\ |\lambda_i^n| \geq c\}.$$
\end{definition}

\begin{definition}[$c$-eigenvalue margin of $\bbG_n$] \label{def:c_eig_margin}
The $c$-eigenvalue margin of $\bbG_n$, denoted $\delta_{nc}$, is given by $$\delta_{nc} = \min_{i, j\neq i} \{ |\lambda_i^n - \lambda_j|\ :\ |\lambda_i^n| \geq c\}$$ where $\lambda_i^n$ and $\lambda_i$ are the eigenvalues of $\bbW_n$ and $\bbW$ respectively.
\end{definition}

\begin{assumption} \label{as1}
The graphon $\bbW$ is $A_1$-Lipschitz, i.e. $|\bbW(u_2,v_2)-\bbW(u_1,v_1)| \leq A_1(|u_2-u_1|+|v_2-v_1|)$.
\end{assumption} 
\begin{assumption} \label{as2}
The spectral response of the convolutional filter, $h$, is $A_2$-Lipschitz and non-amplifying, i.e. {$|h(\lambda)|<1$}.
\end{assumption} 
\begin{assumption} \label{as3}
The graphon signal $X$ is $A_3$-Lipschitz.
\end{assumption}

\begin{theorem}[Graphon filter approximation by graph filter] \label{thm:graphon-graph-filter}
Consider the graphon filter given by $Y=\Phi (X; \bbh, \bbW)$ as in \eqref{eqn:spec-graphon_filter}, where $h(\lambda)$ is constant for $|\lambda| < c$ [cf. Fig. \ref{fig:filter_response}]. For the graph filter instantiated from $\Phi (X; \bbh, \bbW)$ as $\bby_n = \Phi (\bbx_n; \bbh, \bbS_n)$ [cf.  \eqref{eqn:det_graph_filter}], under Assumptions \ref{as1} through \ref{as3} it holds
\begin{equation*} \label{eqn:thm4result1}
\|Y-Y_n\|_{L_2} \leq {\sqrt{A_1}}\left(A_2 + \frac{\pi B_{nc}}{\delta_{nc}}\right)n^{-\frac{1}{2}}\|X\|_{L_2} + \frac{2A_3}{\sqrt{3}}n^{-\frac{1}{2}}
\end{equation*}
where $Y_n = \Phi (X_n; \bbh, \bbW_n)$ is the graph filter induced by $\bby_n = \Phi (\bbx_n; \bbh, \bbS_n)$ [cf. \eqref{eqn:graphon_filter_ind}].
%, $n_c$ is the cardinality of the set $\ccalC =\{i\ |\ |\lambda^{n}_i| \geq c\}$, and $\delta_c = \min_{i \in \ccalC} (|\lambda_i - \lambda^{n}_{i+\mbox{\scriptsize sgn}(i)}|,|\lambda_{i+\mbox{\scriptsize sgn}(i)}-\lambda^{n}_i|, |\lambda_1-\lambda^n_{-1}|,|\lambda^n_1-\lambda_{-1}|)$, with $\lambda_i$ and $\lambda^n_i$ denoting the eigenvalues of $\bbW$ and $\bbW_n$ respectively [cf. Fig. \ref{fig:filter_response}]. 
%In particular, if $X=X_n$ we have
%\purple{
%\begin{equation} \label{eqn:thm4result2}
%\|Y-Y_n\|_{L_2} \leq {\sqrt{A_1}}\left(A_2 + \frac{\pi n_c}{\delta_c}\right)n^{-\frac{1}{2}}\|X\|_{L_2} .
%\end{equation}}
\end{theorem}

Theorem \ref{thm:graphon-graph-filter} gives an asymptotic upper bound to the error incurred when approximating graphon filters with graph filters. This bound depends on the filter transferability constant $\sqrt{A_1}(A_2 + \pi B_{nc}/\delta_{nc})n^{-0.5}$, which multiplies $\|X\|$, and on a fixed error term {depending on the variability $A_3$ of $X$ (Assumption \ref{as3})} and corresponding to the difference between $X$ and the graphon signal $X_n$, which is induced by $\bbx_n$. For large $n$, the first term dominates the second. Hence, the quality of the approximation is closely related to the transferability constant. 

Aside from decreasing asymptotically with $n$, the transferability constant depends on the graphon and on the filter parameters. {The dependence on the graphon is due to $A_1$, which is proportional to the graphon variability (Assumption \ref{as1}). The dependence on the filter parameters happens through the constants $A_2$, $B_{nc}$ and $\delta_{nc}$. The first two determine the variability of the filter's spectral response, which is controlled by both the Lipschitz constant $A_2$ (Assumption \ref{as2}) and the length of the band $[c,1]$, as depicted in Fig. \ref{fig:filter_response}. In particular, the number of eigenvalues within this band, given by $B_{nc}$, should satisfy $B_{nc} \ll n$ (i.e. $B_{nc} < \sqrt{n}$). This restriction on the length of the passing band, which is necessary for asymptotic convergence, is a consequence of two facts. The first is that the eigenvalues of the graph converge to those of the graphon \cite[Chapter 11.6]{lovasz2012large} as illustrated in Fig. \ref{fig:eigenvalues_graph_graphon}. The second is that the eigenvalues of the graphon, when ordered in decreasing order of absolute value, accumulate near zero. Combined, these facts imply that, for small eigenvalues, the graph eigenvalues are hard to match to the corresponding graphon eigenvalues, making consecutive eigenvalues difficult to discriminate. As a consequence, filters $h$ with large variation near zero (i.e., small $c$) may modify matching graphon and graph eigenvalues differently, leading to large approximation error.}
Lastly, note that when the $B_{nc} < \sqrt{n}$ requirement is satisfied, asymptotic convergence is guaranteed by convergence of the eigenvalues of $\bbW_n$ to those of $\bbW$ because $\delta_{nc} \to \min_{i\ :\ \lambda^n_i \geq c}|\lambda_i-\lambda_{i+\mbox{\scriptsize sgn(i)}}| \neq 0$, i.e., $\delta_{nc}$ converges to the minimum eigengap of the graphon in the passing band. 

%%%%%%%%%%%%%%%%%%%%%%%%%%%%%%%%%%%%%%%%%%%%%%%%%%%%%%%%%%%%%%%%%%%%%%%%%%%%%%%%
%%%%                              SUBSECTION                                %%%%
%%%%%%%%%%%%%%%%%%%%%%%%%%%%%%%%%%%%%%%%%%%%%%%%%%%%%%%%%%%%%%%%%%%%%%%%%%%%%%%%

\subsection{Graph filter transferability} \label{subsec:transf1}

By the triangle inequality, transferability of graph filters follows directly from Theorem \ref{thm:graphon-graph-filter}.

\begin{figure}[t]
    \centering
    \scalebox{.88}{\input{plots_transferability/filter_response.tex}} 
    \caption{Lipschitz continuous filter with spectral response $h(\lambda)$ constant for $\lambda < c$. The constant band for $\lambda \in [0,c]$ ensures that the filter has the same response for eigenvalues close to zero, which are harder to discriminate. This is necessary to avoid mismatch of the filter response for the graphon and graph eigenvalues in this range.}
    \label{fig:filter_response}
\end{figure}
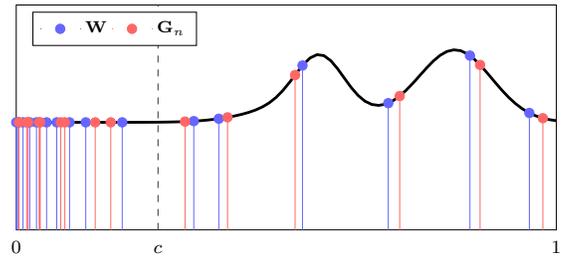

\begin{theorem}[Graph filter transferability]  \label{thm:graph-graph-filter}
Let $\bbG_{n_1}$ and $\bbG_{n_2}$, and $\bbx_{n_1}$ and $\bbx_{n_2}$, be graphs and graph signals obtained from the graphon $\bbW$ and the graphon signal $X$ as in \eqref{eqn:det_graph_filter}, with $n_1 \neq n_2$. Consider the graph filters given by $\bby_{n_1} = \Phi(\bbx_{n_1}; \bbh,\bbS_{n_1})$ and $\bby_{n_2} = \Phi(\bbx_{n_2}; \bbh,\bbS_{n_2})$, and let their shared spectral response $h(\lambda)$ [cf. \eqref{eq:freqResponse}] be constant for $|\lambda| < c$ [cf. Fig. \ref{fig:filter_response}]. Then, under Assumptions \ref{as1} through \ref{as3} it holds
\begin{align*}
\|Y_{n_1}&-Y_{n_2}\|_{L_2} \leq \\
 &\sqrt{A_1}\left(A_2 + \frac{\pi B_{c}}{\delta_{c}}\right)\left({n_1}^{-\frac{1}{2}}+{n_2}^{-\frac{1}{2}}\right)\|X\|_{L_2} \\
&\qquad \qquad \qquad \qquad \qquad \quad + \frac{2A_3}{\sqrt{3}}\left(n_1^{-\frac{1}{2}}+ n_2^{-\frac{1}{2}}\right)
\end{align*}
where $Y_{n_j} = \Phi(X_{n_j}; \bbh, \bbW_{n_j})$ is the graphon filter induced by $\bby_{n_j} = \Phi(\bbx_{n_j}; \bbh, \bbS_{n_j})$ [cf. \eqref{eqn:graphon_filter_ind}], $B_c = \max\{B_{n_1 c},B_{n_2 c}\}$ [cf. Def. \ref{def:c_band_card}] and $\delta_c = \min \{\delta_{n_1 c},\delta_{n_2 c}\}$ [cf. Def. \ref{def:c_eig_margin}].
\end{theorem}

%Relying on the graphon filters induced by $\Phi(\bbx_{n_1}; \bbh,\bbS_{n_1})$ and $\Phi(\bbx_{n_2}; \bbh,\bbS_{n_2})$,
Theorem \ref{thm:graph-graph-filter} upper bounds the difference between the outputs of two identical graph filters on different graphs belonging to the same graphon family. Because this bound decreases asymptotically with $n_1$ and $n_2$, a filter designed for one of these graphs can be transferred to the other with good performance guarantees for large $n_1$ and $n_2$. Beyond values of $n_1$ and $n_2$ satisfying a specific error requirement of, say, $\epsilon$, graph filters are scalable in the sense that they can be applied to any other graph with size $n>\max(n_1,n_2)$ and achieve less than $\epsilon$ error. 
%This is important in problems where the graph size is dynamic, as is the case of recommendation systems for companies with a growing product portfolio [cf. Secs. \ref{sec_reco_systems} and \ref{sec_reco_systems_results}].

The transferability constant in Theorem \ref{thm:graph-graph-filter} is equal to the sum of the transferability constant in Theorem \ref{thm:graphon-graph-filter} for $n=n_1$ and $n=n_2$. %Hence, the same comments about its dependence on the graphon and on the filter's parameters apply. In particular, 
{Even if Theorem \ref{thm:graph-graph-filter} does not require explicitly defining the graphon filter and comparing its spectral response to that of the graph filters, the band $[c,1]$ should be small to guarantee that the filter be able to match the eigenvalues of $\bbG_1$ and $\bbG_2$ and distinguish between consecutive eigenvalues [cf. Fig. \ref{fig:filter_response}]. Therefore, there exists a trade-off between the transferability and discriminability of graph filters.}

%%%%%%%%%%%%%%%%%%%%%%%%%%%%%%%%%%%%%%%%%%%%%%%%%%%%%%%%%%%%%%%%%%%%%%%%%%%%%%%%
%%%%                              SUBSECTION                                %%%%
%%%%%%%%%%%%%%%%%%%%%%%%%%%%%%%%%%%%%%%%%%%%%%%%%%%%%%%%%%%%%%%%%%%%%%%%%%%%%%%%

\subsection{Graphon neural networks} \label{subsec:wnns}

%%%%%%%%%%%%%%%%%%%%%%%%%%%%%%%%%%%%%%%%%%%%%%%%%%%%%%%%%%%%%%%%
%%%   F   I   G   U   R   E   %%%%%%%%%%%%%%%%%%%%%%%%%%%%%%%%%%
%%%%%%%%%%%%%%%%%%%%%%%%%%%%%%%%%%%%%%%%%%%%%%%%%%%%%%%%%%%%%%%%
%
 
%%%%%%%%%%%%%%%%%%%%%%%%%%%%%%%%%%%%%%%%%%%%%%%%%%%%%%%%%%%%%%%%
%%%   F   I   G   U   R   E   %%%%%%%%%%%%%%%%%%%%%%%%%%%%%%%%%%
%%%%%%%%%%%%%%%%%%%%%%%%%%%%%%%%%%%%%%%%%%%%%%%%%%%%%%%%%%%%%%%%
%
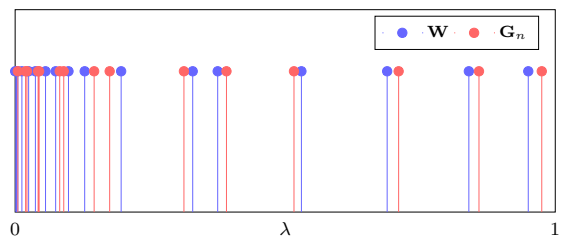
\begin{figure}[t]
    \centering
    \scalebox{.88}{\input{plots_transferability/eigenvalues_graph_graphon.tex}}  
    \caption{Comparison of graphon eigenvalues (blue) and eigenvalues of a graph $\bbG_n$ from a convergent graph sequence (red). As the number of nodes $n$ grows, the eigenvalues of $\bbG_n$ converge to those of $\bbW$.}
    \label{fig:eigenvalues_graph_graphon}
\end{figure}

%%%%%%%%%%%%%%%%%%%%%%%%%%%%%%%%%%%%%%%%%%%%%%%%%%%%%%%%%%%%%%%%
%%%   F   I   G   U   R   E   %%%%%%%%%%%%%%%%%%%%%%%%%%%%%%%%%%
%%%%%%%%%%%%%%%%%%%%%%%%%%%%%%%%%%%%%%%%%%%%%%%%%%%%%%%%%%%%%%%%
%

%Analogously to the definitions of a graphon signal and of the graphon convolution, 
The graphon neural network (WNN) is defined as the limit architecture of a GNN defined on the graphs of a convergent graph sequence. 
%While the WNN processes data supported on graphons, it retains the structure of a GNN by stacking layers of graphon convolutions and nonlinear activation functions.
Denoting the nonlinear activation function $\sigma$, the $\ell$th layer of a multi-layer WNN with $F_\ell=1$ feature per layer (like the GNNs in Sec. \ref{sec_ch8_gnns_multiple_layers}) is given by
\begin{equation}
{
X_{\ell} = \sigma\left(\Phi (X_{\ell-1}; \bbh_\ell, \bbW) \right)
}
\end{equation}
for $1 \leq \ell \leq L$. Note that the input signal at the first layer, $X_0$, is the input data $X$, and the WNN output is given by $Y = X_L$. 

Similarly to the GNN, this WNN can also be written as a map $Y = \Phi(X; \bbH, \bbW)$, where the matrix $\bbH = \{\bbh_\ell\}_{\ell}$ groups the filter coefficients of all layers. Note that the parameters in $\bbH$ are completely independent of the graphon, which is another characteristic WNNs have in common with GNNs.

\subsubsection{Generating GNNs from WNNs} \label{subsec:generating}

An important consequence of the GNN and WNN parametrizations is that, in the maps $\Phi(\bbx; \bbH, \bbS)$ and $\Phi(X; \bbH, \bbW)$, the parameters $\bbH$ can be the same. This allows sampling or evaluating GNNs from a WNN, i.e., the WNN acts as a generating model for GNNs. To see this, consider the WNN $\Phi(X; \bbH, \bbW)$ and define a partition $u_i = (i-1)/n$, $1 \leq i \leq n$, of $[0,1]$. A GNN $\Phi(\bbx_n; \bbH, \bbS_n)$ can be obtained by evaluating the deterministic graph $\bbG_n$ and the deterministic graph signal $\bbx_n$ as in equation \eqref{eqn:det_graph_filter}.

The interpretation of GNNs as instantiantions of a WNN is important because it explicitly disconnects the GNN architecture from the graph. In this interpretation, the graph is not a fixed hyperparameter of the GNN, but a parameter that can be changed according to the underlying graphon and the value of $n$. 
%This has two important consequences. First, it 
This reveals the ability of GNNs to \textit{scale}. 
%Second, it 
It also allows GNNs to be adapted both by optimizing the weights in $\bbH$ and by changing the graph $\bbG_n$, which adds degrees of freedom to the architecture at no additional computational cost.

WNNs induced by GNNs can also be defined. The WNN induced by a GNN $\Phi(\bbx_n; \bbH, \bbS_n)$ is given by $\Phi(X_n; \bbH, \bbW_n)$ where $\bbW_n$, the graphon induced by $\bbG_n$, and $X_n$, the graphon signal induced by $\bbx_n$, are as in \eqref{eqn:graphon_filter_ind}.
This definition allows establishing a direct comparison both between GNNs and WNNs and between GNNs on graphs of different sizes.

%In particular, provided that the filter band $[c,1]$ is narrow enough ($n_c \ll \min (n_1,n_2)$),  

%is important because it relates the performance of two GNNs which, unlike WNNs, are both realizable in practice. 

\subsubsection{Approximating WNNs with GNNs} \label{subsec:approximating}

For large $n$, we can expect the GNNs instantiated from a WNN to become closer to the WNN itself at a similar rate at which the graphs $\bbG_n$ converge to $\bbW$. As such, the outputs of the GNN and WNN maps $\Phi(\bbx_n; \bbH, \bbS_n)$ and $\Phi(X; \bbH, \bbW)$ should also grow closer, allowing the GNN to be used as a proxy for the WNN. To evaluate the quality of this approximation for different values of $n$, the outputs of $\Phi(\bbx_n; \bbH, \bbS_n)$ and $\Phi(X; \bbH, \bbW)$ must be compared. This is done by considering the WNN induced by $\Phi(\bbx_n; \bbH, \bbS_n)$ and given by $Y_n = \Phi(X_n; \bbH, \bbW_n)$ [cf. \eqref{eqn:graphon_filter_ind}]. Under Assumption \ref{as4}, the following theorem from \cite{ruiz2020wnn} holds.
\begin{assumption} \label{as4}
The activation functions are normalized Lipschitz, i.e. $|\sigma(x)-\sigma(y)| \leq |x-y|$, and $\sigma(0)=0$.
\end{assumption} 

%%%%%%%%%%%%%%%%%%%%%%%%%%%%%%%%%%%%%%%%%%%%%%%%%%%%%%%%%%%%%%%%%%%%%%%%%%%%%%%%
%   F   I   G   U   R   E   %%%%%%%%%%%%%%%%%%%%%%%%%%%%%%%%%%%%%%%%%%%%%%%%%%%%
%%%%%%%%%%%%%%%%%%%%%%%%%%%%%%%%%%%%%%%%%%%%%%%%%%%%%%%%%%%%%%%%%%%%%%%%%%%%%%%%
%
\begin{figure*}[h]
\centering
\begin{subfigure}[b]{.32\linewidth}
\centering
\vspace{0.5cm}
\includegraphics[width=0.9\textwidth, height=0.62\textwidth]{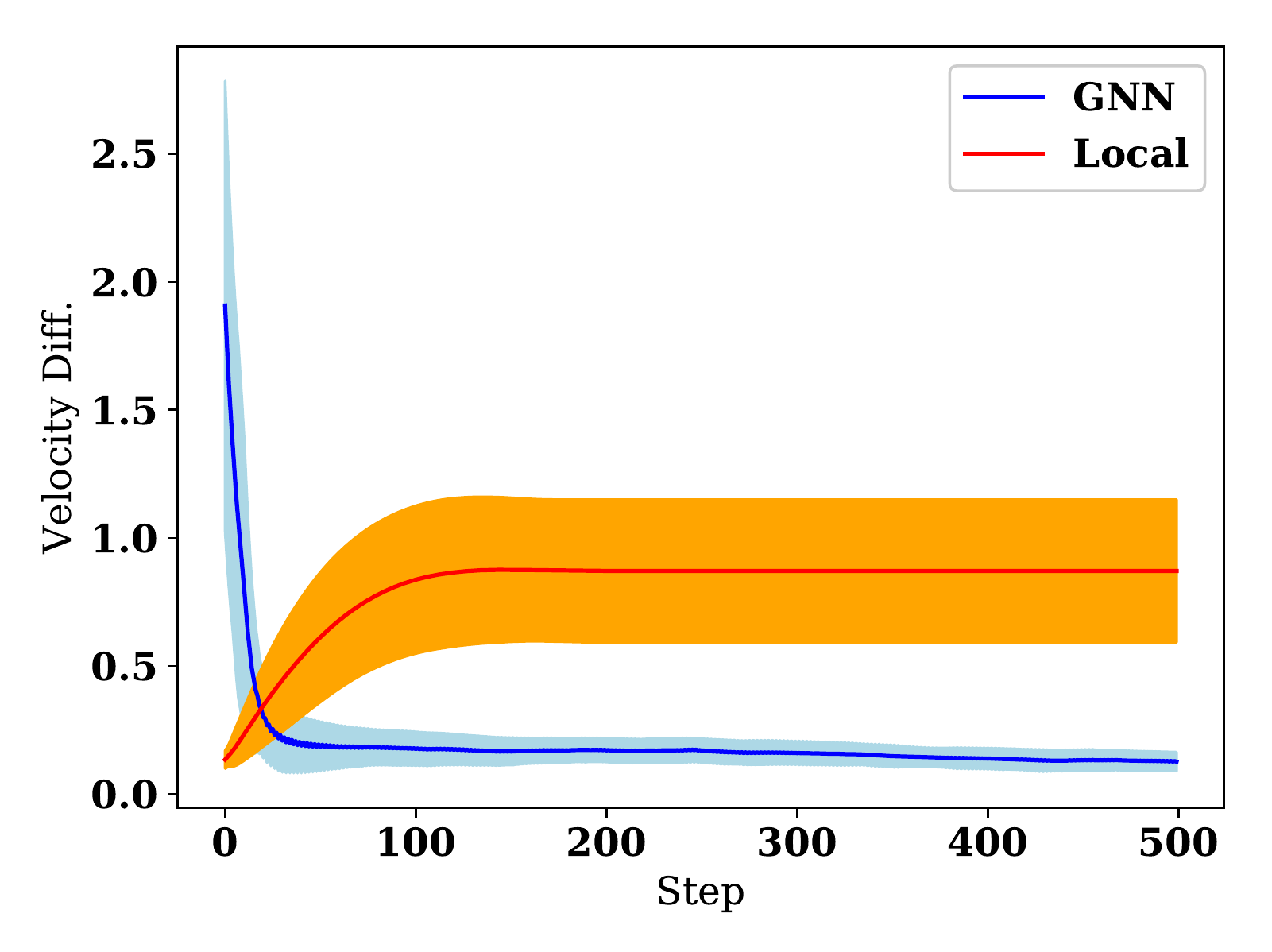} 
\subcaption{}\label{fig:vel}
\end{subfigure}%
\begin{subfigure}[b]{.32\linewidth}
\centering
\includegraphics[width=0.9\textwidth]{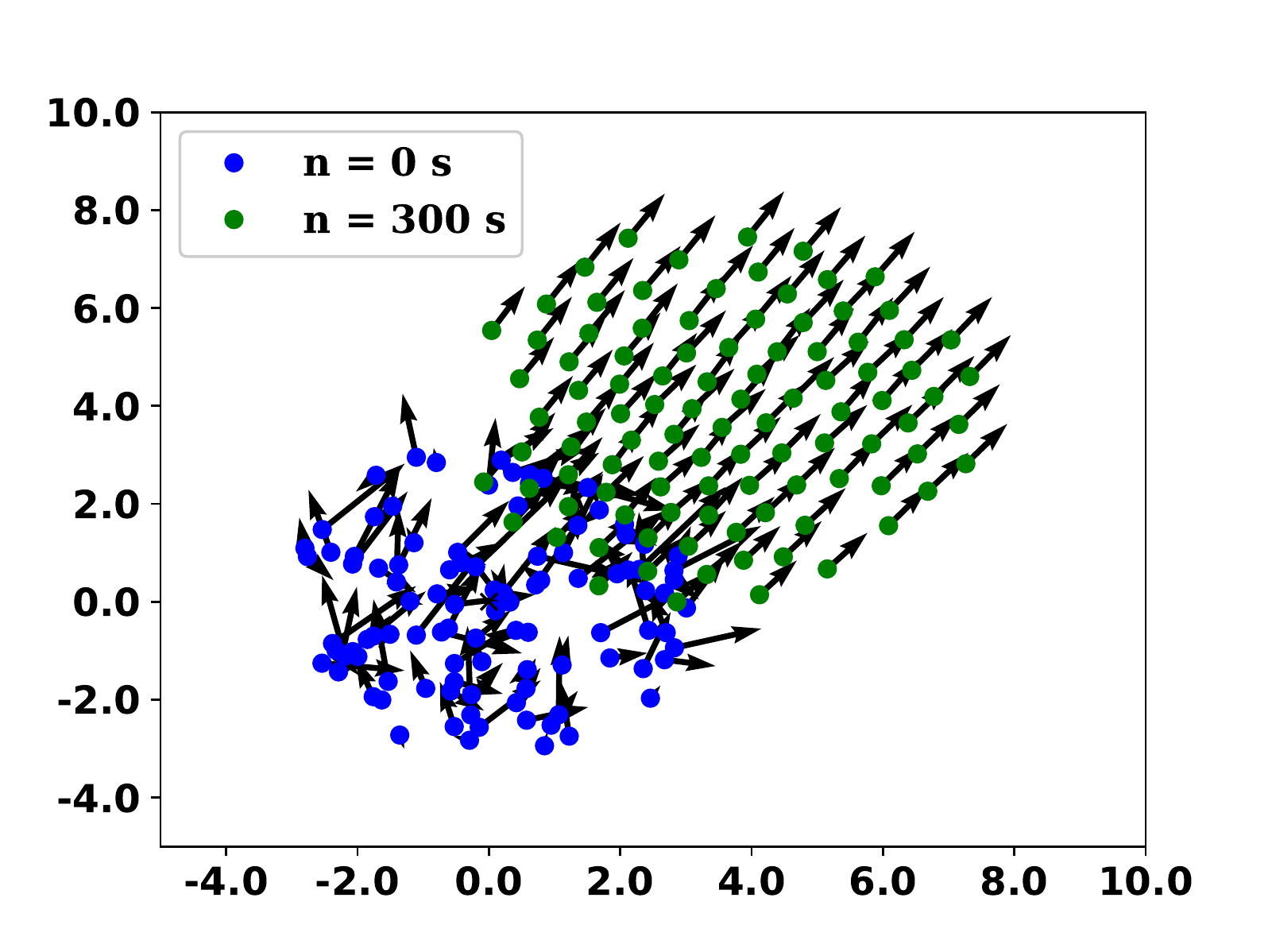}
\subcaption{}\label{fig:grid_traj300}
\end{subfigure}%
\begin{subfigure}[b]{.32\linewidth}
\centering
\includegraphics[width=0.9\textwidth]{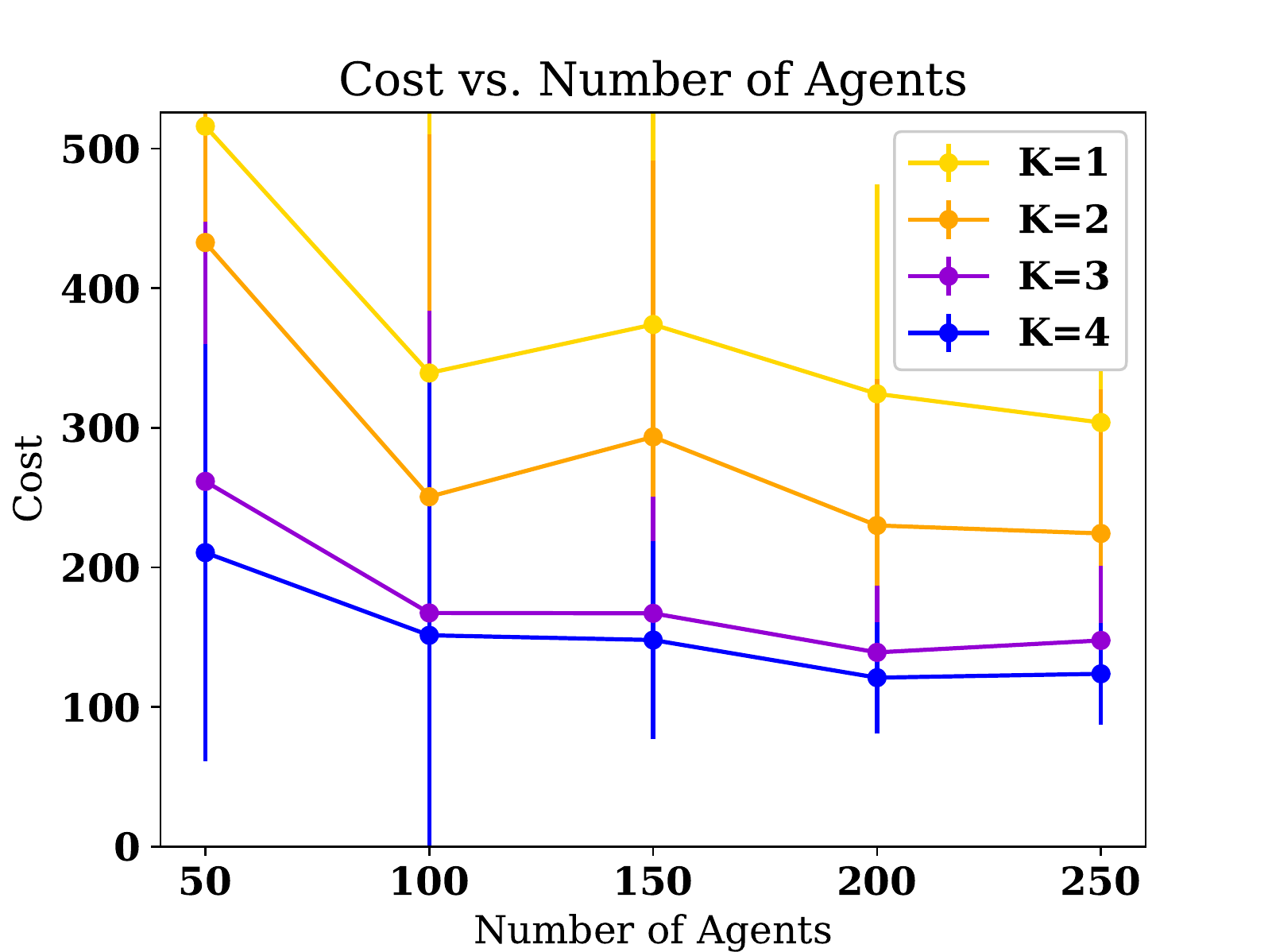}
\subcaption{}\label{fig:transfer_grid}
\end{subfigure}
\caption{The GNN maintains a cohesive flock, while the local controller allows the flock to scatter. (a) Average difference in velocities. Local stands for $K=0$. 
(b) Flock positions using the GNN. (c) Cost vs. number of agents.
\vspace{-0.5cm}
}
\label{fig:flocking_transfer}
\end{figure*} 

{This assumption is satisfied for most conventional nonlinearities, e.g. ReLU and hyperbolic tangent.}

\begin{theorem}[WNN approximation by GNN]  \label{thm:graphon-graph}
Consider the $L$-layer WNN given by $Y=\Phi(X; \bbH, \bbW)$, where {$F_\ell=1$} for $1 \leq \ell \leq L$. Let the graphon convolutions $h(\lambda)$ [cf. \eqref{eqn:spec-graphon_filter}] be such that $h(\lambda)$ is constant for $|\lambda| < c$ [cf. Fig. \ref{fig:filter_response}]. For the GNN instantiated from this WNN as $\bby_n = \Phi(\bbx_n; \bbH, \bbS_n)$ [cf.  \eqref{eqn:det_graph_filter}], under Assumptions \ref{as1} through \ref{as4} it holds
{
\begin{align*}
\|Y_n-Y\|_{L_2} &\leq \\ L&\sqrt{A_1}\left(A_2 + \frac{\pi B_{nc}}{\delta_{nc}}\right)n^{-\frac{1}{2}}\|X\|_{L_2} + \frac{A_3}{\sqrt{3}}n^{-\frac{1}{2}}
\end{align*}
}
where $Y_n = \Phi(X_n; \bbH, \bbW_n)$ is the WNN induced by $\bby_n = \Phi(\bbx_n; \bbH, \bbS_n)$ [cf. \eqref{eqn:graphon_filter_ind}].
% $n_c$ is the cardinality of the set $\ccalC =\{i\ |\ |\lambda^{n}_i| \geq c\}$, and $\delta_c = \min_{i \in \ccalC} (|\lambda_i - \lambda^{n}_{i+\mbox{\scriptsize sgn}(i)}|,|\lambda_{i+\mbox{\scriptsize sgn}(i)}-\lambda^{n}_i|, |\lambda_1-\lambda^n_{-1}|,|\lambda^n_1-\lambda_{-1}|)$, with $\lambda_i$ and $\lambda^n_i$ denoting the eigenvalues of $\bbW$ and $\bbW_n$ respectively.
\end{theorem}

Given a graph $\bbG_n$ and a signal $\bbx_n$ obtained from $\bbW$ and $X$ as in \eqref{eqn:det_graph_filter}, the GNN $\Phi(\bbx_n; \bbH, \bbS_n)$ can approximate the WNN $\Phi(\bbX; \bbH, \bbW)$ with an error that decreases asymptotically with $n$. This error is upper bounded by a term proportional to the input, controlled by the \textit{transferability constant} ${L\sqrt{A_1}}\left(A_2 + {(\pi B_{nc})}/{\delta_{nc}}\right)n^{-0.5}$, and by a fixed error term given by $A_3/\sqrt{3n}$. The fixed error term is a truncation error due to ``discretizing'' $X$ to obtain $\bbx_n$. Besides the dependence on the graphon and on the filter parameters, the transferability constant also depends on $L$. As for the constants $A_1$, $A_2$, $B_{nc}$ and $\delta_{nc}$, the same comments as in Theorem \ref{thm:graphon-graph-filter} apply. %The approximation error is better for low $A_1$, i.e., smooth graphons. Additionally, 

\subsection{GNN transferability} \label{subsec:transf2}

By Theorem \ref{thm:graphon-graph} and the triangle inequality, the following theorem from \cite{ruiz2020wnn} holds.

\begin{theorem}[GNN transferability]  \label{thm:graph-graph}
Let $\bbG_{n_1}$ and $\bbG_{n_2}$, and $\bbx_{n_1}$ and $\bbx_{n_2}$, be graphs and graph signals obtained from the graphon $\bbW$ and the graphon signal $X$ as in \eqref{eqn:det_graph_filter}, with $n_1 \neq n_2$. Consider the $L$-layer GNNs given by $\Phi(\bbx_{n_1};\bbH,\bbS_{n_1})$ and $\Phi(\bbx_{n_2}; \bbH, \bbS_{n_2})$, where $F_\ell=1$ for $1 \leq \ell \leq L$. Let the graph convolutions $h(\lambda)$ [cf. \eqref{eq:freqResponse}] be such that $h(\lambda)$ is constant for $|\lambda| < c$. Then, under Assumptions \ref{as1} through \ref{as4} it holds
\begin{align*}
\|Y_{n_1}&-Y_{n_2}\|_{L_2} \leq \\
 &L\sqrt{A_1}\left(A_2 + \frac{\pi B_{c}}{\delta_{c}}\right)\left({n_1}^{-\frac{1}{2}}+{n_2}^{-\frac{1}{2}}\right)\|X\|_{L_2} \\
&\qquad \qquad \qquad \qquad \qquad \quad + \frac{A_3}{\sqrt{3}}\left(n_1^{-\frac{1}{2}}+ n_2^{-\frac{1}{2}}\right)
\end{align*}
where $Y_{n_j} = \Phi(X_{n_j}; \bbH, \bbW_{n_j})$ is the WNN induced by $\bby_{n_j} = \Phi(\bbx_{n_j}; \bbH, \bbS_{n_j})$ [cf. \eqref{eqn:graphon_filter_ind}], $B_c = \max\{B_{n_1 c},B_{n_2 c}\}$ [cf. Def. \ref{def:c_band_card}] and $\delta_c = \min \{\delta_{n_1 c},\delta_{n_2 c}\}$ [cf. Def. \ref{def:c_eig_margin}].
%$n_c' = \max_{j \in\{1,2\}} |\ccalC_j|$ is the maximum cardinality of the sets $\ccalC_j = \{i\ |\ |\lambda^{n_j}_i| \geq c\}$, and $\delta_c' = \min_{i \in \ccalC_j, j \in \{1,2\}}(|\lambda_i - \lambda^{n_j}_{i+\mbox{\scriptsize sgn}(i)}|,|\lambda_{i+\mbox{\scriptsize sgn}(i)}-\lambda^{n_j}_i|, |\lambda_1-\lambda^{n_j}_{-1}|,|\lambda^{n_j}_1-\lambda_{-1}|)$, with $\lambda_i$ and $\lambda^{n_j}_i$ denoting the eigenvalues of $\bbW$ and $\bbW_{n_j}$ respectively.
\end{theorem}

Theorem \ref{thm:graph-graph} proves that GNNs are transferable between graphs of different sizes belonging to the same graphon family. This has two important implications. If the GNN hyperparameters are chosen carefully, the GNN can be transferred from the graph on which it was trained to another graph with error bound inversely proportional to the sizes of both graphs. In scenarios where the same task has to be replicated on different graphs, e.g., operating the same type of sensor network on multiple plants, this is key because it avoids retraining the GNN.
This result also implies that GNNs, like graph filters, are scalable. They can be trained on smaller graphs than the graphs on which they are deployed (and vice-versa), and are robust to increases in the graph size.
% that are common, for instance, in recommendation systems with a dynamic product base. 
%The obvious advantage, in this case, is that training GNNs on small graphs is easier than training them on large graphs.

The approximation error is given by the transferability constant ${LF^{L-1}\sqrt{A_1}}(A_2 + {\pi B_{c}}/{\delta_{c}})({n_1}^{-0.5}+{n_2}^{-0.5})$ and the fixed error term $A_3(n_1^{-0.5}+{n_2}^{-0.5})/\sqrt{3}$, both of which decrease asymptotically with $n_1$ and $n_2$. The fixed error term measures how different the graph signals $\bbx_{n_1}$ and $\bbx_{n_2}$ are from the graphon signal $X$, therefore, its contribution is small. The transferability constant, on the other hand, is determined by the graphon variability $A_1$, the number of layers $L$ and the convolutional filter parameters $A_2$, $B_c$ and $\delta_c$. {Except for $A_1$, all of these can be tuned. In order to have an asymptotic bound for $n_2>n_1$, the number of eigenvalues in the band $[c,1]$ must satisfy $B_c < \sqrt{n_1}$ [cf. Fig. \ref{fig:filter_response}]. This restriction is necessary to avoid mismatching the filter response for small eigenvalues of $\bbG_{n_1}$ and $\bbG_{n_2}$, which become harder to discriminate as they accumulate around zero [cf. Fig. \ref{fig:eigenvalues_graph_graphon}].} As long as this condition is satisfied, the bound converges asymptotically because, as $n_1, n_2 \to \infty$, $\delta_c$ converges to the minimum eigengap of the graphon in the passing band. 

%Because it requires a restriction on $n_c'$ to be asymptotic, 
The transferability bound in Theorem \ref{thm:graph-graph} thus reflects a similar trade-off between transferability and discriminability to that observed for graph filters. However, in the case of GNNs this is partially overcome by the addition of nonlinearities. Nonlinearities act as rectifiers which \textit{scatter} some spectral components associated with small $\lambda$ around the middle range of the spectrum. This makes for an interesting parallel with the role of nonlinearities in stability, which depends on the components associated with large eigenvalues being scattered around the lower range of the spectrum instead.

%--- if two GNNs can be made arbitrarily close to the same WNN, then they can be made arbitrarily close to one another. %We state it here as Theorem \ref{thm:graph-graph}, which retains the same assumptions from Theorem \ref{thm:graphon-graph}.

%\newpage

%% file: plots_transferability/eigenvalues_graphon.tex
%!TEX root = ../stability.tex

\pgfplotsset{xtick style={draw=none}}

\def \thisplotscale {3.4}
\def \unit {\thisplotscale cm}

\def \frequencyresponse 
     {   0.8}

\begin{tikzpicture}[x = 1*\unit, y=1*\unit]
\begin{axis}[scale only axis,
             width  = 2.4*\unit,
             height = 0.9*\unit,
             xmin = 0, xmax=8,
             xtick = {0, 0.3, 1.03, 1.57, 
              2.63, 3, 4.24, 5.51, 6.72, 7.6, 8},
             xticklabels = {\black{\footnotesize $0$},
			                \blue{\footnotesize $\ \ \ldots$},
             				\blue{\footnotesize $\ \lam_8$}, 
                            \blue{\footnotesize $\ \lam_7$},
                            \blue{\footnotesize $\ \lam_6$}, 
                            \blue{\footnotesize $\ \lam_5$}, 
                            \blue{\footnotesize $\ \lam_4$}, 
                            \blue{\footnotesize $\ \lam_3$},
                            \blue{\footnotesize $\ \lam_2$}, 
                            \blue{\footnotesize $\ \lam_1$},
                            \black{\footnotesize $1$}},
             ymin = -0, ymax = 1.15,
             ytick = {-1},
             typeset ticklabels with strut,
             enlarge x limits=false]

\addplot+[samples at = {0.00, 0.02, 0.04, 0.1, 		              0.2, 0.3, 0.45, 0.6, 0.79, 1.03, 1.57, 
              2.63, 3, 4.24, 5.51, 6.72, 7.6}, 
          color = blue!60, 
          ycomb, 
          mark=otimes*, 
          mark options={blue!60}]
         {\frequencyresponse};

%\addplot[ domain=-0.5:7.5, 
%          samples = 80, 
%          color = black,
%          line width = 1.2]
%         {\frequencyresponse};

\end{axis}
\end{tikzpicture}

%{-0.22, 0.25, 0.61, 1.05, 1.37, 2.11, 
%  2.43, 3.07, 3.67, 4.13, 4.31, 5.05, 
%  5.20, 6.00, 6.42, 7.00}

%% file: plots_transferability/eigenvalues_graph_graphon.tex
%!TEX root = ../stability.tex

\pgfplotsset{xtick style={draw=none}}

\def \thisplotscale {3.4}
\def \unit {\thisplotscale cm}

\def \frequencyresponse 
     {0.8}

\begin{tikzpicture}[x = 1*\unit, y=1*\unit]
\begin{axis}[scale only axis,
             width  = 2.4*\unit,
             height = 0.9*\unit,
             xmin = 0, xmax=8,
             xtick = {0, 4, 8},
             xticklabels = {\black{\footnotesize $0$},
             				\black{\footnotesize $\lambda$}, 
                            \black{\footnotesize $1$}},
             ymin = -0, ymax = 1.15,
             ytick = {-1},
             typeset ticklabels with strut,
             enlarge x limits=false,
             legend pos=north east,
             legend columns=3]

%\addlegendimage{empty legend}

\addplot+[samples at = {0.00, 0.02, 0.04, 0.1, 
			  0.2, 0.3, 0.45, 0.6, 0.79, 1.03, 1.57, 
              2.63, 3, 4.24, 5.51, 6.72, 7.6}, 
          color = blue!60, 
          ycomb, 
          mark=otimes*, 
          mark options={blue!60}]
         {\frequencyresponse};
         
\addplot+[samples at = {0.04, 0.175, 0.155, 0.34, 0.355, 
              0.66, 0.72, 1.17, 1.4, 2.5, 3.13, 4.13, 
              5.68, 6.87, 7.8}, 
          color = red!60, 
          ycomb, 
          mark=otimes*, 
          mark options={red!60}]
         {\frequencyresponse};

%\addlegendentry{}
   \addlegendentry{\scriptsize $\bbW$}
   \addlegendentry{\scriptsize $\bbG_n$}

%\addplot[ domain=-0.5:7.5, 
%          samples = 80, 
%          color = black,
%          line width = 1.2]
%         {\frequencyresponse};

\end{axis}
\end{tikzpicture}

%{-0.22, 0.25, 0.61, 1.05, 1.37, 2.11, 
%  2.43, 3.07, 3.67, 4.13, 4.31, 5.05, 
%  5.20, 6.00, 6.42, 7.00}

%% file: plots_transferability/filter_response.tex
%!TEX root = ../graphons.tex

\pgfplotsset{xtick style={draw=none}}
\pgfplotsset{ytick style={draw=none}}

\def \thisplotscale {3.4}
\def \unit {\thisplotscale cm}

\def \frequencyresponse 
     { 0.1*exp(-(1*(x-4.3))^2) 
       + 0.25*exp(-(2*(x-4.5))^2) 
       + 0.37*exp(-(1.3*(x-6.5))^2) 
       + 0.55}

\begin{tikzpicture}[x = 1*\unit, y=1*\unit]
\def \factorx {3/8}
\def \deltax  {0.5*\factorx}
\def \shadeshift  {0.05}

%\path [fill=black, opacity = 0.1] 
 %             (0.0, 0.55) --
 %             (0.63, 0.75) --  
 %             (0.63, 0.35) -- cycle;

\begin{axis}[scale only axis,
             width  = 2.4*\unit,
             height = 1*\unit,
             xmin = 0, xmax=8,
             xtick = {0, 2.1, 8},
             xticklabels = {\black{\footnotesize $0$},
             				\black\footnotesize {$c$}, 
                            \black{\footnotesize $1$}},
             ymin = -0, ymax = 1.15,
             ytick = {2},
             %yticklabels = {\black{\footnotesize $h(0)$}},
%             ytick = {0.928, 0.55},
%             yticklabels = {\black{\footnotesize $h_{\mbox{\tiny min}}+Lc$},
%             				\black{\footnotesize $h_{\mbox{\tiny min}}$}},
%             yticklabel style={rotate=90},
             typeset ticklabels with strut,
             enlarge x limits=false,             					legend pos=north west,
             legend columns=3]

\addplot+[samples at = {0.00, 0.02, 0.04, 0.1, 		              
			  0.2, 0.3, 0.45, 0.6, 0.79, 1.03, 1.57, 
              2.63, 3, 4.24, 5.51, 6.72, 7.6}, 
          color = blue!60, 
          ycomb, 
          mark=otimes*, 
          mark options={blue!60}]
         {\frequencyresponse};
         
\addplot+[samples at = {0.04, 0.175, 0.155, 0.34, 0.355, 
              0.66, 0.72, 1.17, 1.4, 2.5, 3.13, 4.13, 
              5.68, 6.87, 7.8}, 
          color = red!60, 
          ycomb, 
          mark=otimes*, 
          mark options={red!60}]
         {\frequencyresponse};
         
\addplot+[samples at = {2.1}, 
          color = black!80, 
          dashed,
          ycomb,
          mark=none]
         {1.2};

\addplot[ domain=0:8, 
          samples = 80, 
          color = black,
          line width = 1.2]
         {\frequencyresponse};

%\addplot[ domain=0:1.95, 
%          samples = 80, 
%          color = black!30,
%          line width = 0.4]
%         {0.4};
         
%\addplot[ domain=0:2.1, 
%		  dashed,
%          samples = 80, 
%          color = black!80,
%          line width = 0.5]
%         {0.928-0.175*x};

   \addlegendentry{\scriptsize $\bbW$}
   \addlegendentry{\scriptsize $\bbG_n$}

\end{axis}

\end{tikzpicture}

%{-0.22, 0.25, 0.61, 1.05, 1.37, 2.11, 
%  2.43, 3.07, 3.67, 4.13, 4.31, 5.05, 
%  5.20, 6.00, 6.42, 7.00}

%% file: LR-numerical.tex
% !TEX root = root.tex

%%%%%%%%%%%%%%%%%%%%%%%%%%%%%%%%%%%%%%%%%%%%%%%%%%%%%%%%%%%%%%%%%%%%%%%%%%%%%%%%
%   S   E   C   T   I   O   N   %%%%%%%%%%%%%%%%%%%%%%%%%%%%%%%%%%%%%%%%%%%%%%%%
%%%%%%%%%%%%%%%%%%%%%%%%%%%%%%%%%%%%%%%%%%%%%%%%%%%%%%%%%%%%%%%%%%%%%%%%%%%%%%%%
%
\section{Decentralized Collaborative Systems} \label{sec_dcis}

GNNs have been applied with success to learn decentralized control policies \cite{Tolstaya19-Flocking, Li20-Planning}. Consider then a team of $n$ agents that endeavor to accomplish a shared goal. Each agent has access to local states $\bbx_i$ and has to produce local control actions $\bba_i$. Agent proximity determines the ability to exchange information between pairs of agents and results in access to delayed information about the state of the system. If agents $i$ and $j$ are separated by $k$ communication hops they know about their respective states with a delay of $k$ time units. We capture this limitation with the definition of the information history of agent $i$,
\begin{equation} \label{eqn:partialInformation}
    \ccalX_{i}(t) 
       = \bigcup_{k=0}^{K-1} 
          \Big\{ 
             \bbx_{j}(t-k) 
                : j \in \ccalN_{i}^{k}(t) 
                   \Big\} .
\end{equation}
As per \eqref{eqn:partialInformation}, agent $i$ has access to its current state $\bbx_i(t)$, but only knows the states of $k$-hop neighbors at time $t-k$. A decentralized controller is one in which actions $\bba_i(t)$ are functions of the history $\ccalX_{i}(t)$. It is notable that the graph filters in \eqref{eqn_ch8_gnn_recursion_filter_matrix_diffusion_sequence} can be modified to have this property. Doing so requires that we rewrite \eqref{eqn_ch8_gnn_recursion_filter_matrix_diffusion_sequence} in terms of a diffusion sequence that takes time delays into consideration. Thus, replace $\bbZ_{lk}$ in \eqref{eqn_ch8_gnn_recursion_filter_matrix_diffusion_sequence} by $\bbZ_{lk}(t)$ defined as
\begin{equation} \label{eqn_gnn_recursion_filter_dcs}
   \bbZ_{lk}(t) 
      = \bbS \bbZ_{l,k-1}(t-1), 
           \quad 
              \text{with~} 
                 \bbZ_{l0}(t) = \bbX_{l}(t) .
\end{equation}
This is the same as \eqref{eqn_MIMO_diffusion_sequence} except for the use of time delays to respect the information structure described by \eqref{eqn:partialInformation}.

GNNs have proven successful in learning policies for flocking \cite{Tolstaya19-Flocking} and collaborative navigation \cite{Li20-Planning}. We describe here some flocking results from \cite{Tolstaya19-Flocking}. In this scenario we are given a team of $n$ agents with random initial positions and velocities. The goal is for agents to form a cohesive flock in which: (i) They all move with the same velocity. (ii) There are no collisions between agents. To solve this problem we consider local states $\bbx_{i}(t) \in \reals^{6}$ with components,
\begin{align} \label{eqn:flockingState}
   \bbx^T_{i}(t) 
      = \Bigg[ 
           \sum_{j \in \ccalN_{i}} 
              \bbv_{ij}(t) ;
                 \sum_{j \in \ccalN_{i}(t)} 
                    \frac{\bbr_{ij}}
                       {\| \bbr_{ij}(t)\|^{4}} ; 
                          \sum_{j \in \ccalN_{i}} 
                             \frac{\bbr_{ij}(t)}
                                {\| \bbr_{ij}(t)\|^{2}}
                                   \Bigg]
\end{align}
In \eqref{eqn:flockingState}, $\bbr_{ij}(t)$ and $\bbv_{ij}(t)$ denote the positions and velocities of agent $j$ measured relative to the position and velocity of agent $i$, respectively. The neighborhood $\ccalN_i$ is made up of nodes $j$ for which the distance $\|\bbr_{ij}\|\leq R$. The distance $R$ represents a communication and sensing radius. The components of the state in \eqref{eqn:flockingState} are somewhat arbitrary. They are motivated by their use in a benchmark decentralized controller \cite{tanner2003stable}.

It is important to observe that an optimal centralized controller is trivial as we can just order all the agents to move in the same direction. The optimal decentralized controller is, however, unknown. We therefore choose to train a decentralized GNN to mimic the centralized controller while respecting the information structure in \eqref{eqn:partialInformation}. While perfect mimicry is not attained we do oberve improvement relative to existing decentralized controllers. This is illustrated in Figs. \ref{fig:flocking_transfer}a and \ref{fig:flocking_transfer}b where we show the velocities for a swarm that is controlled with a GNN and a swarm that is controlled with the decentralized controller in \cite{tanner2003stable}. A more comprehensive evaluation is shown in Fig. \ref{fig:flocking_transfer}c where we illustrate the cost that is attained by different GNN architectures as we vary the flock size. The ability to attain small cost for large swarms is worth emphasizing. 

Since we are training to mimic a centralized controller, the training of the GNN is an offline process. This fact implies that the networks that are observed during training and the networks that are observed during execution are different. This is not expected to be an issue because of the stability and transferability results of Sections \ref{sec:stability} and \ref{sec:transferability}. The numerical results in Figure \ref{fig:flocking_transfer} corroborate that this is true. 

%%%%%%%%%%%%%%%%%%%%%%%%%%%%%%%%%%%%%%%%%%%%%%%%%%%%%%%%%%%%%%%%%%%%%%%%%%%%%%%%
%   S   E   C   T   I   O   N   %%%%%%%%%%%%%%%%%%%%%%%%%%%%%%%%%%%%%%%%%%%%%%%%
%%%%%%%%%%%%%%%%%%%%%%%%%%%%%%%%%%%%%%%%%%%%%%%%%%%%%%%%%%%%%%%%%%%%%%%%%%%%%%%%
%
\section{Wireless Communication Networks} \label{sbs:transf}
GNNs have also been applied with success to learn optimal resource allocations in wireless communication networks \cite{Eisen20-REGNN}. Consider an ad-hoc wireless network with $n$ transmitter and receiver pairs indexed by $i\in\{1,n\}$. Wireless link states are represented with fading coefficients $s_{ij} \in \reals_{+}$, which denote the fading state between a transmitter $i$ and a receiver $j$. The fading channel $s_{ii}$ connects transmitter $i$ to its intended receiver $j$. The fading channel $s_{ij}$ with $j\neq i$ links $i$ to other receivers on which the transmission of $i$ manifests as interference. All channels are arranged in the matrix $\bbS \in\reals^{m\times m}$. The goal is to map fading state observations $\bbS$ to power allocations $\bbp := [p_1; \hdots; p_m] = \bbp(\bbS)$. The combination of channel realizations $\bbS$ and power allocations $\bbp(\bbS)$ determines the communication rate between each transmitter-receiver pair. For instance, if using  capacity achieving codes without interference cancellation, rates are determined by the function 
\begin{align}\label{eq_interference_problem1}
  f_i(\bbp; \bbS) 
     := \log \bigg( 1 + 
           \frac{s_{ii} p_i(\bbS)}
              {1 + \sum_{j \neq i} s_{ji} p_j(\bbS)}\bigg).
\end{align}
The expression in \eqref{eq_interference_problem1} represents an instantaneous performance metric. It is customary to focus on the long term performance given by the expectation $\mbE[f_i(\bbp; \bbS)]$ over realizations of the fading channels $\bbS$. A particular problem of interest is the maximization of the expected sum rate, which leads to the optimal power allocation being given by  
\begin{alignat}{3} \label{eq_param_problem}
   \bbp^*(\bbS) 
      ~=~ \argmax 
             \sum_{i=1}^n 
                \mbE\Big[\,
                   f_i(\bbp; \bbS) 
                      \,\Big].
\end{alignat}
The problem in \eqref{eq_param_problem} is a statistical risk minimization problem of the form in \eqref{sec_ch8_learning}. We advocate its solution with a GNN and therefore choose to parametrize the power allocation as a $\bbp(\bbS) = \Phi(\bbx;\ccalH,\bbS)$. The important observation to make is that in \eqref{eq_param_problem} we want to find a power allocation $\bbp(\bbS)$ associated to each fading realization $\bbS$. Thus, we are reinterpreting the shift operator $\bbS$ as an input to the GNN. To emphasize this fact we say that the parametrization is a random-edge (RE)GNN. There is also no input $\bbx$ in \eqref{eq_param_problem}. We can therefore set $\bbx=\bbone$ in the GNN parametrization.  

%%%%%%%%%%%%%%%%%%%%%%%%%%%%%%%%%%%%%%%%%%%%%%%%%%%%%%%%%%%%%%%%%%%%%%%%%%%%%%%%
%   F   I   G   U   R   E   %%%%%%%%%%%%%%%%%%%%%%%%%%%%%%%%%%%%%%%%%%%%%%%%%%%%
%%%%%%%%%%%%%%%%%%%%%%%%%%%%%%%%%%%%%%%%%%%%%%%%%%%%%%%%%%%%%%%%%%%%%%%%%%%%%%%%
%
\begin{figure}[t]
\centering
\includegraphics[width=0.85\linewidth]{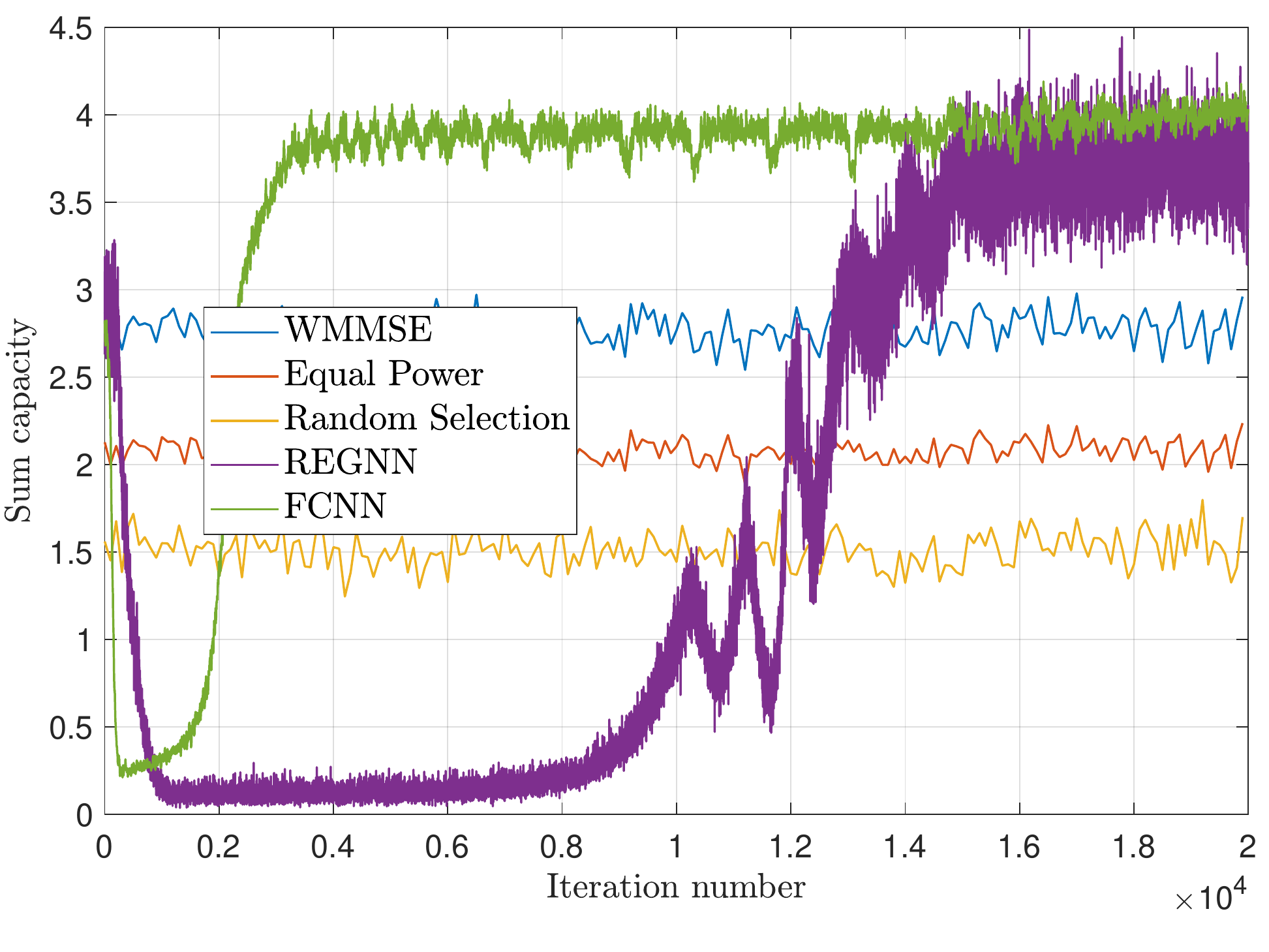}
\caption{Performance of GNN during training for $m=20$ pairs, in comparison with FCNN and three heuristic baselines: WMMSE \cite{shi2011iteratively}, equal power division across all users and across a random subset of users.
%Mean and percentile sum-rates achieved by the GNN trained on network of size $n=50$ and executed on networks of increasing size $n^{\prime}$ in comparison to heuristic baselines: WMMSE \cite{shi2011iteratively}, equal power division across all users and across a random subset of users.
}
\label{fig_compare}
\end{figure}

%%%%%%%%%%%%%%%%%%%%%%%%%%%%%%%%%%%%%%%%%%%%%%%%%%%%%%%%%%%%%%%%%%%%%%%%%%%%%%%%
%   F   I   G   U   R   E   %%%%%%%%%%%%%%%%%%%%%%%%%%%%%%%%%%%%%%%%%%%%%%%%%%%%
%%%%%%%%%%%%%%%%%%%%%%%%%%%%%%%%%%%%%%%%%%%%%%%%%%%%%%%%%%%%%%%%%%%%%%%%%%%%%%%%
%
\begin{figure}[t]
\centering
\includegraphics[width=0.85\linewidth]{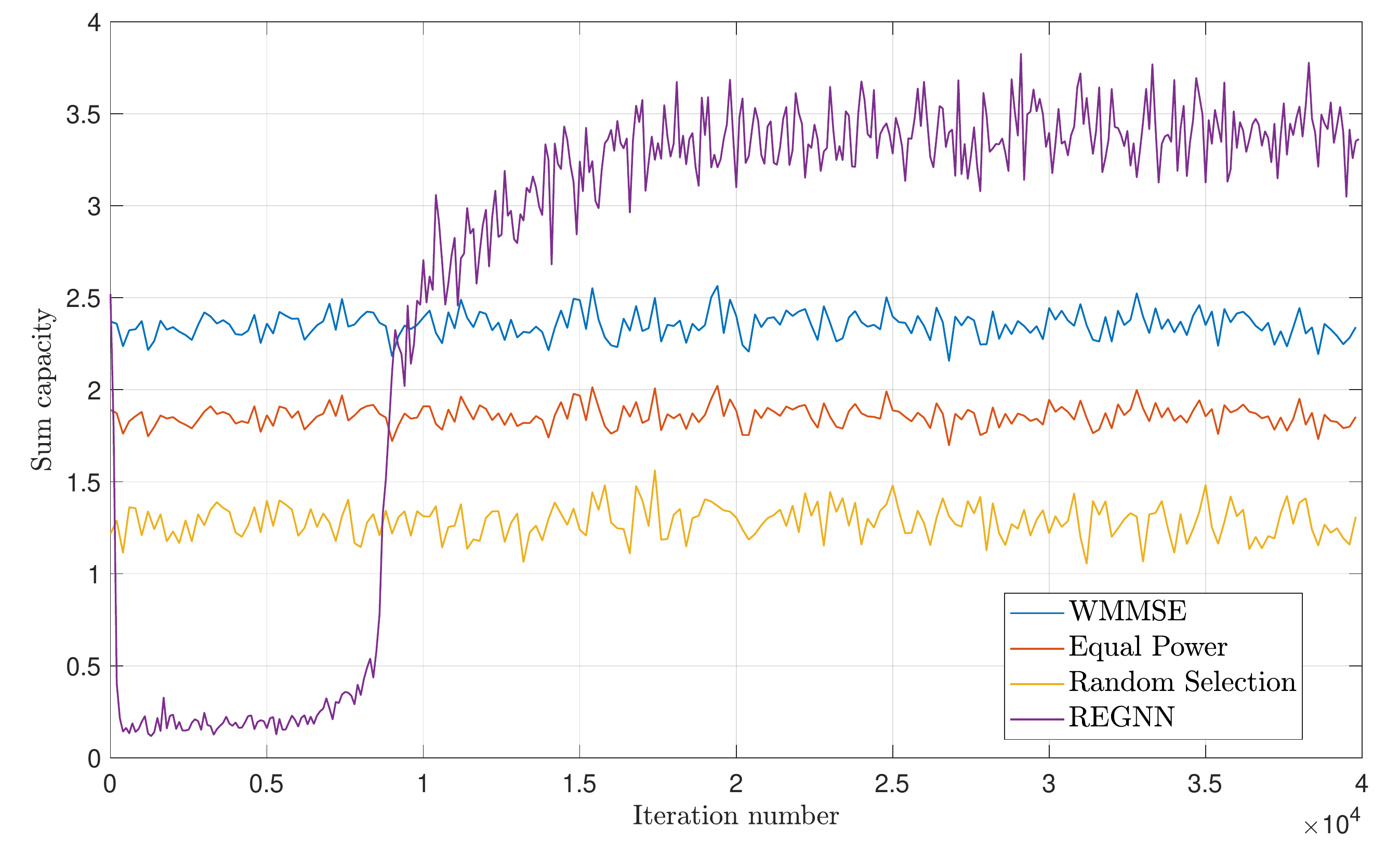}
\caption{Performance of GNN during training for $m=50$ pairs, in comparison with three heuristic baselines: WMMSE \cite{shi2011iteratively}, equal power division across all users and across a random subset of users.
%Mean and percentile sum-rates achieved by the GNN trained on network of size $n=50$ and executed on networks of increasing size $n^{\prime}$ in comparison to heuristic baselines: WMMSE \cite{shi2011iteratively}, equal power division across all users and across a random subset of users.
}
\label{fig_compare_2}
\end{figure}

Figs. \ref{fig_compare} and \ref{fig_compare_2} are training curves for the solution of \eqref{eq_param_problem} with a REGNN parametrization \cite{Eisen20-REGNN}. For comparison, training curves for a fully connected neural network are also shown along with heuristics \cite{shi2011iteratively}. Fig. \ref{fig_compare} considers 20 communicating pairs. It is notable that both, the REGNN and the fully connected neural network outperform existing heuristics and attain similar performance. The advantage of the REGNN is that it utilizes a smaller number of parameters. In Fig. \ref{fig_compare_2} we consider 50 communicating pairs. The REGNN still outperforms standard heuristics. Missing from this picture is a curve for a fully connected neural network. This is because it fails to train in a network of this size. 

The formulation in \eqref{eq_interference_problem1}-\eqref{eq_param_problem} can be generalized to different rate functions and it can be modified to incorporate constraints and network state representations. We refer the interested reader to \cite{Eisen20-REGNN}.

%% file: 99_conclusions.tex
% !TEX root = root.tex

%%%%%%%%%%%%%%%%%%%%%%%%%%%%%%%%%%%%%%%%%%%%%%%%%%%%%%%%%%%%%%%%%%%%%%%%%%%%%%%%
%   S   E   C   T   I   O   N   %%%%%%%%%%%%%%%%%%%%%%%%%%%%%%%%%%%%%%%%%%%%%%%%
%%%%%%%%%%%%%%%%%%%%%%%%%%%%%%%%%%%%%%%%%%%%%%%%%%%%%%%%%%%%%%%%%%%%%%%%%%%%%%%%
%
\section{Conclusions} \label{sec:conclusions}

Graph neural networks (GNNs) are becoming the tool of choice for the processing of signals supported on graphs. In this paper we have shown that GNNs are minor variations of graph convolutional filters. They differ in the incorporation of pointwise nonlinear functions and the addition of multiple layers. Being minor variations of graph filters, the good empirical performance of GNNs is expected: we have ample evidence supporting the usefulness of graph filters. What is unexpected is the appearance of significant gains for what is such a minor variation. In this paper we attempted to explain this phenomenon with a perturbation stability analysis showing that pointwise nonlinearities make it possible to discriminate signals while retaining robustness with respect to perturbations of the graph.

We further introduced graphon filters and graphon neural networks so as to understand the limit behavior of GNNs. This analysis uncovers the ability to transfer a GNN across graphs with different numbers of nodes. As in the case of our stability analysis, we discovered that GNNs exhibit more robust transferability than linear graph filters.

In both domains there remains much to be done. For instance, our stability analysis has much to say about perturbation of eigenvalues of a graph shift operator but little to say about the perturbation of its eigenvectors. There are also other ways of defining graph limits that are not graphons and several other GNN architectures whose fundamental properties have not been studied. We hope that this contribution can spark interest in understanding the fundamental properties of GNNs.